\documentclass{article}

\usepackage[preprint]{neurips_2025}

\usepackage[utf8]{inputenc} 
\usepackage[T1]{fontenc}    
\usepackage{hyperref}       
\usepackage{url}            
\usepackage{booktabs}       
\usepackage{amsfonts}       
\usepackage{nicefrac}       
\usepackage{microtype}      
\usepackage[table,xcdraw]{xcolor}

\usepackage{hhline}
\usepackage{boldline}
\usepackage{color}
\usepackage{soul}
\usepackage[most]{tcolorbox}
\usepackage{enumitem}
\setlist[itemize]{topsep=-0.18cm}
\usepackage{wrapfig}
\usepackage{amsmath}
\usepackage{caption}
\usepackage{algorithm}
\usepackage{algorithmic}
\usepackage[most]{tcolorbox}  
\usepackage{graphicx}
\usepackage{subcaption}

\usepackage{pifont}

\definecolor{lightgrey}{rgb}{0.925, 0.925, 0.925}
\sethlcolor{lightgrey}

\newcommand{\argmax}{\mathop{\mathrm{arg\,max}}\limits}

\newcommand{\icml}[1]{\textcolor{black}{#1}}
\newcommand{\edit}[1]{\textcolor{black}{#1}}

\newcommand{\yunfei}[1]{\textcolor{black}{#1}} 
\newcommand{\leo}[1]{\textcolor{black}{#1}} 

\title{Toward Foundation Model for Multivariate Wearable Sensing of Physiological Signals}

\author{Yunfei Luo$^{1,}$\thanks{These authors contributed equally to this work.} 
, Yuliang Chen$^{1,}$\footnotemark[1]
, Asif Salekin$^{2}$
, Tauhidur Rahman$^{1}$ \\
$^{1}$Halıcıoğlu Data Science Institute, University of California San Diego\\
$^{2}$Schools of Engineering, Arizona State University\\
}

\begin{document}
\tcbset{
  promptbox/.style={
    colback=gray!10,    
    colframe=gray!50,   
    boxrule=0.4pt,
    arc=2pt,
    left=6pt,
    right=6pt,
    top=4pt,
    bottom=4pt,
    enhanced,
    sharp corners
  }
}

\maketitle

\vspace{-4mm}
\begin{abstract}
Time-series foundation models excel at tasks like forecasting across diverse data types by leveraging informative waveform representations. Wearable sensing data, however, pose unique challenges due to their variability in patterns and frequency bands, especially for healthcare-related outcomes. The main obstacle lies in crafting generalizable representations that adapt efficiently across heterogeneous sensing configurations and applications.
\edit{To address this, we propose \textsc{NormWear}, the first multi-modal and ubiquitous foundation model designed to extract generalized and informative representations from wearable sensing data. 
Specifically, we design a channel-aware attention mechanism 
with a shared special liaison [CLS] token
to detect signal patterns in both intra-sensor and inter-sensors. This helps the model to extract more meaningful information considering both time series themselves and the relationships between input sensors. 
This helps the model to be widely compatible with various sensors settings. 
}
\textsc{NormWear} is pretrained on a diverse set of physiological signals, including PPG, ECG, EEG, GSR, and IMU, from various public datasets. 
\edit{Our model shows exceptional generalizability across 11 public wearable sensing datasets, spanning 18 applications in mental health, body state inference, vital sign estimation, and disease risk evaluation. 
It consistently outperforms competitive baselines under zero-shot, partial-shot, and full-shot settings, indicating broad applicability in real-world health applications. 
}
The code is available: \url{https://github.com/Mobile-Sensing-and-UbiComp-Laboratory/NormWear}.
\end{abstract}

\vspace{-2mm}
\section{Introduction} \label{sec:intro}
\vspace{-1mm}
Mobile and wearable sensors have been shown to be valuable for the field of healthcare by passively and continuously tracking physiological signals such as photoplethysmography (PPG) for pulse, electrocardiography (ECG) for heart activity, galvanic skin response (GSR), and electroencephalography (EEG) for brain activity. These time series signals are beneficial for early diagnosis, personalized health insights, and remote patient monitoring \citep{ts-large-survey}.

\begin{figure*}[ht]
\centering
\includegraphics[width=0.95\textwidth]{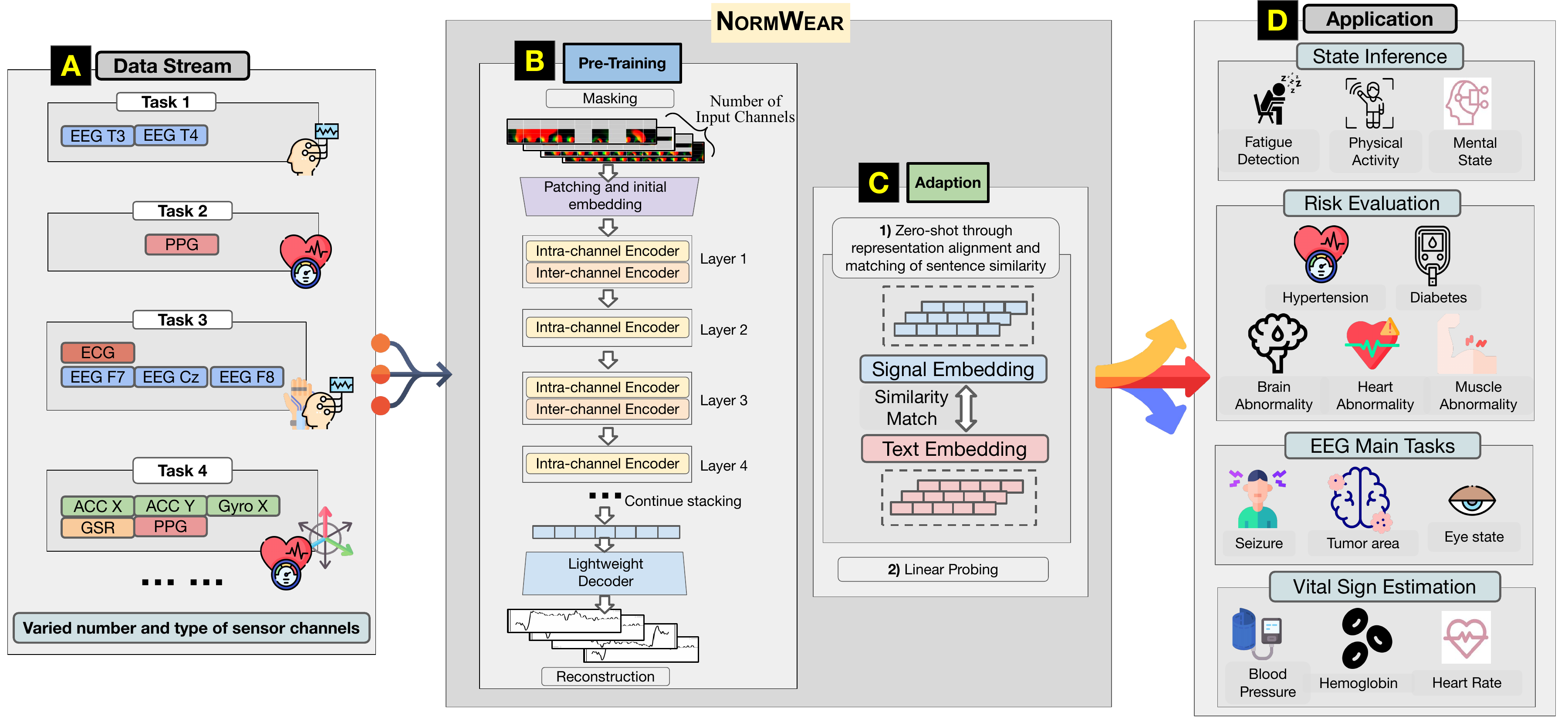}
\caption{\edit{The role of our framework. 
Several icons from \citet{medi-icons,ts-large-survey}.)}}
\label{fig:overview}
\vspace{-6mm}
\end{figure*}

Recently, several foundation models have emerged for time series modeling, including \citet{chronos,apple-ppg-ecg,UniTS,convtran}.
Another common approach for signal modeling involves converting raw signal series into 2D images or spectrograms, using fixed-size sliding windows, followed by the use of visual encoders like Vision Transformers (ViT) to extract representations for making inferences
\citep{img-ts,ts-as-plot,mel-spectrum-music,sft-ecg-user,sft-eeg-sleep,vit}. 
\edit{These works have significantly advanced the field and provided valuable insights, yet two main issues still exists which need further exploration to fully understand their potential in wearable scenarios.}
First, contrastive learning-based foundation models \citep{apple-ppg-ecg} rely on a predefined set of input signal types, making them unsuitable when transferring to scenarios with different types and numbers of sensors. Second, while both time series foundation models \citep{chronos,tfc,UniTS} and \edit{spectral}-based approaches \citep{img-ts,ts-as-plot} attempt to address this issue by training a generic encoder that can handle type-agnostic series, they remain limited to processing only univariate series. Because of this constraint, these previous works fail to account for the heterogeneity of multivariate input data; specifically, they do not capture the complex relationships between signals from sensors located on different body parts. These two limitations of recent approaches hinder their generalization and usefulness for wearable health monitoring. 

Moreover, Wearable-based multimodal physiological signals present unique challenges that distinguish them from general time series data, such as stock prices or weather patterns. \edit{Wearable signal modalities, such as PPG and EEG, vary in characteristics like dimensionality, sampling rate, and resolution, often requiring modality-specific preprocessing. Existing methods tokenize raw signals \citep{chronos,tfc} or convert them into image or spectral representations \citep{clap,spec-ecg,ts-img-ecg}. While effective for specific tasks, these approaches lack generalizability and fail to provide a consistent preprocessing pipeline across multiple modalities. A consistent framework that accommodates diverse signal requirements is essential for training deep learning-based foundation models and advancing multimodal signal analysis.}

In this work, we present \textsc{NormWear}, a normative foundation model, aiming to learn effective wearable sensing representations, addressing the above-discussed research gaps.
\textsc{NormWear} has been pretrained on more than \edit{2.5 million multivariate wearable sensing segments, comprising total of 14,943 hours of sensor signal series, using publicibly avaliable datasets. We evaluated \textsc{NormWear}} on \edit{18} public downstream tasks against \edit{competitive} baselines across \edit{zero-shot, few-show, and full-shot settings. } 
Overall, our contributions with the proposed \textsc{NormWear} healthcare modeling framework can be summarized as follows:
\begin{itemize}[itemsep=2.5pt, parsep=0pt]
    \item To our knowledge, we are the first to develop a foundation model specifically designed for wearable sensing data, capable of processing \edit{arbitrary configuration} of multivariate signals from sources such as the heart, skin, brain, and physical body.
    \item \textsc{NormWear} comprises novel methodologies built upon the advanced practice in both the fields of signal processing and deep learning, including (a) continuous wavelet transform (CWT) based multi-scale representations for modality- and number-agnostic tokenization, (b) channel-aware attention layer that enables the model to process arbitrary multivariate inputs, and (c) a human sensing adapted fusion mechanism that \edit{enabled \textsc{NormWear} to achieve zero-shot inference on health related wearable sensing tasks.}
    \item We are also the first to integrate and process a comprehensive wearable signals dataset \edit{with varied number of input channels} for training self-supervised learning algorithms, with thorough downstream evaluation. These datasets cover key health applications, including mental and physical state inference, vital sign estimation, and disease risk evaluation. 
\end{itemize}

Our proposed \textsc{NormWear} aims to provide a generalized data representation solution for smart health monitoring, benefiting the general public, and serving as a fundamental tool for researchers and professionals to address future healthcare challenges. 
\edit{We made the code and cleaned data to be publicly available to spur reproducible research.}

\vspace{-2mm}
\section{Related Work}
\vspace{-2mm}
\yunfei{Foundation models have emerged as a transformative paradigm in machine learning, enabling generalizable and reusable representations across diverse downstream tasks \citep{foundation_model}.
In the time series domain, recent works \citep{chronos,convtran,apple-ppg-ecg,google-scale-foundation} have demonstrated success in tasks such as forecasting, classification, and anomaly detection. However, their generalizability to health-related wearable signals remains limited due to the lack of in-depth evaluation, reliance on specific sensor types \citep{cbramod,labram,biot} and univariate data \citep{papagei,ecgfm}, as well as the inability to handle the heterogeneity of multivariate wearable signals. 
In contrast, \textsc{NormWear} builds upon these principles by introducing a modeling framework that is agnostic to the sensor modality and number of input channels, as stated in section \ref{sec:intro}, and is presented in details in section \ref{sec:method}. \textsc{NormWear} has been evaluated on 18 digital healthcare tasks and demonstrate peak performance against solid time series modeling baselines, including common statistical approach, SoTA model in time series with self-supervised learning \citep{tfc}, SoTA spectrum based modeling approach \citep{clap}, and SoTA time series forecasting model \citep{chronos}. Our work not only generalizes to arbitrary sensor configurations but also ensures compatibility across multivariate data, addressing key limitations of earlier approaches.
}

\vspace{-2mm}
\section{Method} \label{sec:method}
\vspace{-2mm}
\subsection{Dataset construction for model pretraining and downstream evaluation}
\label{sec2.1}
\vspace{-2mm}
We curated a collection of \edit{9} publicly available datasets (\edit{Table \ref{tab:pretrain-info})} exclusively for model pretraining, resulting in approximately \edit{230,962} multivariate time series segments, \edit{comprising 4,294 hours of total sensor signal series}, across various modalities, including PPG, ECG, EEG, GSR, PCG, and inertial measurement unit (IMU) data. \edit{To address the dataset size limitation, we then applied herustic data augmentation (algorithm \ref{alg:augment_single_series}) to expand the pretrain dataset to 2.5 million \edit{segments, comprising 14,943 hours of total sensor signal series}. Notably, each sample segment may contain a variable number of input channels depending on the sensor signals provided by the respective datasets. This input configuration aligns seamlessly with our model's design, which is optimized to flexibly handle arbitrary numbers and configurations of sensor signal inputs.}

To prevent potential data leakage in downstream tasks, we evaluate our model's transferability using an additional \edit{11} publicly available datasets encompassing \edit{18} modeling tasks, 
which include affective state classification, physical state recognition, biological estimation, and disease risk evaluation. 
Details about the datasets \edit{is presented in Table \ref{tab:downstream-info}}.


\vspace{-2mm}
\subsection{\edit{Tokenization}}
\vspace{-2mm}
Tokenization is a fundamental term widely used in natural language processing. In the context of wearable sensing, we leverage this term 
to represent the stage of signal processing before sending the processed data to the deep learning-based encoder. 
\edit{
Spectral methods, which utilize the short-time Fast Fourier Transform (FFT) \citep{fft} with a sliding window to compute spectrograms, are widely regarded as the benchmark approach for tokenization.
However, due to the inherent trade-off between time and frequency resolution, the spectral representation with a fixed window size cannot be generalized. This is because the window size has to be modulated accordingly when the modality varies.}
To enhance transferability, we propose a well-designed \edit{signal processing} pipeline that preserves information in both the frequency and time domains across multiple scales. We begin by calculating the first and second derivatives for each single signal series, as suggested by \citet{ppg-bp-resnet}, followed by computing the continuous wavelet transform (CWT) on both the raw and derivative series, resulting in three scalograms. Then, we stack the three scalograms to form data in RGB-image-like format. The derivatives capture the rate of signal change at different moments, while the wavelet transform provides a multi-resolution encoding that preserves information from both the time and frequency domains \citet{wavelet-analsis}. 
For the wavelet transform, we use the Mexican Hat wavelet for signal convolution, as recommended by previous studies \citep{ricker-ecg,ricker-ppg,ricker-gsr,ricker-eeg,ricker-acc}. We apply scales ranging from 1 to 64, following the guidance of \citep{har-cwt,har-cwt2}, which sufficiently covers most frequency bands of interest for physiological signals. \edit{Finally, this RGB-like scalogram is divided into patches, which is treated in the same way as tokens in an ViT \citep{vit}.}
In this way, this tokenization approach can be applied to various types of sensing signals without sensor-specific adjustments or reconfigurations. 

\begin{figure*}[t!]
\vspace{-4mm}
\centering
\includegraphics[width=0.95\textwidth]{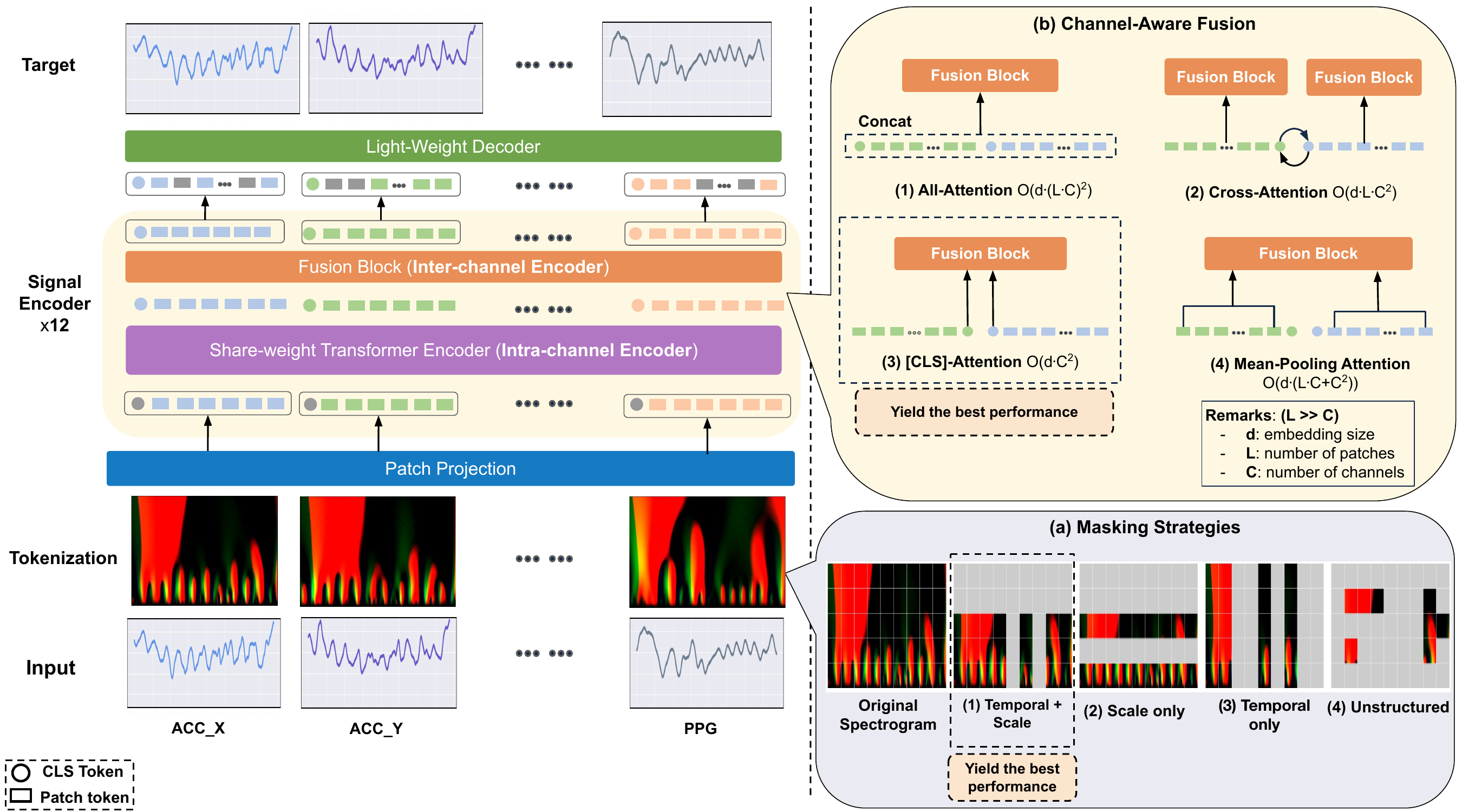}
\vspace{-2mm}
\caption{Overview of the pretrain pipeline.}
\label{fig:pipeline}
\vspace{-5mm}
\end{figure*}

\vspace{-2mm}
\subsection{\edit{Share-weighted Encoder}}
\vspace{-2mm}
\edit{Rather than concatenating tokens from all channels into a single long sequence and processing them with a full attention transformer, we treat each channel of the multivariate signal as an independent input stream. Although all channels share the same transformer backbone, the forward pass is executed separately for each one. This design allows the model to first learn the temporal characteristics of each sensor without interference from others. It not only reduces computational cost but also increases flexibility. Because each channel is processed independently, the model can be pretrained on datasets with varying numbers or types of sensors and later fine-tuned on a target task with a different sensor configuration.}
\vspace{-2mm}
\subsection{\edit{Channel-Aware Attention with Liaison Special Token}}
\vspace{-2mm}
\label{sec2.3}
Following the tokenization step, 
\edit{we adopt common reconstruction-based pretraining strategies from Masked Auto Encoder (MAE) \citep{mae,audiomae,ts-reconstruct-pretrain},
where input tokens are randomly masked and the model is trained to reconstruct the original time series using mean squared error (MSE) loss.
} 
Inspired by \citet{audiomae}, we experiment with four masking strategies, as shown in Figure \ref{fig:pipeline} (a), including masking on (1) temporal and scale, (2) scale only, (3) temporal only, and (4) unstructured axes.
We observe that the temporal and scalar masking yields the best performance for the downstream tasks. 
\edit{For the model architecture, we construct} the backbone of our proposed framework with a convolutional patching layer followed by 12 standard Transformer blocks \citep{attn-all-you-need}. For the same reason, \textsc{NormWear} uses a lightweight decoder consisting of 2 Transformer blocks, combined with a linear projection layer and a convolution layer to reconstruct the raw physiological signals both temporally and spatially. 
We also prepend a [CLS] token to each signal channel, following standard practice in transformer models, for learning a global representation of the input sequence for that channel. 

Another important point to consider is that although empirical studies \citep{patchtst,apple-ppg-ecg} show that channel-independent structures effectively capture local patterns, they fail to account for relationships across channels. To address this, \edit{we use the [CLS] token from each signal channel as a liaison token, allowing them to exchange information through the channel-aware fusion layer afrer every other encoder block.}
We explore several fusion approaches \edit{and different design of liaison token} as shown in Figure \ref{fig:pipeline} (b), with each method described below: 

\vspace{-1mm}
(1) \textbf{All-Attention Fusion:} 
This approach involves concatenating all tokens from each modality without considering their individual properties and fusing the information through a self-attention module. However, this method requires quadratic computation time, as every token passes through the self-attention module, making it impractical for real-world applications.

\vspace{-1mm}
(2) \textbf{Cross-Attention Fusion:} In addition to the cross-attention mechanism used in Cross-ViT \citep{chen2021crossvit}, we introduce a slight modification to fit in our problem setting. We propose a symmetric fusion method, using the \textsc[CLS] token from each modality as an intermediary to exchange information between the patch tokens of another modality, then projecting the information back to its original modality in the subsequent Transformer layer. While this strategy is efficient, it restricts the model to handling only two time series signals or modalities, which deviates from our goal of building a general model capable of processing an arbitrary number of channels.

\vspace{-1mm}
(3) \textbf{\textsc[CLS]-Attention Fusion} The \textsc[CLS] token serves as an abstract global representation for each signal modality. Here, we propose a hybrid fusion approach. We stack the \textsc[CLS] tokens from all signal modalities and perform feature fusion using a self-attention mechanism. The fused \textsc[CLS] token is then reattached to its original channel, enabling the newly learned information to be propagated to each patch token in subsequent transformer encoder layers.

\vspace{-1mm}
(4) \textbf{Mean-Pooling Fusion} Similar to the \textsc[CLS]-Attention Fusion approach, we employ mean-pooling within each channel instead of using the \textsc[CLS] token as an abstract global representation. 

Our empirical results show that \textsc[CLS]-attention fusion achieves the best downstreaming performance for our proposed \textsc{NormWear} model. Details of all the ablation studies are reported in Appendix \ref{appx:ablation}. \edit{Beyond accuracy, we want to emphasize that the [CLS]-Attention Fusion design is highly flexible. This flexibility arises from the fact that self-attention is length-flexible and permutation-invariant \citep{attn-all-you-need}. Consequently, it integrates naturally with our shared-weight encoder, allowing the model to accommodate a variable number of sensor channels presented in any order. We provide additional empirical evidence of NormWear's permutation invariance in Table \ref{tab:channel-permutation}, Appendix \ref{appx:ablation}.} 

\begin{figure*}[t!]
\vspace{-4mm}
\centering
\includegraphics[width=0.95\textwidth]{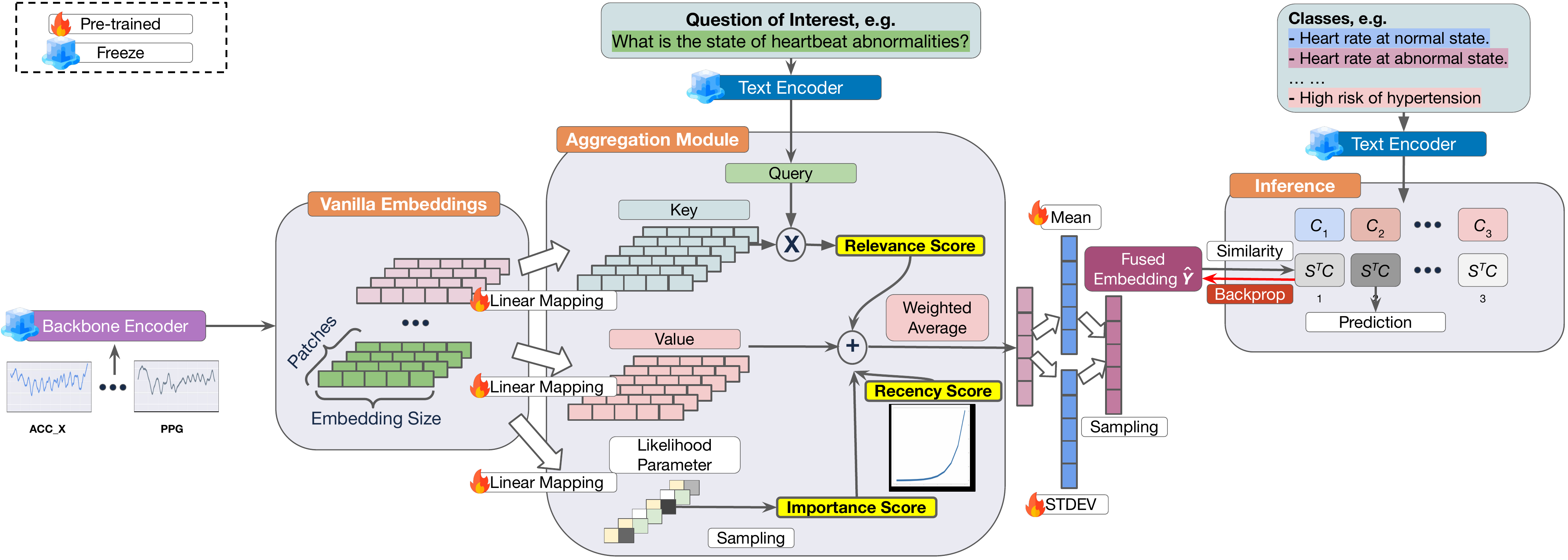}
\vspace{-1mm}
\caption{Memory stream inspired temporal fusion mechanism for representation alignment. }
\label{fig:fusion_pipeline}
\vspace{-4mm}
\end{figure*}

\vspace{-2mm}
\subsection{\edit{Sensor-Semantic Representation Alignment}}
\vspace{-2mm}
\edit{
Zero-shot inference is an important aspect to evaluate foundation model. 
We evaluate our model in this setting by retrieving the closest text-derived label for each unseen task in the shared embedding space. Specifically, to unify information across diverse modalities, we incorporate a representation alignment objective that encourages the embeddings of physiological sensor data to reside in the same latent space as paired textual descriptions. Once this shared space is established, it naturally supports zero-shot inference by allowing unseen sensor inputs to be interpreted through their proximity to text-derived anchors, without additional task-specific training.
Several important works in this direction focusing on domains of vision-language \cite{clip}, audio-language \cite{clap}, and motion-language \citep{unimts}. These works leverage end-to-end training to bind their modality of interest into semantic space. 
In this work, we extend this methodology to explore \textsc{Normwear}'s ability to generalize across unseen datasets and tasks. 
} 

\edit{
Building on prior work in representation alignment, we notice that in healthcare-related tasks where flexible inference across diverse scenarios is often required, the ground truth labels often have substantial overlap. For instance, depression is inferred from stress levels \citep{stress-depress}, and running and cycling produce similar IMU signals \citep{imu-adapt}. Due to these nested relationships, it create potential challenge to representation alignment when using 
contrastive learning, which requires clearly defined positive and negative pairs.
To address this, we first propose a novel way to fuse the signal representations together with improved qualities, then align the representation with vector distance as an auxiliary loss for contrastive learning method. 
} 
In addition, to reduce computation cost and counteract the issue of catastrophic forgetting \citep{blip2}, we use off-the-shelf frozen encoders for both signal and text modalities. 

\edit{Human physiological signals are task-specific, dynamic, and often weakly labeled \citep{he2018weakly,kim2022detector,qian2021weakly,ma2021unsupervised}. To address these characteristics, we introduce three complementary scoring mechanisms during feature aggregation: \textit{relevance scores} prioritize patches aligned with the task objective (e.g., IMU for activity recognition), guided by query sentences such as ``\textit{What activity is the subject doing?}''; \textit{recency scores} emphasize recent segments to better reflect the current physiological or emotional state \citep{roelofs2017freeze,ppg-bp-temporal-pattern,human-infect-recency}; and \textit{importance scores} weigh signal segments that contain meaningful or transient patterns often buried in weakly labeled sequences. Together, these scores guide the MSiTF fusion module to generate compact, task-aware representations. This design is inspired by memory-stream retrieval mechanisms \citep{memory-stream} and is tailored to the demands of human-centered sensing tasks such as risk assessment, affect detection, and activity recognition.}

\textbf{Memory Stream inspired Temporal Fusion (MSiTF). }
Our Aggregation or Fusion Module, MSiTF, is designed to addresses the above-discussed three challenges through three scores discussed below.
Specifically, we denote $f$ as the function that takes the semantic embedding of query sentence $q$ and backbone output $H \in \mathbb{R}^{P \times E}$ as input, where $P$ is the patch size and $E$ is the embedding size, thus having the final fused representation $f(q, H) = \hat{Y} \in \mathbb{R}^E$.

We define the \textit{Relevance} score as the cross attention between the key representations of each sensor time step and the query sentence embedding, obtained from a pretrained language model (Clinical TinyLlama \citep{clinical-tinyllama}). This mechanism allows the model to identify distinct but contextually relevant segments in the sensor input.
For the \textit{Recency} score, we use an exponential decay function to reflect the intuition that recent time steps are more important than earlier ones.
Finally, we consider the importance score \edit{IMP} in this case to be whether to keep the representation at each time step or not. In order to achieve this, we assign binary parameters to each time step, denoted as $\theta_t = p(v_t) \in \mathbb{R}^2$ where $v_t \in \mathbb{R}^E$ is the representation vector at time step $t$ and $p$ is a \edit{trainable} linear transformation function \edit{which will be optimized during pretraining}. We then have the importance score \edit{for each patch} defined as 
\vspace{-2mm}
\begin{equation}
\label{eq:importance}
W_{imp}(t) = \argmax_{i \in \{0, 1\}} \frac{\exp \bigg( \Big(\log (\theta_{t,i}) + \epsilon \Big) / \tau \bigg)}{\sum_{j \in \{0, 1\}} \exp \bigg( \Big(\log (\theta_{t, j}) + \epsilon_j \Big) / \tau \bigg)}
\end{equation}
where $\epsilon$ is the noise term sampled from Gumbel distribution \citep{gumbel-trick}, and $\tau$ is the temperature controlling the sharpness of the softmax function. Because $\argmax$ is not a differentiable function, we will directly take the resulting probability corresponding to index at $j=1$ \edit{to be the \textit{importance} score,} with $\tau$ being set to a small number to push the result closer to one hot vector from the softmax function. 
\yunfei{
As a result, this logit function will determine to what extent to activate the gate during forward pass on each patch of the input signals. 
}
\edit{The final score for each patch is the summation of the three scores as described above. This score will be treated as the weight for aggregating the representations from all the patches to form the fixed length embedded output (vector with size of 768 in our case). 
}

\edit{Once the signal embeddings are aggregated, we adopt a variational-inspired approach \citep{vae}. 
This design injects stochasticity into the representation, encouraging the model to explore and capture nuanced variations in semantic representations.
Finally, we leverage contrastive learning with auxiliary loss on vector distance to train the MSiTF module with a projection layer to text representation on the pretraining datasets. 
The sentence template formation and training details are presented in Appendix \ref{app:ctl_loss}.
}

\vspace{-2mm}
\section{Experiments}
\vspace{-2mm}
\edit{\textsc{NormWear} is pretrained exclusively on the data shown in Table \ref{tab:pretrain-info}. 
In this section, we present a comprehensive evaluation across \edit{11} downstream publicly available datasets, focusing on \edit{18} widely-recognized digital healthcare tasks. We evaluate the methods following order of zero-shot capability, partial-shot learning, and full-shot learning. }

\vspace{-2mm}
\subsection{\edit{Selection of baselines covering representative modeling strategies}} \label{sec:baseline-model}
\vspace{-2mm}
\edit{Modeling multivariate wearable signals with arbitrary input channels and sensor types, such as those capturing activities of heart, brain, and body physical motions, presents unique challenges, as no universally recognized open-source baseline or state-of-the-art (SoTA) model exists in this domain. To evaluate our approach, we selected diverse and representative baselines (as shown in Table \ref{tab:baseline}).} 

In the literature, various modeling strategies have been proposed. Firstly, early approaches involved handcrafting statistical features, which was a widely adopted practice in signal processing \citep{nld,uci-har,studenglife-gps}.  We include this simple baseline as sanity check.
Secondly, since sensory data can be naturally represented as time series \citep{UniTS, img-ts}, we benchmarked our model against Chronos \citep{chronos} , as well as a self-supervised framework TF-C \citep{tfc}.
Finally, the spectrum-based modeling methods \citep{mel-spectrum-music,sft-ecg-user,sft-eeg-sleep} are widely used for signal modeling. Therefore, we incorporate CLAP \citep{clap} into baselines that has demonstrates SoTA performance in spectrogram-based modeling. 
\icml{Regarding the comparison with concurrent works proposing foundation models for a specific sensor modality, we leverage PaPaGei \citep{papagei} for PPG datasets, ECG-FM \citep{ecgfm} for ECG datasets, and CBraMod \citep{cbramod} for EEG datasets. }
These baselines span distinct paradigms, providing a solid foundation to demonstrate the strengths of our model in wearable signal tasks.
\icml{For uni-modal baselines like Chronos and CLAP, we feed each signal separately into model and concatenate their representations after the forward pass. This ensures that all models have the same field of view, making the comparison fair. }



\vspace{-2mm}
\subsection{Zero-shot Evaluation}
\vspace{-2mm}

We achieve zero-shot inference by pretraining our proposed novel temporal fusion module on the task of representation alignment. 
\yunfei{We include the SoTA spectral-based model CLAP \cite{clap} as a baseline to provide a more comprehensive comparison of the results. For CLAP, we experimented with both Manhattan distance (MD) and dot product (DP) as similarity metrics during inference.
We observe that there are no statistically significant differences in performance when using MD and DP for label retrieval in CLAP. 
}
From table \ref{tab:zero-shot-res}, we could observe that overall, \textsc{NormWear} equipped with MSiTF outperforms the baselines. 
We compare \textsc{NormWear} with a few ablations by removing importance score (w/o IMP) and removing text augmentation (w/o text aug). We can observe that performance drop after removing each of the above components, verifying their respective importance in improving generalization across various downstream tasks. 
\edit{We present this outcome to demonstrate the zero-shot capability in the wearable signal domain, an aspect not present in recent studies. We also hope this outcome could potentially provide a new perspective that can help drive progress in this direction within the field.} 

\renewcommand{\arraystretch}{1.2}
\begin{table*}[ht!]
\vspace{-1mm}
\normalsize
\centering
\caption{
\edit{Zero-shot performance on the downstream datasets, with AUC ROC being reported. The last two columns show the average across the tasks and across group types respectively.}
}
\vspace{-1mm}
\resizebox{0.95\textwidth}{!}{
\begin{tabular}{|l|l|l|l|l|l|l|l|l|l|l|l|l|l|l|l|l|l|}
\hlineB{2}
\hlineB{2}
\bf Model&
\cellcolor{blue!20} \rotatebox{85}{WESAD}&
\cellcolor{blue!20} \rotatebox{85}{UCI-HAR}&
\cellcolor{blue!20} \rotatebox{85}{DriverFatigue}&
\cellcolor{teal!20} \rotatebox{85}{GAMEEMO}&
\cellcolor{teal!20} \rotatebox{85}{Epilepsy (eye open state)}&
\cellcolor{teal!20} \rotatebox{85}{Epilepsy (eye relaxation)}&
\cellcolor{teal!20} \rotatebox{85}{Epilepsy (health area)}&
\cellcolor{teal!20} \rotatebox{85}{Epilepsy (tumor area)}&
\cellcolor{teal!20} \rotatebox{85}{Epilepsy (seizure)}&
\cellcolor{red!20} \rotatebox{85}{PPG-BP (HTN)}&
\cellcolor{red!20} \rotatebox{85}{PPG-BP (DM)}&
\cellcolor{red!20} \rotatebox{85}{PPG-BP (CVA)}&
\cellcolor{red!20} \rotatebox{85}{PPG-BP (CVD)}&
\cellcolor{red!20} \rotatebox{85}{ECG-Abnormal}&
\cellcolor{red!20} \rotatebox{85}{PhysioNet EMG}&
\rotatebox{85}{Micro Avg.}&
\rotatebox{85}{Macro Avg.} \\
\hline
\hlineB{2}
\hlineB{2}
\textsc{CLAP} - MD&
45.3&
62.8&
58.5&
53.1&
44.9&
45.1&
47.6&
30.5&
\bf 84.9&
\bf 59.4&
41.8&
46.0&
57.4&
22.9&
55.4&
50.4&
51.2  
\\\hline
\textsc{CLAP} - DP&
50.7&
52.3&
\bf 61.1&
51.6&
54.4&
41.9&
58.6&
46.4&
74.3&
52.2&
41.4&
50.6&
58.9&
42.7&
38.3&
51.7&
52.2 
\\\hline
\hlineB{2}
\hlineB{2}
\begin{tabular}{@{\hspace{-0.5\tabcolsep}}l@{\hspace{-0.5\tabcolsep}}}
    before bind
\end{tabular}&
44.1&
48.2&
52.1&
48.4&
54.1&
62.6&
53.9&
52.5&
24.6&
48.8&
49.6&
46.3&
56.8&
54.3&
48.2&
49.6&
49.4
\\\hline
\begin{tabular}{@{\hspace{-0.5\tabcolsep}}l@{\hspace{-0.5\tabcolsep}}}
     \textbf{\textsc{NormWear}} \\
    \scriptsize w/ MSiTF
\end{tabular}&
55.8 &
71.2 &
57.2 &
51.0 &
\bf 55.7 &
61.3 &
\bf 67.6 &
55.8 &
66.0 &
57.1 &
\bf 62.5 &
\bf 70.0 &
59.0 &
63.1 &
70.1 & 
\bf \cellcolor{gray!60}61.6 &
\bf \cellcolor{gray!60}61.5
\\\hline
\begin{tabular}{@{\hspace{-0.5\tabcolsep}}l@{\hspace{-0.5\tabcolsep}}}
   - w/o IMP
\end{tabular}&
56.2&
70.3&
55.4&
49.8&
54.0&
56.5&
66.9&
\bf 57.3&
52.9&
56.5&
54.3&
61.7&
\bf 60.7&
\bf 73.4&
65.2&
\cellcolor{gray!30}59.4&
\cellcolor{gray!30}59.6
\\\hline
\begin{tabular}{@{\hspace{-0.5\tabcolsep}}l@{\hspace{-0.5\tabcolsep}}}
    - w/o text aug
\end{tabular}&
54.8&
65.8&
55.2&
49.2&
31.0&
58.4&
58.6&
32.8&
58.1&
50.2&
52.6&
50.8&
50.6&
47.7&
33.6&
50.0&
51.4 
 \\\hline
\begin{tabular}{@{\hspace{-0.5\tabcolsep}}l@{\hspace{-0.5\tabcolsep}}}
    - w/o refine
\end{tabular}&
\bf 59.5&
\bf 72.8&
42.7&
\bf 57.3&
50.6&
\bf 69.0&
43.3&
50.5&
74.8&
48.3&
38.8&
44.6&
44.1&
72.4&
\bf 75.7&
56.3&
56.6
 \\
\hlineB{2}
\hlineB{2}
\end{tabular}}
\label{tab:zero-shot-res}
\vspace{-2mm}
\end{table*}

\vspace{-1mm}
\begin{figure*}[t!]
\vspace{-4mm}
\centering
\includegraphics[trim=0 170 0 100, clip, width=0.95\textwidth]{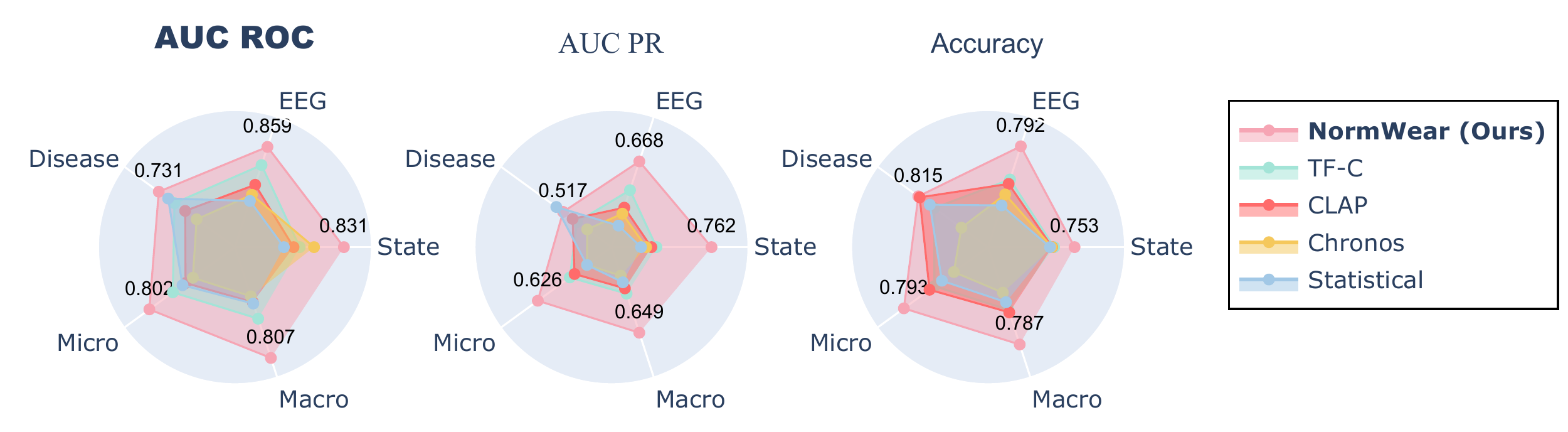}
\vspace{-2mm}
\caption{\textbf{Overview of performance trend of \textsc{Normwear}, \edit{under full-shot linear probing}, against competitive baselines in downstream tasks:} (1) Disease risk predictions. (2) EEG main tasks (mental and abnormal states prediction). 
(3) State recognition: physical and mental activities. (4) Macro: Average performance over types of tasks. 
(5) Micro: Average performance over each task.
}
\label{fig:radar-plot}
\vspace{-4mm}
\end{figure*}

\begin{table*}[t!]
\vspace{-2mm}
\scriptsize
\centering
\caption{
\textbf{Detailed performance on various downstream wearable-signal-based health related applications under full-shot linear probing evaluation.}
}
\vspace{-1mm}
\resizebox{0.99\textwidth}{!}{
        \begin{tabular}{|l|c|c|c|c|c|c|}
        \hlineB{2}
        \hlineB{2}
        \bf \begin{tabular}{@{\hspace{-0.5\tabcolsep}}l@{\hspace{-0.5\tabcolsep}}}
            Downstream
            Tasks \\ 
        \end{tabular} &
        \bf Statistical
       &
        \bf Chronos
       &
        \bf CLAP 
       &
        \bf TF-C
       &
        \bf Modality-Specific 
       &
        \bf \begin{tabular}{@{\hspace{-0.5\tabcolsep}}l@{\hspace{-0.5\tabcolsep}}}
            \textsc{Normwear} (Ours) \\
        \end{tabular}  
        \\\hline
        \hlineB{2}
        \cellcolor{blue!20} WESAD&
        66.213&71.489&72.383&69.865&56.656&\bf 76.060
        \\\hline
        \cellcolor{blue!20} UCI-HAR&
        95.784&91.593&96.420&96.892&-&\bf 98.954
        \\\hline
        \cellcolor{blue!20} DriverFatigue&
        63.249&76.722&61.889&66.882&\bf 80.430&74.292
        \\\hline
        \hlineB{2}
        \bf Activity Recognition Avg.&
        75.082&\cellcolor{gray!40}79.935&76.897&\cellcolor{gray!20}77.880&-&\cellcolor{gray!40}\bf 83.102
        \\\hline
        \hlineB{2}
        \hlineB{2}
        \cellcolor{teal!20} Epilepsy (eye open state)&
        82.489&82.41&85.094&89.153&90.436&\bf 92.743
        \\\hline
        \cellcolor{teal!20} Epilepsy (eye relaxation)&
        87.457&88.218&89.867&94.416&\bf 95.552&94.828
        \\\hline
        \cellcolor{teal!20} Epilepsy (health area)&
        86.274&81.08&83.711&85.619&88.065&\bf 88.541
        \\\hline
        \cellcolor{teal!20} Epilepsy (tumor area)&
        82.816&81.034&83.644&86.348&\bf87.258 &87.197
        \\\hline
        \cellcolor{teal!20} Epilepsy (seizure)&
        88.272&97.572&\bf 97.734&93.998&94.616&97.053
        \\\hline
        \cellcolor{teal!20} GAMEEMO&
        51.009&53.747&52.551&\bf 56.275&55.420&54.937
        \\\hline
        \hlineB{2}
        \bf EEG Main Tasks Avg.&
        79.720&80.677&\cellcolor{gray!20}82.100&\cellcolor{gray!40}84.302&85.225&\cellcolor{gray!60}\bf 85.883
        \\\hline
        \hlineB{2}
        \hlineB{2}
        \cellcolor{red!20} ECG-Abnormal&
        97.092&98.585&97.23&98.275&89.898&\bf 99.140
        \\\hline
         \cellcolor{red!20} PPG-BP (HTN)&
         59.499&52.425&56.757&\bf 65.229&61.839&62.341
        \\\hline
        \cellcolor{red!20} PPG-BP (DM)&
         47.823&51.164&42.455&\bf 57.883&55.668&55.893
        \\\hline
        \cellcolor{red!20} PPG-BP (CVA)&
        71.250&50.278&51.667&58.125& \bf73.125&70.625
        \\\hline
        \cellcolor{red!20} PPG-BP (CVD)&
         51.219&58.31&50.91&\bf 58.674&49.066&51.773
        \\\hline
         \cellcolor{red!20} PhysioNet EMG&
         \bf 99.309&61.6&98.627&78.308&-&99.216
        \\\hline
        \hlineB{2}
        \bf Risk Evaluation Avg.&
        \cellcolor{gray!40}71.032&62.060&66.274&\cellcolor{gray!20}69.416&-&\cellcolor{gray!60}\bf73.165 
        \\\hline
        \hlineB{2}
        \hlineB{2}
        \cellcolor{orange!10}Noninvasive-BP&
        92.310&91.79&91.922&87.481&90.596&\bf 92.420 
        \\\hline
        \cellcolor{orange!10} PPG-Hgb&
        94.219&\bf 95.005&94.291&93.408&94.912&94.632
        \\\hline
        \cellcolor{orange!10} Fetal-fPCG&
        98.929&99.048&\bf	99.195&99.077&-&99.072
        \\\hline
        \hlineB{2}
        \bf Vital Signs Avg.&
        \cellcolor{gray!20}95.153&\cellcolor{gray!40}95.281&95.136&93.322&-&\cellcolor{gray!60}\bf 95.375
        \\\hline
        \hlineB{2}
        \bf Micro Avg.&
        \cellcolor{gray!20}78.623&76.782&78.130&\cellcolor{gray!40}79.773&-&\cellcolor{gray!60}\bf82.762
        \\\hline
        \bf Macro Avg.&
        \cellcolor{gray!20}80.247&79.488&80.103&\cellcolor{gray!40}81.230&-&\cellcolor{gray!60}\bf84.381
        \\\hline
        \hlineB{2}
        \end{tabular}
}
\label{tab:main-res}
\vspace{-2mm}
\end{table*}

\begin{table*}[t!]
\scriptsize
\centering
\caption{
\textbf{Baselines}
}
\label{tab:baseline}
\vspace{-2mm}
\resizebox{1.0\textwidth}{!}{
\begin{tabular}{|l|l|}
        \hlineB{2}
        \hlineB{2}
        \bf Baseline Methods&
        \bf 
        Modeling Strategies
        \\\hline
        \hlineB{2}
        \begin{tabular}{@{\hspace{-0.5\tabcolsep}}l@{\hspace{-0.5\tabcolsep}}}
            Modality Specific \citep{tfc}
        \end{tabular} 
        &
        \begin{tabular}{@{\hspace{-0.5\tabcolsep}}l@{\hspace{-0.5\tabcolsep}}}
            PaPaGei \citep{papagei}, ECG-FM \citep{ecgfm}, CBraMod \citep{cbramod}. 
        \end{tabular} 
        \\\hline
        \begin{tabular}{@{\hspace{-0.5\tabcolsep}}l@{\hspace{-0.5\tabcolsep}}}
            TF-C \citep{tfc}
        \end{tabular} 
        &
        \begin{tabular}{@{\hspace{-0.5\tabcolsep}}l@{\hspace{-0.5\tabcolsep}}}
            SoTA in TS SSL; modeling 
            time and frequency domain  
            information at same time.
        \end{tabular} 
        \\\hline
        \begin{tabular}{@{\hspace{-0.5\tabcolsep}}l@{\hspace{-0.5\tabcolsep}}}
            CLAP \citep{clap}
        \end{tabular} 
        &
        \begin{tabular}{@{\hspace{-0.5\tabcolsep}}l@{\hspace{-0.5\tabcolsep}}}
            SoTA in audio modeling;
            process signal as spectrogram 
        \end{tabular} 
        \\\hline
        \begin{tabular}{@{\hspace{-0.5\tabcolsep}}l@{\hspace{-0.5\tabcolsep}}}
            Chronos \citep{chronos}
        \end{tabular} 
        &
        \begin{tabular}{@{\hspace{-0.5\tabcolsep}}l@{\hspace{-0.5\tabcolsep}}}
            SoTA in TS forecasting,
            leverage LLM for modeling
        \end{tabular} 
        \\\hline
        \begin{tabular}{@{\hspace{-0.5\tabcolsep}}l@{\hspace{-0.5\tabcolsep}}}
            Statistical approach
        \end{tabular} 
        &
        \begin{tabular}{@{\hspace{-0.5\tabcolsep}}l@{\hspace{-0.5\tabcolsep}}}
            Reserve full interpretability
        \end{tabular} 
        \\\hline
        \hlineB{2}
    \end{tabular}
}
\vspace{-4mm}
\end{table*}

\vspace{-2mm}
\subsection{\edit{Partial-shot and Full-shot Evaluation}}
\label{sec:linear-prob-res}
\vspace{-2mm} 
\leo{We evaluate the learned representations using linear probing through supervised training on each downstream dataset, and report performance on the corresponding held-out test set. To ensure fair comparison, we use a unified evaluation protocol with identical hyperparameter settings and implementation across all models and the dataset \citep{harnet}. This design ensures that performance differences are not due to variations in learning rate, regularization, or data augmentation \citep{hp-tune}. }\edit{Specifically, the classification tasks, using logistic regression, are solved by Newton's method with conjugate gradient, with AUC ROC being reported as main metric. The regression (vital signs) tasks, using ridge regression, are solved by Cholesky's method with closed form solution, with relative accuracy being reported. } 
\edit{For partial-shot evaluation, we leverage 10\% of the training data for the linear probing, and detailed performance result is presented in Table \ref{tab:semi-supervised}. The full-shot evaluation results is presented in Table \ref{tab:main-res}. All scores are the higher the better.}

\edit{From Figure \ref{fig:radar-plot}, Table \ref{tab:main-res}, and Table \ref{tab:main-res-metrics}, we observe that \textsc{Normwear} consistently achieves peak performance across all task groups, including activity recognition, EEG signal analysis, disease risk evaluation, and vital sign estimation. Furthermore, its leading performance remains consistent across various evaluation metrics. 
Based on the macro-averaged total score across task groups, \textsc{Normwear} delivers a 3.9\% improvement over the state-of-the-art (SoTA) time-series self-supervised learning framework \citep{tfc}, a 5.3\% improvement over the SoTA spectrum-based modeling method \citep{clap}, a 6.1\% improvement over SoTA time-series forecasting models with LLM backbones \citep{chronos}, and a 5.2\% improvement over standard statistical baselines.
On larger datasets, \textsc{Normwear} significantly outperforms the statistical baseline by 9.0\% and 7.5\% for activity recognition and EEG brain activity monitoring tasks, respectively. On smaller datasets, it still achieves peak performance in disease risk evaluation. For vital sign estimation, all methods yield comparable results, suggesting inherent challenges in these regression tasks that warrant further investigation but are beyond the scope of this study. 
}

\icml{When comparing with recent modality specific foundation models, NormWear's main benefit is that it capture cross-modal relationships, making it more versatile for wearable health tasks. While it sacrifices modality-specific optimization for adaptability, this may slightly reduce performance in highly specialized tasks. Single-signal models excel in their domains due to deeper modality-focused training. Instead of maximizing single-modality data, we prioritize signal diversity for better generalization. Benchmarking shows that NormWear, trained on a smaller dataset than EEG-only models, still achieves competitive results, highlighting the effectiveness of our pre-training approach.}
\edit{These findings illustrate \textsc{Normwear}'s capacity to balance consistency and adaptability across a diverse range of tasks and conditions. By excelling across standard benchmarks while addressing the intricacies of varied applications, \textsc{Normwear} exemplifies the philosophy of a foundation model: a reliable generalist capable of performing robustly across both typical and challenging scenarios.}

\begin{figure*}[ht!]
  \vspace{-2mm}
  \centering
  \begin{subfigure}[t]{0.48\textwidth}
    \centering
    \includegraphics[width=\textwidth]{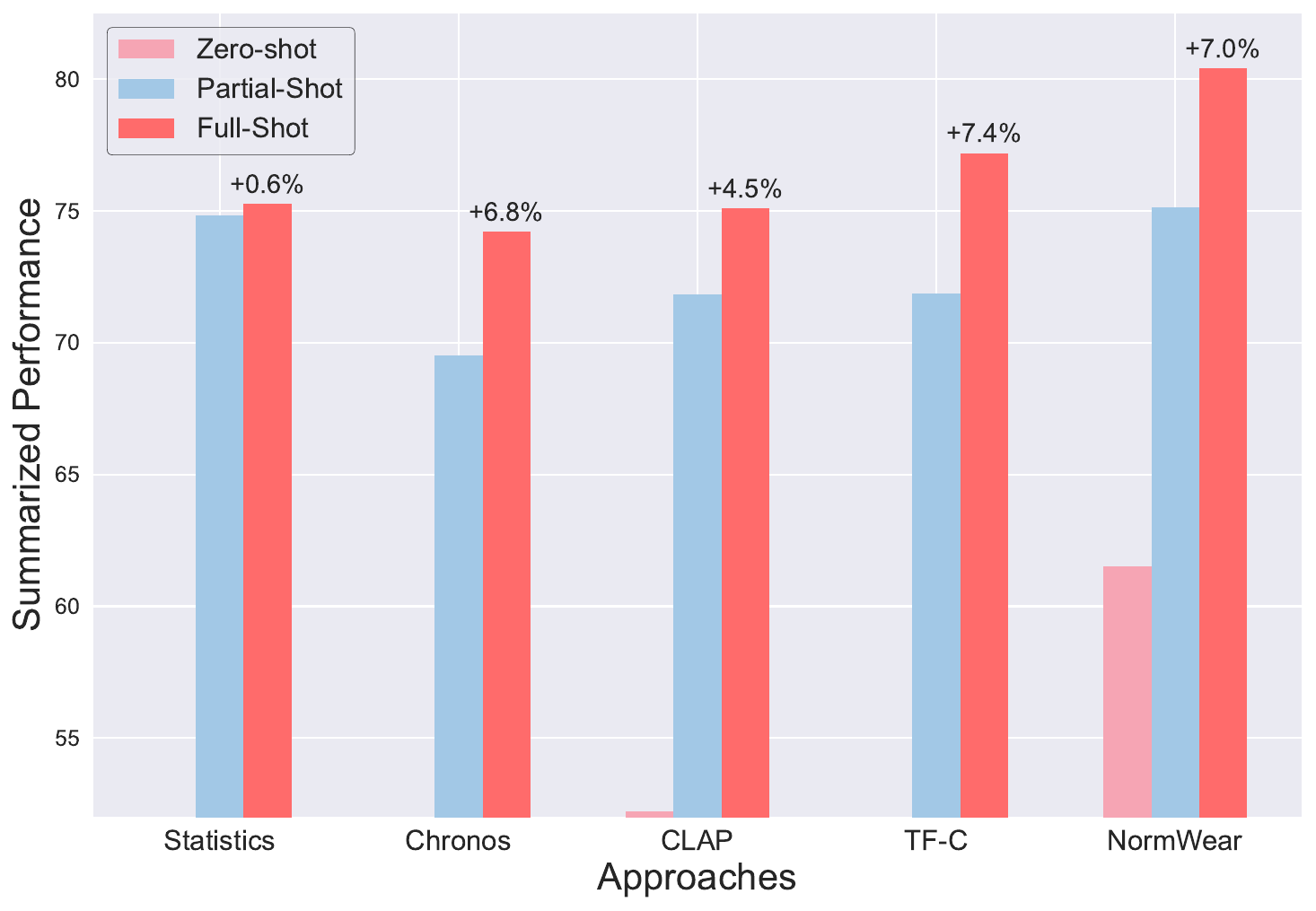}
    \caption{\edit{Adaptation summary}}
    \label{fig:adapt-summary}
  \end{subfigure}
  \hfill
  \begin{subfigure}[t]{0.5\textwidth}
    \centering
    \raisebox{3mm}{\includegraphics[width=\textwidth]{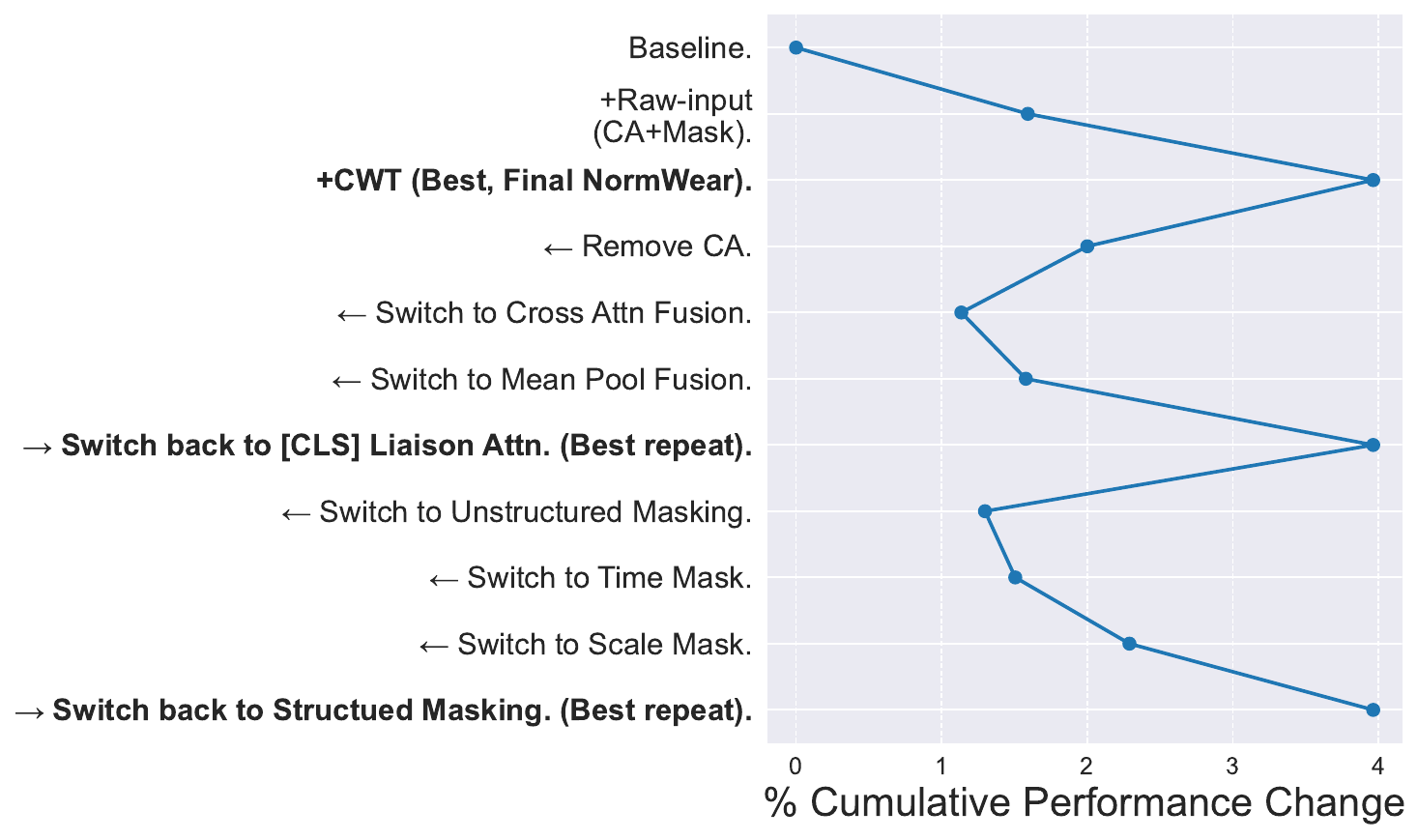}}
    \caption{\edit{Inspect performance contribution. `CA' refers to channel-aware attention.}}
    \label{fig:ablation-waterfall}
  \end{subfigure}
  \vspace{-2mm}
  \caption{\edit{Summary of adaptation performance and module-level performance contributions. Details of ablation study results are presented in Appendix \ref{appx:ablation}.}}
  \label{fig:combined-figure}
  \vspace{-4mm}
\end{figure*}



\vspace{-1mm}
\section{Conclusion and Discussion}
\vspace{-3mm}
\textbf{Conclusion.}
In this work, we mainly propose a foundation model for wearable physiological signals. 
\textsc{NormWear} is a practical tool that could serve as a starting point for researchers and clinicians when tackling a problem with wearable based signal data. Our proposed model could extract informative representations from raw signal series, which can be leveraged for further machine learning modeling, clustering, embedding vector-based information retrieval, and deployment of real-time health states monitoring with minimal tuning. We've justified the utilizability and generalization of \textsc{NormWear} through an extensive evaluation of various ubiquitous health applications.
As for future works, it is important to leverage our framework on larger scale clinical applications and explore the applicability of embedding vectors as state representations for intervention modeling problems that comprise the decision-making process. 

\edit{
\textbf{Limitation and Future Work.}
We acknowledge several limitations to be addressed in future work. (1) The representation alignment component is currently trained on a limited set of healthcare-related objectives, and expanding the pretraining corpus with more diverse semantic labels may improve generalization. (2) While our design supports classification tasks well, adapting the framework for regression remains an open challenge, and future work may explore alternative formulations beyond label discretization. (3) NormWear currently focuses on physiological signals with relatively narrow frequency bands; extending its applicability to higher-frequency modalities such as audio or lower-resolution clinical summaries is a promising direction. 
}

\edit{
\textbf{Broad Impact.} \textsc{NormWear} is the first foundation model tailored for multivariate physiological signals that supports a wide range of wearable health tasks across sensor modalities, device types, and clinical applications. Through a unified CWT-based tokenization pipeline and a channel-aware fusion mechanism, it enables robust, modality-agnostic representation learning. Our extensive evaluation across zero-shot, partial-shot, and full-shot settings demonstrates NormWear’s strong generalizability and practical relevance. 
We believe NormWear provides a valuable resource for advancing foundation modeling in digital health and promoting more unified benchmarks in the community.
}

\section*{Ethics Statement}
This study contains applications in the field of healthcare. We ensured that all the data being used during pretraining and evaluations were made publicly available by the original authors, and all these works were cited properly.

\section*{Reproducibility Statement}
The full code base, comprising all the code and documentation required for model construction, pretraining, downstream evaluation, and reproduction of the experimental results, is published at \url{https://github.com/Mobile-Sensing-and-UbiComp-Laboratory/NormWear}. 

\bibliography{example_paper}

\begin{thebibliography}{90}
\providecommand{\natexlab}[1]{#1}
\providecommand{\url}[1]{\texttt{#1}}
\expandafter\ifx\csname urlstyle\endcsname\relax
  \providecommand{\doi}[1]{doi: #1}\else
  \providecommand{\doi}{doi: \begingroup \urlstyle{rm}\Url}\fi

\bibitem[Abbaspourazad et~al.(2023)Abbaspourazad, Elachqar, Miller, Emrani, Nallasamy, and Shapiro]{apple-ppg-ecg}
Abbaspourazad, S., Elachqar, O., Miller, A.~C., Emrani, S., Nallasamy, U., and Shapiro, I.
\newblock Large-scale training of foundation models for wearable biosignals.
\newblock \emph{arXiv preprint arXiv:2312.05409}, 2023.

\bibitem[Abuzairi et~al.(2024)Abuzairi, Vinia, Yudhistira, Rizkinia, and Eriska]{ppg-hgb}
Abuzairi, T., Vinia, E., Yudhistira, M.~A., Rizkinia, M., and Eriska, W.
\newblock A dataset of hemoglobin blood value and photoplethysmography signal for machine learning-based non-invasive hemoglobin measurement.
\newblock \emph{Data in Brief}, 52:\penalty0 109823, 2024.
\newblock ISSN 2352-3409.
\newblock \doi{https://doi.org/10.1016/j.dib.2023.109823}.

\bibitem[Achiam et~al.(2023)Achiam, Adler, Agarwal, Ahmad, Akkaya, Aleman, Almeida, Altenschmidt, Altman, Anadkat, et~al.]{chatgpt}
Achiam, J., Adler, S., Agarwal, S., Ahmad, L., Akkaya, I., Aleman, F.~L., Almeida, D., Altenschmidt, J., Altman, S., Anadkat, S., et~al.
\newblock Gpt-4 technical report.
\newblock \emph{arXiv preprint arXiv:2303.08774}, 2023.

\bibitem[Alakus et~al.(2020)Alakus, Gonen, and Turkoglu]{gameemo}
Alakus, T.~B., Gonen, M., and Turkoglu, I.
\newblock Database for an emotion recognition system based on eeg signals and various computer games--gameemo.
\newblock \emph{Biomedical Signal Processing and Control}, 60:\penalty0 101951, 2020.

\bibitem[Alzahab et~al.(2022)Alzahab, Di~Iorio, Apollonio, Alshalak, Gravina, Antognoli, Baldi, Scalise, and Alchalabi]{audi-eeg}
Alzahab, N.~A., Di~Iorio, A., Apollonio, L., Alshalak, M., Gravina, A., Antognoli, L., Baldi, M., Scalise, L., and Alchalabi, B.
\newblock Auditory evoked potential eeg-biometric dataset, 2022.

\bibitem[Andrzejak et~al.(2023)Andrzejak, Lehnertz, Rieke, Mormann, David, and Elger]{epilepsy}
Andrzejak, R.~G., Lehnertz, K., Rieke, C., Mormann, F., David, P., and Elger, C.~E.
\newblock {Indications of nonlinear deterministic and finite-dimensional structures in time series of brain electrical activity: Dependence on recording region and brain state [dataset]}.
\newblock \emph{Physical Review E}, 2023.
\newblock \doi{10.34810/data490}.
\newblock URL \url{https://doi.org/10.34810/data490}.

\bibitem[Ansari et~al.(2024)Ansari, Stella, Turkmen, Zhang, Mercado, Shen, Shchur, Rangapuram, Arango, Kapoor, et~al.]{chronos}
Ansari, A.~F., Stella, L., Turkmen, C., Zhang, X., Mercado, P., Shen, H., Shchur, O., Rangapuram, S.~S., Arango, S.~P., Kapoor, S., et~al.
\newblock Chronos: Learning the language of time series.
\newblock \emph{arXiv preprint arXiv:2403.07815}, 2024.

\bibitem[Bajaj et~al.(2020)Bajaj, Carri{\'o}n, and Bellotti]{phyaat}
Bajaj, N., Carri{\'o}n, J.~R., and Bellotti, F.
\newblock Phyaat: Physiology of auditory attention to speech dataset.
\newblock \emph{arXiv preprint arXiv:2005.11577}, 2020.

\bibitem[Beh et~al.(2021)Beh, Wu, and Wu]{maus}
Beh, W.-K., Wu, Y.-H., and Wu, A.-Y.~A.
\newblock Maus: A dataset for mental workload assessment on n-back task using wearable sensor, 2021.
\newblock URL \url{https://dx.doi.org/10.21227/q4td-yd35}.

\bibitem[Bhaskaran et~al.(2022)Bhaskaran, J, George, and Arora]{india-fpcg}
Bhaskaran, A., J, S.~K., George, S., and Arora, M.
\newblock Heart rate estimation and validation algorithm for fetal phonocardiography.
\newblock \emph{Physiological Measurement}, 43\penalty0 (7):\penalty0 075008, jul 2022.
\newblock \doi{10.1088/1361-6579/ac7a8c}.
\newblock URL \url{https://dx.doi.org/10.1088/1361-6579/ac7a8c}.

\bibitem[Bommasani et~al.(2022)Bommasani, Hudson, Adeli, Altman, Arora, von Arx, Bernstein, Bohg, Bosselut, Brunskill, Brynjolfsson, Buch, Card, Castellon, Chatterji, Chen, Creel, Davis, Demszky, Donahue, Doumbouya, Durmus, Ermon, Etchemendy, Ethayarajh, Fei-Fei, Finn, Gale, Gillespie, Goel, Goodman, Grossman, Guha, Hashimoto, Henderson, Hewitt, Ho, Hong, Hsu, Huang, Icard, Jain, Jurafsky, Kalluri, Karamcheti, Keeling, Khani, Khattab, Koh, Krass, Krishna, Kuditipudi, Kumar, Ladhak, Lee, Lee, Leskovec, Levent, Li, Li, Ma, Malik, Manning, Mirchandani, Mitchell, Munyikwa, Nair, Narayan, Narayanan, Newman, Nie, Niebles, Nilforoshan, Nyarko, Ogut, Orr, Papadimitriou, Park, Piech, Portelance, Potts, Raghunathan, Reich, Ren, Rong, Roohani, Ruiz, Ryan, Ré, Sadigh, Sagawa, Santhanam, Shih, Srinivasan, Tamkin, Taori, Thomas, Tramèr, Wang, Wang, Wu, Wu, Wu, Xie, Yasunaga, You, Zaharia, Zhang, Zhang, Zhang, Zhang, Zheng, Zhou, and Liang]{foundation_model}
Bommasani, R., Hudson, D.~A., Adeli, E., Altman, R., Arora, S., von Arx, S., Bernstein, M.~S., Bohg, J., Bosselut, A., Brunskill, E., Brynjolfsson, E., Buch, S., Card, D., Castellon, R., Chatterji, N., Chen, A., Creel, K., Davis, J.~Q., Demszky, D., Donahue, C., Doumbouya, M., Durmus, E., Ermon, S., Etchemendy, J., Ethayarajh, K., Fei-Fei, L., Finn, C., Gale, T., Gillespie, L., Goel, K., Goodman, N., Grossman, S., Guha, N., Hashimoto, T., Henderson, P., Hewitt, J., Ho, D.~E., Hong, J., Hsu, K., Huang, J., Icard, T., Jain, S., Jurafsky, D., Kalluri, P., Karamcheti, S., Keeling, G., Khani, F., Khattab, O., Koh, P.~W., Krass, M., Krishna, R., Kuditipudi, R., Kumar, A., Ladhak, F., Lee, M., Lee, T., Leskovec, J., Levent, I., Li, X.~L., Li, X., Ma, T., Malik, A., Manning, C.~D., Mirchandani, S., Mitchell, E., Munyikwa, Z., Nair, S., Narayan, A., Narayanan, D., Newman, B., Nie, A., Niebles, J.~C., Nilforoshan, H., Nyarko, J., Ogut, G., Orr, L., Papadimitriou, I., Park, J.~S., Piech, C., Portelance, E., Potts, C.,
  Raghunathan, A., Reich, R., Ren, H., Rong, F., Roohani, Y., Ruiz, C., Ryan, J., Ré, C., Sadigh, D., Sagawa, S., Santhanam, K., Shih, A., Srinivasan, K., Tamkin, A., Taori, R., Thomas, A.~W., Tramèr, F., Wang, R.~E., Wang, W., Wu, B., Wu, J., Wu, Y., Xie, S.~M., Yasunaga, M., You, J., Zaharia, M., Zhang, M., Zhang, T., Zhang, X., Zhang, Y., Zheng, L., Zhou, K., and Liang, P.
\newblock On the opportunities and risks of foundation models, 2022.
\newblock URL \url{https://arxiv.org/abs/2108.07258}.

\bibitem[Bousseljot et~al.(2009)Bousseljot, Kreiseler, and Schnabel]{ecg-abnormal}
Bousseljot, R., Kreiseler, D., and Schnabel, A.
\newblock Nutzung der ekg-signaldatenbank cardiodat der ptb {\"u}ber das internet.
\newblock In \emph{PTB-XL, a large publicly available electrocardiography dataset}, 2009.
\newblock URL \url{https://api.semanticscholar.org/CorpusID:111121953}.

\bibitem[Brigham(1988)]{fft}
Brigham, E.~O.
\newblock \emph{The fast Fourier transform and its applications}.
\newblock Prentice-Hall, Inc., 1988.

\bibitem[Burke \& Nasor(2004)Burke and Nasor]{ricker-ecg}
Burke, M. and Nasor, M.
\newblock Wavelet based analysis and characterization of the ecg signal.
\newblock \emph{Journal of Medical Engineering \& Technology}, 28\penalty0 (2):\penalty0 47--55, 2004.

\bibitem[Carmona et~al.(2021)Carmona, Aubet, Flunkert, and Gasthaus]{data_aug}
Carmona, C.~U., Aubet, F.-X., Flunkert, V., and Gasthaus, J.
\newblock Neural contextual anomaly detection for time series, 2021.
\newblock URL \url{https://arxiv.org/abs/2107.07702}.

\bibitem[Caron et~al.(2021)Caron, Misra, Mairal, Goyal, Bojanowski, and Joulin]{swav}
Caron, M., Misra, I., Mairal, J., Goyal, P., Bojanowski, P., and Joulin, A.
\newblock Unsupervised learning of visual features by contrasting cluster assignments, 2021.
\newblock URL \url{https://arxiv.org/abs/2006.09882}.

\bibitem[Chaudhury et~al.(2021)Chaudhury, Yu, Liu, Kumar, Hornby, Duplessis, Sklar, Epstein, and Reifman]{human-infect-recency}
Chaudhury, S., Yu, C., Liu, R., Kumar, K., Hornby, S., Duplessis, C., Sklar, J.~M., Epstein, J.~E., and Reifman, J.
\newblock Wearables detect malaria early in a controlled human-infection study.
\newblock \emph{IEEE Transactions on Biomedical Engineering}, 69\penalty0 (6):\penalty0 2119--2129, 2021.

\bibitem[Chen et~al.(2021)Chen, Fan, and Panda]{chen2021crossvit}
Chen, C.-F., Fan, Q., and Panda, R.
\newblock Crossvit: Cross-attention multi-scale vision transformer for image classification, 2021.

\bibitem[Chowdhury et~al.(2020)Chowdhury, Shuzan, Chowdhury, Mahbub, Uddin, Khandakar, and Reaz]{ppg-bp-temporal-pattern}
Chowdhury, M.~H., Shuzan, M. N.~I., Chowdhury, M.~E., Mahbub, Z.~B., Uddin, M.~M., Khandakar, A., and Reaz, M. B.~I.
\newblock Estimating blood pressure from the photoplethysmogram signal and demographic features using machine learning techniques.
\newblock \emph{Sensors}, 20\penalty0 (11):\penalty0 3127, 2020.

\bibitem[Chun et~al.(2016)Chun, Kang, Kim, Lee, Oakley, and Kim]{sft-ecg-user}
Chun, S.~Y., Kang, J.-H., Kim, H., Lee, C., Oakley, I., and Kim, S.-P.
\newblock Ecg based user authentication for wearable devices using short time fourier transform.
\newblock In \emph{2016 39th international conference on telecommunications and signal processing (tsp)}, pp.\  656--659. IEEE, 2016.

\bibitem[Dar et~al.(2022)Dar, Rahim, Akram, Khawaja, and Rahim]{yaad}
Dar, M.~N., Rahim, A., Akram, M.~U., Khawaja, S.~G., and Rahim, A.
\newblock Yaad: young adult’s affective data using wearable ecg and gsr sensors.
\newblock In \emph{2022 2nd International Conference on Digital Futures and Transformative Technologies (ICoDT2)}, pp.\  1--7. IEEE, 2022.

\bibitem[Dosovitskiy et~al.(2020)Dosovitskiy, Beyer, Kolesnikov, Weissenborn, Zhai, Unterthiner, Dehghani, Minderer, Heigold, Gelly, et~al.]{vit}
Dosovitskiy, A., Beyer, L., Kolesnikov, A., Weissenborn, D., Zhai, X., Unterthiner, T., Dehghani, M., Minderer, M., Heigold, G., Gelly, S., et~al.
\newblock An image is worth 16x16 words: Transformers for image recognition at scale.
\newblock \emph{arXiv preprint arXiv:2010.11929}, 2020.

\bibitem[Esmaili et~al.(2017)Esmaili, Kachuee, and Shabany]{ppg-bp}
Esmaili, A., Kachuee, M., and Shabany, M.
\newblock Nonlinear cuffless blood pressure estimation of healthy subjects using pulse transit time and arrival time.
\newblock \emph{IEEE Transactions on Instrumentation and Measurement}, 66\penalty0 (12):\penalty0 3299--3308, 2017.

\bibitem[Fekri~Azgomi et~al.(2023)Fekri~Azgomi, Branco, Amin, et~al.]{brain_cog}
Fekri~Azgomi, H., Branco, L. R.~F., Amin, M.~R., et~al.
\newblock Regulation of brain cognitive states through auditory, gustatory, and olfactory stimulation with wearable monitoring.
\newblock \emph{Scientific Reports}, 13:\penalty0 12399, 2023.
\newblock \doi{10.1038/s41598-023-37829-z}.
\newblock URL \url{https://doi.org/10.1038/s41598-023-37829-z}.

\bibitem[Foumani et~al.(2024)Foumani, Tan, Webb, and Salehi]{convtran}
Foumani, N.~M., Tan, C.~W., Webb, G.~I., and Salehi, M.
\newblock Improving position encoding of transformers for multivariate time series classification.
\newblock \emph{Data Mining and Knowledge Discovery}, 38\penalty0 (1):\penalty0 22--48, 2024.

\bibitem[Freepik(n.d.)]{medi-icons}
Freepik.
\newblock Hypertension; blood pressure gauge; motion sensor; student sleeping in class; diabetes; blood cells; edge computing; galvanic skin response; motion sensor; accelerometer sensor; eeg, n.d.
\newblock URL prefix: https://www.flaticon.com/free-icon/ , IDs: hypertension\_4939229; blood-pressure-gauge\_3184052; motion-sensor\_2818201; student-sleeping-in-class\_43739; diabetes\_2750352; blood-cells\_3400003; edge-computing\_11068838;galvanic-skin-response\_11228469; motion-sensor\_17881894; accelerometer-sensor\_11330476; eeg\_9851782.

\bibitem[Goldberger et~al.(2000)Goldberger, Amaral, Glass, Hausdorff, Ivanov, Mark, Mietus, Moody, Peng, and Stanley]{PhysioNet}
Goldberger, A.~L., Amaral, L. A.~N., Glass, L., Hausdorff, J.~M., Ivanov, P.~C., Mark, R.~G., Mietus, J.~E., Moody, G.~B., Peng, C.-K., and Stanley, H.~E.
\newblock {PhysioBank, PhysioToolkit, and PhysioNet}: Components of a new research resource for complex physiologic signals.
\newblock \emph{Circulation}, 101\penalty0 (23):\penalty0 e215--e220, 2000.
\newblock Circulation Electronic Pages: http://circ.ahajournals.org/content/101/23/e215.full PMID:1085218; doi: 10.1161/01.CIR.101.23.e215.

\bibitem[Hassani(2021)]{ricker-gsr}
Hassani, T.
\newblock Federated emotion recognition with physiological signals-gsr, 2021.

\bibitem[He et~al.(2018)He, Zhang, Wang, and Pei]{he2018weakly}
He, J., Zhang, Q., Wang, L., and Pei, L.
\newblock Weakly supervised human activity recognition from wearable sensors by recurrent attention learning.
\newblock \emph{IEEE Sensors Journal}, 19\penalty0 (6):\penalty0 2287--2297, 2018.

\bibitem[He et~al.(2021)He, Chen, Xie, Li, Dollár, and Girshick]{mae}
He, K., Chen, X., Xie, S., Li, Y., Dollár, P., and Girshick, R.
\newblock Masked autoencoders are scalable vision learners, 2021.
\newblock URL \url{https://arxiv.org/abs/2111.06377}.

\bibitem[Hosni \& Atef(2023)Hosni and Atef]{ricker-ppg}
Hosni, A. and Atef, M.
\newblock Remote real-time heart rate monitoring with recursive motion artifact removal using ppg signals from a smartphone camera.
\newblock \emph{Multimedia Tools and Applications}, 82\penalty0 (13):\penalty0 20571--20588, 2023.

\bibitem[Hu et~al.(2001)Hu, Ivanov, Chen, Carpena, and Stanley]{dfa}
Hu, K., Ivanov, P.~C., Chen, Z., Carpena, P., and Stanley, H.~E.
\newblock Effect of trends on detrended fluctuation analysis.
\newblock \emph{Physical Review E}, 64\penalty0 (1):\penalty0 011114, 2001.

\bibitem[Huang et~al.(2023)Huang, Xu, Li, Baevski, Auli, Galuba, Metze, and Feichtenhofer]{audiomae}
Huang, P.-Y., Xu, H., Li, J., Baevski, A., Auli, M., Galuba, W., Metze, F., and Feichtenhofer, C.
\newblock Masked autoencoders that listen, 2023.
\newblock URL \url{https://arxiv.org/abs/2207.06405}.

\bibitem[Jang et~al.(2017)Jang, Gu, and Poole]{gumbel-trick}
Jang, E., Gu, S., and Poole, B.
\newblock Categorical reparameterization with gumbel-softmax, 2017.

\bibitem[Jiang et~al.(2024)Jiang, Zhao, and Lu]{labram}
Jiang, W.-B., Zhao, L.-M., and Lu, B.-L.
\newblock Large brain model for learning generic representations with tremendous eeg data in bci, 2024.
\newblock URL \url{https://arxiv.org/abs/2405.18765}.

\bibitem[Jolliffe \& Cadima(2016)Jolliffe and Cadima]{pca}
Jolliffe, I.~T. and Cadima, J.
\newblock Principal component analysis: a review and recent developments.
\newblock \emph{Philosophical transactions of the royal society A: Mathematical, Physical and Engineering Sciences}, 374\penalty0 (2065):\penalty0 20150202, 2016.

\bibitem[Kachuee et~al.(2016)Kachuee, Kiani, Mohammadzade, and Shabany]{cuff-less}
Kachuee, M., Kiani, M.~M., Mohammadzade, H., and Shabany, M.
\newblock Cuffless blood pressure estimation algorithms for continuous health-care monitoring.
\newblock \emph{IEEE Transactions on Biomedical Engineering}, 64\penalty0 (4):\penalty0 859--869, 2016.

\bibitem[Kazemnejad et~al.(2021)Kazemnejad, Gordany, and Sameni]{EPHNOGRAM}
Kazemnejad, A., Gordany, P., and Sameni, R.
\newblock {EPHNOGRAM: A Simultaneous Electrocardiogram and Phonocardiogram Database (version 1.0.0)}, 2021.
\newblock URL \url{https://doi.org/10.13026/tjtq-5911}.

\bibitem[Kim et~al.(2022)Kim, Lee, Cho, and Kwak]{kim2022detector}
Kim, D., Lee, J., Cho, M., and Kwak, S.
\newblock Detector-free weakly supervised group activity recognition.
\newblock In \emph{Proceedings of the IEEE/CVF Conference on Computer Vision and Pattern Recognition}, pp.\  20083--20093, 2022.

\bibitem[Kingma \& Welling(2022)Kingma and Welling]{vae}
Kingma, D.~P. and Welling, M.
\newblock Auto-encoding variational bayes, 2022.
\newblock URL \url{https://arxiv.org/abs/1312.6114}.

\bibitem[Krishnan et~al.(2020)Krishnan, Yaacob, Krishnan, Rizon, and Ang]{sft-eeg-sleep}
Krishnan, P., Yaacob, S., Krishnan, A.~P., Rizon, M., and Ang, C.~K.
\newblock Eeg based drowsiness detection using relative band power and short-time fourier transform.
\newblock \emph{J. Robotics Netw. Artif. Life}, 7\penalty0 (3):\penalty0 147--151, 2020.

\bibitem[LeMoult(2020)]{stress-depress}
LeMoult, J.
\newblock From stress to depression: Bringing together cognitive and biological science.
\newblock \emph{Current Directions in Psychological Science}, 29\penalty0 (6):\penalty0 592--598, 2020.

\bibitem[Li et~al.(2019)Li, Derrode, and Pieczynski]{imu-adapt}
Li, H., Derrode, S., and Pieczynski, W.
\newblock An adaptive and on-line imu-based locomotion activity classification method using a triplet markov model.
\newblock \emph{Neurocomputing}, 362:\penalty0 94--105, 2019.

\bibitem[Li et~al.(2023)Li, Li, Savarese, and Hoi]{blip2}
Li, J., Li, D., Savarese, S., and Hoi, S.
\newblock {BLIP}-2: Bootstrapping language-image pre-training with frozen image encoders and large language models.
\newblock In Krause, A., Brunskill, E., Cho, K., Engelhardt, B., Sabato, S., and Scarlett, J. (eds.), \emph{Proceedings of the 40th International Conference on Machine Learning}, volume 202 of \emph{Proceedings of Machine Learning Research}, pp.\  19730--19742. PMLR, 23--29 Jul 2023.
\newblock URL \url{https://proceedings.mlr.press/v202/li23q.html}.

\bibitem[Liang et~al.(2018)Liang, Chen, Liu, and Elgendi]{ppg-china}
Liang, Y., Chen, Z., Liu, G., and Elgendi, M.
\newblock A new, short-recorded photoplethysmogram dataset for blood pressure monitoring in china.
\newblock \emph{Scientific data}, 5\penalty0 (1):\penalty0 1--7, 2018.
\newblock \doi{10.6084/m9.figshare.5459299.v5}.

\bibitem[Ma et~al.(2021)Ma, Zhang, Li, and Lu]{ma2021unsupervised}
Ma, H., Zhang, Z., Li, W., and Lu, S.
\newblock Unsupervised human activity representation learning with multi-task deep clustering.
\newblock \emph{Proceedings of the ACM on Interactive, Mobile, Wearable and Ubiquitous Technologies}, 5\penalty0 (1):\penalty0 1--25, 2021.

\bibitem[Mathew et~al.(2024)Mathew, Barbosa, Prince, and Venkatraman]{spec-ecg}
Mathew, G., Barbosa, D., Prince, J., and Venkatraman, S.
\newblock Foundation models for cardiovascular disease detection via biosignals from digital stethoscopes.
\newblock \emph{npj Cardiovascular Health}, 1\penalty0 (1):\penalty0 25, Oct 2024.
\newblock ISSN 2948-2836.
\newblock \doi{10.1038/s44325-024-00027-5}.
\newblock URL \url{https://doi.org/10.1038/s44325-024-00027-5}.

\bibitem[McKeen et~al.(2024)McKeen, Oliva, Masood, Toma, Rubin, and Wang]{ecgfm}
McKeen, K., Oliva, L., Masood, S., Toma, A., Rubin, B., and Wang, B.
\newblock Ecg-fm: An open electrocardiogram foundation model, 2024.
\newblock URL \url{https://arxiv.org/abs/2408.05178}.

\bibitem[Mikelsons et~al.(2017)Mikelsons, Smith, Mehrotra, and Musolesi]{studenglife-gps}
Mikelsons, G., Smith, M., Mehrotra, A., and Musolesi, M.
\newblock Towards deep learning models for psychological state prediction using smartphone data: Challenges and opportunities.
\newblock In \emph{ML4H Workshop at 31st Conference on Neural Information Processing Systems (NIPS)}, 2017.
\newblock URL \url{https://arxiv.org/abs/1711.06350}.

\bibitem[Min et~al.(2017)Min, Wang, and Hu]{driver-fatigue}
Min, J., Wang, P., and Hu, J.
\newblock {The original EEG data for driver fatigue detection}.
\newblock \emph{figshare.Dataset.}, 7 2017.
\newblock \doi{10.6084/m9.figshare.5202739.v1}.

\bibitem[Muzammil(2021)]{clinical-tinyllama}
Muzammil, M.
\newblock Finetuning endevsols/tinyllama-2.5t-clinical model on clinical dataset., 2021.
\newblock URL \url{https://huggingface.co/muzammil-eds/tinyllama-2.5T-Clinical-v2}.

\bibitem[Narayanswamy et~al.(2024)Narayanswamy, Liu, Ayush, Yang, Xu, Liao, Garrison, Tailor, Sunshine, Liu, Althoff, Narayanan, Kohli, Zhan, Malhotra, Patel, Abdel-Ghaffar, and McDuff]{google-scale-foundation}
Narayanswamy, G., Liu, X., Ayush, K., Yang, Y., Xu, X., Liao, S., Garrison, J., Tailor, S., Sunshine, J., Liu, Y., Althoff, T., Narayanan, S., Kohli, P., Zhan, J., Malhotra, M., Patel, S., Abdel-Ghaffar, S., and McDuff, D.
\newblock Scaling wearable foundation models, 2024.
\newblock URL \url{https://arxiv.org/abs/2410.13638}.

\bibitem[Nedorubova et~al.(2021{\natexlab{a}})Nedorubova, Kadyrova, and Khlyupin]{har-cwt2}
Nedorubova, A., Kadyrova, A., and Khlyupin, A.
\newblock Human activity recognition using continuous wavelet transform and convolutional neural networks.
\newblock \emph{arXiv preprint arXiv:2106.12666}, 2021{\natexlab{a}}.

\bibitem[Nedorubova et~al.(2021{\natexlab{b}})Nedorubova, Kadyrova, and Khlyupin]{ricker-acc}
Nedorubova, A., Kadyrova, A., and Khlyupin, A.
\newblock Human activity recognition using continuous wavelet transform and convolutional neural networks.
\newblock \emph{arXiv preprint arXiv:2106.12666}, 2021{\natexlab{b}}.

\bibitem[Negi et~al.(2024)Negi, Giri, Sharma, Sharma, et~al.]{ricker-eeg}
Negi, P.~C., Giri, H., Sharma, S., Sharma, N., et~al.
\newblock A comparative study of scalograms for human activity classification.
\newblock In \emph{2024 IEEE 4th International Conference on Human-Machine Systems (ICHMS)}, pp.\  1--5. IEEE, 2024.

\bibitem[Nie et~al.(2023)Nie, Nguyen, Sinthong, and Kalagnanam]{patchtst}
Nie, Y., Nguyen, N.~H., Sinthong, P., and Kalagnanam, J.
\newblock A time series is worth 64 words: Long-term forecasting with transformers, 2023.
\newblock URL \url{https://arxiv.org/abs/2211.14730}.

\bibitem[Oliver et~al.(2018)Oliver, Odena, Raffel, Cubuk, and Goodfellow]{hp-tune}
Oliver, A., Odena, A., Raffel, C.~A., Cubuk, E.~D., and Goodfellow, I.
\newblock Realistic evaluation of deep semi-supervised learning algorithms.
\newblock In Bengio, S., Wallach, H., Larochelle, H., Grauman, K., Cesa-Bianchi, N., and Garnett, R. (eds.), \emph{Advances in Neural Information Processing Systems}, volume~31. Curran Associates, Inc., 2018.
\newblock URL \url{https://proceedings.neurips.cc/paper_files/paper/2018/file/c1fea270c48e8079d8ddf7d06d26ab52-Paper.pdf}.

\bibitem[Park et~al.(2023)Park, O'Brien, Cai, Morris, Liang, and Bernstein]{memory-stream}
Park, J.~S., O'Brien, J.~C., Cai, C.~J., Morris, M.~R., Liang, P., and Bernstein, M.~S.
\newblock Generative agents: Interactive simulacra of human behavior, 2023.

\bibitem[Pillai et~al.(2024)Pillai, Spathis, Kawsar, and Malekzadeh]{papagei}
Pillai, A., Spathis, D., Kawsar, F., and Malekzadeh, M.
\newblock Papagei: Open foundation models for optical physiological signals, 2024.
\newblock URL \url{https://arxiv.org/abs/2410.20542}.

\bibitem[Pimentel et~al.(2017)Pimentel, Johnson, Charlton, Birrenkott, Watkinson, Tarassenko, and Clifton]{bidmc}
Pimentel, M. A.~F., Johnson, A. E.~W., Charlton, P.~H., Birrenkott, D., Watkinson, P.~J., Tarassenko, L., and Clifton, D.~A.
\newblock Toward a robust estimation of respiratory rate from pulse oximeters.
\newblock \emph{IEEE Transactions on Biomedical Engineering}, 64\penalty0 (8):\penalty0 1914--1923, 2017.
\newblock \doi{10.1109/TBME.2016.2613124}.

\bibitem[Qian \& Rasheed(2004)Qian and Rasheed]{hurst-exponent}
Qian, B. and Rasheed, K.
\newblock Hurst exponent and financial market predictability.
\newblock In \emph{IASTED conference on Financial Engineering and Applications}, pp.\  203--209. Proceedings of the IASTED International Conference Cambridge, MA, 2004.

\bibitem[Qian et~al.(2021)Qian, Pan, and Miao]{qian2021weakly}
Qian, H., Pan, S.~J., and Miao, C.
\newblock Weakly-supervised sensor-based activity segmentation and recognition via learning from distributions.
\newblock \emph{Artificial Intelligence}, 292:\penalty0 103429, 2021.

\bibitem[Radford et~al.(2021)Radford, Kim, Hallacy, Ramesh, Goh, Agarwal, Sastry, Askell, Mishkin, Clark, Krueger, and Sutskever]{clip}
Radford, A., Kim, J.~W., Hallacy, C., Ramesh, A., Goh, G., Agarwal, S., Sastry, G., Askell, A., Mishkin, P., Clark, J., Krueger, G., and Sutskever, I.
\newblock Learning transferable visual models from natural language supervision, 2021.
\newblock URL \url{https://arxiv.org/abs/2103.00020}.

\bibitem[Reiss~Attila(2019)]{ppg-dalia}
Reiss~Attila, Indlekofer~Ina, S.~P.
\newblock {PPG-DaLiA}.
\newblock UCI Machine Learning Repository, 2019.
\newblock {DOI}: https://doi.org/10.24432/C53890.

\bibitem[Reyes-Ortiz et~al.(2012)Reyes-Ortiz, Anguita, Ghio, Oneto, and Parra]{uci-har}
Reyes-Ortiz, J., Anguita, D., Ghio, A., Oneto, L., and Parra, X.
\newblock {Human Activity Recognition Using Smartphones}.
\newblock UCI Machine Learning Repository, 2012.
\newblock {DOI}: https://doi.org/10.24432/C54S4K.

\bibitem[Roelofs(2017)]{roelofs2017freeze}
Roelofs, K.
\newblock Freeze for action: neurobiological mechanisms in animal and human freezing.
\newblock \emph{Philosophical Transactions of the Royal Society B: Biological Sciences}, 372\penalty0 (1718):\penalty0 20160206, 2017.

\bibitem[Schmidt et~al.(2018)Schmidt, Reiss, Duerichen, Marberger, and Van~Laerhoven]{wesad}
Schmidt, P., Reiss, A., Duerichen, R., Marberger, C., and Van~Laerhoven, K.
\newblock Introducing wesad, a multimodal dataset for wearable stress and affect detection.
\newblock In \emph{Proceedings of the 20th ACM international conference on multimodal interaction}, pp.\  400--408, 2018.

\bibitem[Semenoglou et~al.(2023)Semenoglou, Spiliotis, and Assimakopoulos]{img-ts}
Semenoglou, A.-A., Spiliotis, E., and Assimakopoulos, V.
\newblock Image-based time series forecasting: A deep convolutional neural network approach.
\newblock \emph{Neural Networks}, 157:\penalty0 39--53, 2023.
\newblock ISSN 0893-6080.
\newblock \doi{https://doi.org/10.1016/j.neunet.2022.10.006}.
\newblock URL \url{https://www.sciencedirect.com/science/article/pii/S0893608022003902}.

\bibitem[Sengupta et~al.(2022)Sengupta, Polian, and Hayes]{har-cwt}
Sengupta, R., Polian, I., and Hayes, J.~P.
\newblock Wavelet transform assisted neural networks for human activity recognition.
\newblock In \emph{2022 IEEE International Symposium on Circuits and Systems (ISCAS)}, pp.\  1254--1258. IEEE, 2022.

\bibitem[Slapni{\v{c}}ar et~al.(2019)Slapni{\v{c}}ar, Mlakar, and Lu{\v{s}}trek]{ppg-bp-resnet}
Slapni{\v{c}}ar, G., Mlakar, N., and Lu{\v{s}}trek, M.
\newblock Blood pressure estimation from photoplethysmogram using a spectro-temporal deep neural network.
\newblock \emph{Sensors}, 19\penalty0 (15):\penalty0 3420, 2019.

\bibitem[Thompson et~al.(1990)Thompson, Stewart, and Turner]{nld-chaos}
Thompson, J. M.~T., Stewart, H.~B., and Turner, R.
\newblock Nonlinear dynamics and chaos.
\newblock \emph{Computers in Physics}, 4\penalty0 (5):\penalty0 562--563, 1990.

\bibitem[Torrence \& Compo(1998)Torrence and Compo]{wavelet-analsis}
Torrence, C. and Compo, G.~P.
\newblock A practical guide to wavelet analysis.
\newblock \emph{Bulletin of the American Meteorological society}, 79\penalty0 (1):\penalty0 61--78, 1998.

\bibitem[Vaid et~al.(2023)Vaid, Jiang, Sawant, Lerakis, Argulian, Ahuja, Lampert, Charney, Greenspan, Narula, Glicksberg, and Nadkarni]{ts-img-ecg}
Vaid, A., Jiang, J., Sawant, A., Lerakis, S., Argulian, E., Ahuja, Y., Lampert, J., Charney, A., Greenspan, H., Narula, J., Glicksberg, B., and Nadkarni, G.~N.
\newblock A foundational vision transformer improves diagnostic performance for electrocardiograms.
\newblock \emph{npj Digital Medicine}, 6\penalty0 (1):\penalty0 108, Jun 2023.
\newblock ISSN 2398-6352.
\newblock \doi{10.1038/s41746-023-00840-9}.
\newblock URL \url{https://doi.org/10.1038/s41746-023-00840-9}.

\bibitem[Van~der Maaten \& Hinton(2008)Van~der Maaten and Hinton]{tsne}
Van~der Maaten, L. and Hinton, G.
\newblock Visualizing data using t-sne.
\newblock \emph{Journal of machine learning research}, 9\penalty0 (11), 2008.

\bibitem[Vaswani et~al.(2023)Vaswani, Shazeer, Parmar, Uszkoreit, Jones, Gomez, Kaiser, and Polosukhin]{attn-all-you-need}
Vaswani, A., Shazeer, N., Parmar, N., Uszkoreit, J., Jones, L., Gomez, A.~N., Kaiser, L., and Polosukhin, I.
\newblock Attention is all you need, 2023.
\newblock URL \url{https://arxiv.org/abs/1706.03762}.

\bibitem[Vishnupriya \& Meenakshi(2018)Vishnupriya and Meenakshi]{mel-spectrum-music}
Vishnupriya, S. and Meenakshi, K.
\newblock Automatic music genre classification using convolution neural network.
\newblock In \emph{2018 International Conference on Computer Communication and Informatics (ICCCI)}, pp.\  1--4, 2018.
\newblock \doi{10.1109/ICCCI.2018.8441340}.
\newblock URL \url{https://ieeexplore.ieee.org/document/8441340}.

\bibitem[Wang et~al.(2025)Wang, Zhao, Luo, Zhou, Jiang, Li, Li, and Pan]{cbramod}
Wang, J., Zhao, S., Luo, Z., Zhou, Y., Jiang, H., Li, S., Li, T., and Pan, G.
\newblock Cbramod: A criss-cross brain foundation model for eeg decoding, 2025.
\newblock URL \url{https://arxiv.org/abs/2412.07236}.

\bibitem[Wimmer \& Rekabsaz(2023)Wimmer and Rekabsaz]{ts-as-plot}
Wimmer, C. and Rekabsaz, N.
\newblock Leveraging vision-language models for granular market change prediction, 2023.
\newblock URL \url{https://arxiv.org/abs/2301.10166}.

\bibitem[Wolf et~al.(1985)Wolf, Swift, Swinney, and Vastano]{lyapunov-exponent}
Wolf, A., Swift, J.~B., Swinney, H.~L., and Vastano, J.~A.
\newblock Determining lyapunov exponents from a time series.
\newblock \emph{Physica D: nonlinear phenomena}, 16\penalty0 (3):\penalty0 285--317, 1985.

\bibitem[Woo et~al.(2024)Woo, Liu, Kumar, Xiong, Savarese, and Sahoo]{UniTS}
Woo, G., Liu, C., Kumar, A., Xiong, C., Savarese, S., and Sahoo, D.
\newblock Unified training of universal time series forecasting transformers, 2024.
\newblock URL \url{https://arxiv.org/abs/2402.02592}.

\bibitem[Wu et~al.(2023)Wu, Chen, Zhang, Hui, Berg-Kirkpatrick, and Dubnov]{clap}
Wu, Y., Chen, K., Zhang, T., Hui, Y., Berg-Kirkpatrick, T., and Dubnov, S.
\newblock Large-scale contrastive language-audio pretraining with feature fusion and keyword-to-caption augmentation.
\newblock In \emph{ICASSP 2023-2023 IEEE International Conference on Acoustics, Speech and Signal Processing (ICASSP)}, pp.\  1--5. IEEE, 2023.

\bibitem[Yan et~al.(2023{\natexlab{a}})Yan, Huang, Zhao, Liu, Ma, Yang, Yan, Xiong, and Wang]{nld}
Yan, Y., Huang, Y.-C., Zhao, J., Liu, Y.-S., Ma, L., Yang, J., Yan, X.-D., Xiong, J., and Wang, L.
\newblock Topological nonlinear analysis of dynamical systems in wearable sensor-based human physical activity inference.
\newblock \emph{IEEE Transactions on Human-Machine Systems}, 53\penalty0 (4):\penalty0 792--801, 2023{\natexlab{a}}.
\newblock \doi{10.1109/THMS.2023.3275774}.

\bibitem[Yan et~al.(2023{\natexlab{b}})Yan, Huang, Zhao, Liu, Ma, Yang, Yan, Xiong, and Wang]{persist-entropy}
Yan, Y., Huang, Y.-C., Zhao, J., Liu, Y.-S., Ma, L., Yang, J., Yan, X.-D., Xiong, J., and Wang, L.
\newblock Topological nonlinear analysis of dynamical systems in wearable sensor-based human physical activity inference.
\newblock \emph{IEEE Transactions on Human-Machine Systems}, 53\penalty0 (4):\penalty0 792--801, 2023{\natexlab{b}}.
\newblock \doi{10.1109/THMS.2023.3275774}.

\bibitem[Yang et~al.(2023)Yang, Westover, and Sun]{biot}
Yang, C., Westover, M., and Sun, J.
\newblock Biot: Biosignal transformer for cross-data learning in the wild.
\newblock In Oh, A., Naumann, T., Globerson, A., Saenko, K., Hardt, M., and Levine, S. (eds.), \emph{Advances in Neural Information Processing Systems}, volume~36, pp.\  78240--78260. Curran Associates, Inc., 2023.
\newblock URL \url{https://proceedings.neurips.cc/paper_files/paper/2023/file/f6b30f3e2dd9cb53bbf2024402d02295-Paper-Conference.pdf}.

\bibitem[Yuan et~al.(2024)Yuan, Chan, Creagh, Tong, Acquah, Clifton, and Doherty]{harnet}
Yuan, H., Chan, S., Creagh, A.~P., Tong, C., Acquah, A., Clifton, D.~A., and Doherty, A.
\newblock Self-supervised learning for human activity recognition using 700,000 person-days of wearable data.
\newblock \emph{npj Digital Medicine}, 7\penalty0 (1), April 2024.
\newblock ISSN 2398-6352.
\newblock \doi{10.1038/s41746-024-01062-3}.
\newblock URL \url{http://dx.doi.org/10.1038/s41746-024-01062-3}.

\bibitem[Zhang et~al.(2017)Zhang, Cisse, Dauphin, and Lopez-Paz]{zhang2017mixup}
Zhang, H., Cisse, M., Dauphin, Y.~N., and Lopez-Paz, D.
\newblock mixup: Beyond empirical risk minimization, 2017.

\bibitem[Zhang et~al.(2023)Zhang, Yang, Geng, and Hong]{ts-reconstruct-pretrain}
Zhang, W., Yang, L., Geng, S., and Hong, S.
\newblock Self-supervised time series representation learning via cross reconstruction transformer.
\newblock \emph{IEEE Transactions on Neural Networks and Learning Systems}, 2023.

\bibitem[Zhang et~al.(2022)Zhang, Zhao, Tsiligkaridis, and Zitnik]{tfc}
Zhang, X., Zhao, Z., Tsiligkaridis, T., and Zitnik, M.
\newblock Self-supervised contrastive pre-training for time series via time-frequency consistency.
\newblock \emph{Advances in Neural Information Processing Systems}, 35:\penalty0 3988--4003, 2022.

\bibitem[Zhang et~al.(2024{\natexlab{a}})Zhang, Chowdhury, Gupta, and Shang]{ts-large-survey}
Zhang, X., Chowdhury, R.~R., Gupta, R.~K., and Shang, J.
\newblock Large language models for time series: A survey.
\newblock \emph{arXiv preprint arXiv:2402.01801}, 2024{\natexlab{a}}.

\bibitem[Zhang et~al.(2024{\natexlab{b}})Zhang, Teng, Chowdhury, Li, Hong, Gupta, and Shang]{unimts}
Zhang, X., Teng, D., Chowdhury, R.~R., Li, S., Hong, D., Gupta, R.~K., and Shang, J.
\newblock Unimts: Unified pre-training for motion time series, 2024{\natexlab{b}}.
\newblock URL \url{https://arxiv.org/abs/2410.19818}.

\end{thebibliography}
\bibliographystyle{nips}

\clearpage
\appendix
\section{Datasets}
\label{appx:dataset}
\begin{figure}[hbt]
    \centering
    \includegraphics[width=1\linewidth,trim=0 500 0 0, clip]{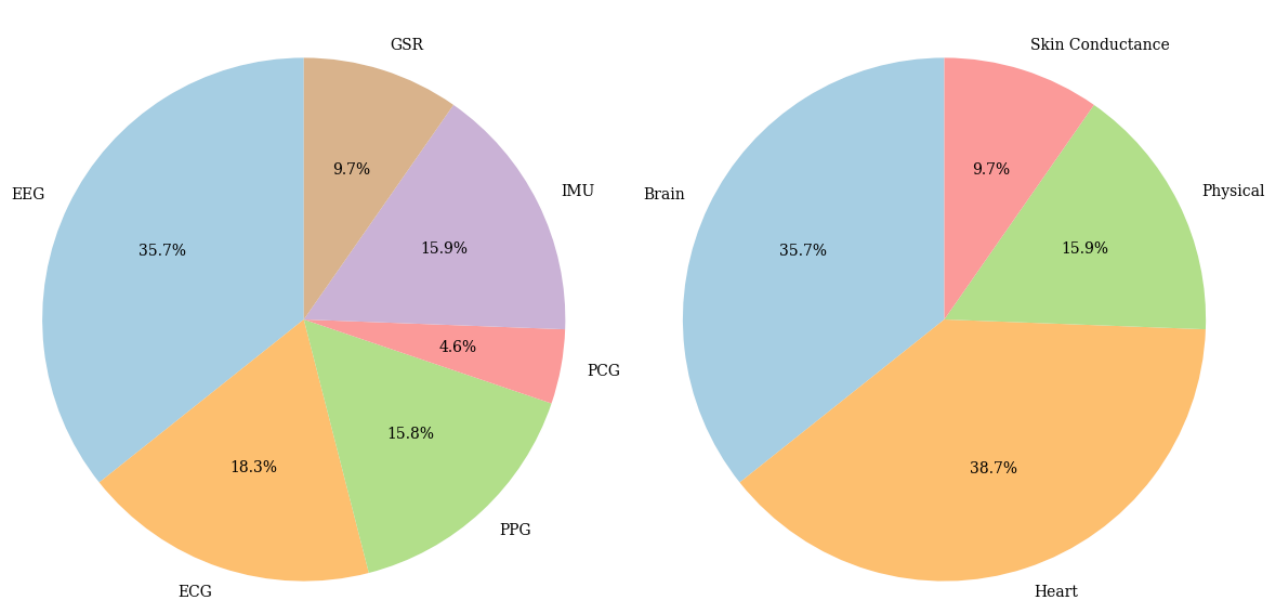}
    \caption{\textbf{Distribution of sensor signals used for pretraining.} \textit{Left:} Distribution by sensor modality. \textit{Right:} Distribution by type of physiological information.}
    \label{fig:pretrain-dist}
    \vspace{-2mm}
\end{figure}

Few openly accessible multi-channel or multi-device datasets for physiological signals exist, limiting advancements in this field. To address this gap, we curated a dataset comprising approximately 385 hours of recordings. Using the augmentation algorithm described below, we expanded this dataset to 4294 hours. The distribution of the pretraining dataset, as shown in Figure \ref{fig:pretrain-dist}, reflects the inherent diversity of the original recordings, ensuring balanced representation across channels and devices. This curated and augmented dataset provides a critical resource for developing robust models, facilitating progress in multi-channel physiological signal research.
\begin{table*}[!ht]
\vspace{-1mm}
    \begin{minipage}{.5\linewidth}
    \caption{\textbf{Downstream evaluation data that are unseen during pretraining.}}
      \scriptsize
      \centering
      \resizebox{1.0\textwidth}{!}{
        \begin{tabular}{|l|l|l|l|l|}
        \hlineB{2}
        \hlineB{2}
        \bf \begin{tabular}{@{\hspace{-0.5\tabcolsep}}l@{\hspace{-0.5\tabcolsep}}}
            Downstream
            Dataset \\ 
        \end{tabular} &
        \bf 
        \begin{tabular}{@{\hspace{-0.5\tabcolsep}}l@{\hspace{-0.5\tabcolsep}}}
            Sensor \\
        \end{tabular}
       &
        \bf \# Channels  
       &
        \bf \begin{tabular}{@{}l@{}}
            Tasks \\ 
        \end{tabular} 
       &
        \bf \begin{tabular}{@{}l@{}}
            \#Samp.
            (\#Subj.) \\ 
        \end{tabular}
      
        \\\hline
        \hlineB{2}
        \cellcolor{blue!20} \begin{tabular}{@{\hspace{-0.5\tabcolsep}}l@{\hspace{-0.5\tabcolsep}}} WESAD \\
            \citep{wesad} \\ 
        \end{tabular}&
        \begin{tabular}{@{\hspace{-0.5\tabcolsep}}l@{\hspace{-0.5\tabcolsep}}}
            IMU, PPG, \\
            ECG, GSR \\ 
        \end{tabular}&
        10&
        \begin{tabular}{@{}l@{}}
            Stress \\ 
            Detection \\
        \end{tabular}&
        11050(15) 
        \\\hline
        \cellcolor{blue!20} \begin{tabular}{@{\hspace{-0.5\tabcolsep}}l@{\hspace{-0.5\tabcolsep}}}UCI-HAR \\
            \citep{uci-har} \\ 
        \end{tabular}&
        \begin{tabular}{@{}l@{}}
            IMU \\ 
        \end{tabular}&
        6
       &
        \begin{tabular}{@{}l@{}}
            HAR \\ 
        \end{tabular}&
        10299(30) 
        \\\hline
        \cellcolor{blue!20} \begin{tabular}{@{\hspace{-0.5\tabcolsep}}l@{\hspace{-0.5\tabcolsep}}}DriverFatigue \\
            \citep{driver-fatigue} \\ 
        \end{tabular}&
        \begin{tabular}{@{}l@{}}
            EEG \\ 
        \end{tabular}&
        4
       &
        \begin{tabular}{@{}l@{}}
            Fatigue \\
            Detection \\
        \end{tabular}&
        2400(12) 
        \\\hline
        \bf Activity Recognition Total&-&-&-&\bf 23749(57)
        \\\hline
        \hlineB{2}
        \hlineB{2}
        \cellcolor{teal!20} \begin{tabular}{@{}l@{}}Epilepsy\\
            \citep{epilepsy} \\ 
        \end{tabular}&
        \begin{tabular}{@{}l@{}}
            EEG \\ 
        \end{tabular}&
        1
       &
        \begin{tabular}{@{}l@{}}
            State \\
            Recognize \\
        \end{tabular}&
        11500(500) 
        \\\hline
        \cellcolor{teal!20} \begin{tabular}{@{}l@{}}GAMEEMO \\
            \citep{gameemo} \\ 
        \end{tabular}&
        \begin{tabular}{@{}l@{}}
            EEG \\ 
        \end{tabular}&
        4
       &
        \begin{tabular}{@{}l@{}}
            Valence- \\
            Arousal \\ 
        \end{tabular}&
        5600(28) 
        \\\hline
        \bf EEG Main Tasks Total&-&-&-&\bf 17100(528)
        \\\hline
        \hlineB{2}
        \hlineB{2}
        \cellcolor{red!20} \begin{tabular}{@{}l@{}}ECG-Abnormal\\
            \citep{ecg-abnormal} \\ 
        \end{tabular}&
        \begin{tabular}{@{}l@{}}
            ECG \\ 
        \end{tabular}&
        1
       &
        \begin{tabular}{@{}l@{}}
            Abnormal \\
            Detection \\
        \end{tabular}&
        11640(249) 
        \\\hline
         \cellcolor{red!20} \begin{tabular}{@{}l@{}}PPG-BP\\
            \citep{ppg-china} \\ 
        \end{tabular}&
        \begin{tabular}{@{}l@{}}
            PPG \\ 
        \end{tabular}&
        1
       &
        \begin{tabular}{@{}l@{}}
            Risk of \\
            Diseases \\
        \end{tabular}&
        657(219) 
        \\\hline
         \cellcolor{red!20} \begin{tabular}{@{}l@{}}PhysioNet EMG\\
            \citep{PhysioNet} \\ 
        \end{tabular}&
        \begin{tabular}{@{}l@{}}
            EMG \\ 
        \end{tabular}&
        1
       &
        \begin{tabular}{@{}l@{}}
            Muscular \\
            Diseases \\
        \end{tabular}&
        163(3) 
        \\\hline
        \bf Risk Evaluation Total&-&-&-&\bf 12460(471)
        \\\hline
        \hlineB{2}
        \hlineB{2}
        \cellcolor{orange!10} \begin{tabular}{@{}l@{}}Noninvasive-BP\\
            \citep{ppg-bp} \\ 
        \end{tabular}&
        \begin{tabular}{@{}l@{}}
            PPG \\ 
        \end{tabular}&
        3
       &
        \begin{tabular}{@{}l@{}}
            BP \\
            Estimate \\
        \end{tabular}&
        125(26) 
        \\\hline
        \cellcolor{orange!10} \begin{tabular}{@{}l@{}}PPG-Hgb\\
            \citep{ppg-bp} \\ 
        \end{tabular}&
        \begin{tabular}{@{}l@{}}
            PPG \\ 
        \end{tabular}&
        2
       &
        \begin{tabular}{@{}l@{}}
            Hgb \\
            Estimate \\
        \end{tabular}&
        68(68) 
        \\\hline
        \cellcolor{orange!10} \begin{tabular}{@{}l@{}}Fetal-fPCG\\
            \citep{india-fpcg} \\ 
        \end{tabular}&
        \begin{tabular}{@{}l@{}}
            PCG \\ 
        \end{tabular}&
        1
       &
        \begin{tabular}{@{}l@{}}
            Fetal HR \\
            Estimate \\
        \end{tabular}&
        47(47)
        \\\hline
        \bf Vital Signs Total&-&-&-&\bf 240(141)
        \\\hline
        \bf Total All&-&-&-&\bf 53549(1197)
        \\\hline
        \hlineB{2}
        \end{tabular}
    } 
    \label{tab:downstream-info}
    \end{minipage}
%
%
    \begin{minipage}{.52\linewidth}
    \caption{\textbf{Pretraining data.}}
    \vspace{3mm}
    \scriptsize
    \centering
        %
    \resizebox{0.78\textwidth}{!}{
    \hspace{-10mm}
    \begin{tabular}{|l|l|l|}
        \hlineB{2}
        \hlineB{2}
        \bf Pretrain Dataset&
        \bf 
        Sensors
       &
        \bf \#Samp (hours).
        \\\hline
        \hlineB{2}
        \begin{tabular}{@{}l@{}}Cuff-Less-BP \\
            \citep{cuff-less} \\ 
        \end{tabular}&
        \begin{tabular}{@{}l@{}}
            ECG, PPG\\ 
        \end{tabular}&
        42934(72)
        \\\hline
        \begin{tabular}{@{}l@{}}PPG-Dalia \\
            \citep{ppg-dalia} \\ 
        \end{tabular}&
        \begin{tabular}{@{}l@{}}
            ECG, PPG\\ 
            IMU, GSR\\
        \end{tabular}&
        42606(71)
        \\\hline
        \begin{tabular}{@{}l@{}}Auditory-EEG \\
            \citep{audi-eeg} \\ 
        \end{tabular}&
        \begin{tabular}{@{}l@{}}
            EEG\\
        \end{tabular}&
        13601(23)
        \\\hline
        \begin{tabular}{@{}l@{}}PhyAAt \\
            \citep{phyaat} \\ 
        \end{tabular}&
        \begin{tabular}{@{}l@{}}
            EEG\\
        \end{tabular}&
        19550(33)
        \\\hline
        \begin{tabular}{@{}l@{}}MAUS \\
            \citep{maus} \\ 
        \end{tabular}&
        \begin{tabular}{@{}l@{}}
            ECG, PPG\\
            GSR\\
        \end{tabular}&
        13068(22)
        \\\hline
        \begin{tabular}{@{}l@{}}Mendeley-YAAD \\
            \citep{yaad} \\ 
        \end{tabular}&
        \begin{tabular}{@{}l@{}}
            ECG, GSR\\
        \end{tabular}&
        2964(5)
        \\\hline
        \begin{tabular}{@{}l@{}}Brain-Cognitive \\
            \citep{brain_cog} \\ 
        \end{tabular}&
        \begin{tabular}{@{}l@{}}
            EEG\\
        \end{tabular}&
        51201(85)
        \\\hline
        \begin{tabular}{@{}l@{}}EPHNOGRAM \\
            \citep{EPHNOGRAM} \\ 
        \end{tabular}&
        \begin{tabular}{@{}l@{}}
            ECG, PCG\\
        \end{tabular}&
        36611(61)
        \\\hline
        \begin{tabular}{@{}l@{}}BIDMC \\
            \citep{bidmc} \\ 
        \end{tabular}&
        \begin{tabular}{@{}l@{}}
            ECG, PPG\\
        \end{tabular}&
        8427(14)
        \\\hline
        \hlineB{2}
        \bf Num Segments (\# Segm.)&-&\begin{tabular}{@{}l@{}}230,962(385) \\
        \end{tabular}
        \\
        \cellcolor{gray!30}\bf \# Segm. w/ Augment&\cellcolor{gray!30}-& \cellcolor{gray!30}\begin{tabular}{@{}l@{}}2,576,418(4,294) \\
        \end{tabular}
        \\\hline
        \bf Num Sensor Signals (\# Sign.)&-&\bf \begin{tabular}{@{}l@{}}802,019(1,337) \\
        \end{tabular}
        \\
        \cellcolor{gray!30}\bf \# Sign. w/ Augment&\cellcolor{gray!30}-&\cellcolor{gray!30}\bf \begin{tabular}{@{}l@{}}8,965,538(14,943) \\
        \end{tabular}
        \\\hline
        \hlineB{2}
        \end{tabular}
    } 
    \label{tab:pretrain-info}
    \end{minipage} 
\vspace{-3mm}
\end{table*}

Table \ref{tab:downstream-info} overviews used dataset in our experiement along with the modality and task type. We will gives further details for each dataset below:

\textbf{WESAD} \citep{wesad} is a publicly available multimodal dataset used for wearable stress and affect detection, formulated as a classification task with labels: neutral, stress, and amusement. The dataset includes physiological and motion data collected from 15 subjects during a lab study, using a chest-worn RespiBAN device and a wrist-worn Empatica E4 device. From the chest device, we use electrocardiogram (ECG), galvanic skin response (GSR), and triaxial acceleration (ACC-X, ACC-Y, ACC-Z), all sampled at 700 Hz. From the wrist device, we use photoplethysmogram (PPG), galvanic skin response (GSR, 4 Hz), and triaxial acceleration (ACC-X, ACC-Y, ACC-Z, 32 Hz). The selected channels span multiple physiological and motion modalities from both chest and wrist sensors. Each data segment is labeled with one of the three affective states, serving as the target output for classification tasks.

\textbf{UCI-HAR} \citep{uci-har} dataset is publicly available and is used for classifying human activities based on sensor data. It comprises data from 30 volunteers, aged 19 to 48, each performing six activities: walking, walking upstairs, walking downstairs, sitting, standing, and laying. During these activities, participants carried a waist-mounted smartphone equipped with embedded accelerometer and gyroscope sensors. The input channels consist of triaxial linear acceleration and triaxial angular velocity, totaling six channels. Each data segment is labeled with one of the six activities, serving as the target output for classification tasks. The sensors recorded data at a constant rate of 50 Hz.

\textbf{Driver Fatigue EEG Dataset} \citep{driver-fatigue} is a publicly available dataset used for detecting driver fatigue based on electroencephalogram (EEG) signals. EEG data were collected using a 40-channel Neuroscan amplifier. The recordings include EEG data corresponding to two states: alert and fatigued. Each data segment is labeled with one of these states, serving as the target output for classification tasks.

\textbf{Epileptic Seizure Recognition} \citep{epilepsy} dataset is publicly available and is used for classifying neurological and physiological states based on EEG signals. It comprises data from 500 subjects, each recorded for 23.6 seconds using a single EEG channel at a sampling rate of 178 Hz. Each sample is labeled with one of five brain states, allowing for the construction of multiple binary classification tasks that target different aspects of neurological assessment. Specifically, we formulated five tasks: 
\begin{itemize}
    \item \textit{Eye Relaxation}: Detects eye fatigue by distinguishing between relaxed and alert states based on EEG changes related to eye closure.
    \item \textit{Health Area}: Classifies brain regions as healthy or affected by neurological abnormalities.
    \item \textit{Tumor Area}: Detects EEG patterns indicative of tumor presence in specific brain regions.
    \item \textit{Seizure}: Identifies seizure activity from non-seizure states.
    \item \textit{Eyes Open vs. Closed}: Differentiates EEG signals associated with visual input states.
\end{itemize}

\textbf{GAMEEMO} \citep{gameemo} is a publicly available dataset used for emotion recognition based on EEG signals. It comprises data from 28 subjects, each playing four emotion-inducing computer games (boring, calm, horror, and funny) for five minutes per game, totaling 20 minutes of EEG data per subject. EEG signals were recorded using the EMOTIV EPOC+ headset, which includes 14 channels (AF3, AF4, F3, F4, F7, F8, FC5, FC6, O1, O2, P7, P8, T7, and T8) positioned according to the 10–20 system. The signals were sampled at 128 Hz. After each gameplay session, subjects rated their emotional response using the Self-Assessment Manikin (SAM) form, providing continuous scores for arousal and valence. These scores were quantized into binary values using subject-specific median thresholds: arousal and valence ratings above the median were labeled as high, and those below or equal to the median as low. Combining the binarized arousal and valence ratings yields four discrete emotional classes: low arousal and low valence, low arousal and high valence, high arousal and low valence, and high arousal and high valence. Each data segment is labeled with one of these four classes, serving as the target output for four-class emotion classification tasks.

\textbf{ECG Heartbeat Categorization} \citep{ecg-abnormal} is a publicly available dataset used for classifying heartbeat signals based on electrocardiogram (ECG) recordings. It comprises two collections of heartbeat signals derived from PhysioNet's MIT-BIH Arrhythmia Dataset and the PTB Diagnostic ECG Database. The first collection includes 109,446 samples categorized into five classes: normal (N), supraventricular ectopic (S), ventricular ectopic (V), fusion (F), and unknown (Q), with ECG signals sampled at 125 Hz. The second collection consists of 14,552 samples categorized into two classes: normal and abnormal, also sampled at 125 Hz. For our analysis, we restructured the dataset into a binary classification framework by consolidating the original categories into two classes: normal and abnormal heartbeats.

\textbf{PPG-China} \citep{ppg-china} is a publicly available dataset used for classifying cardiovascular and metabolic conditions based on photoplethysmography (PPG) signals. It comprises 657 data records from 219 subjects, aged 20 to 89 years, including individuals with conditions such as hypertension and diabetes. PPG signals were recorded using a single channel at a sampling rate of 125 Hz. Each subject's data includes PPG waveforms and corresponding clinical information, facilitating the construction of multiple classification tasks focused on cardiovascular and systemic health monitoring. Specifically, we formulated four tasks:
\begin{itemize}
    \item \textit{PPG-HTN}: Identifies stages of hypotension severity by classifying PPG signals into four levels.
    \item \textit{PPG-DM}: Detects diabetes by distinguishing between diabetic and non-diabetic individuals.
    \item \textit{PPG-CVA}: Identifies the presence or absence of cerebrovascular accidents (strokes) based on PPG patterns.
    \item \textit{PPG-CVD}: Assesses cardiovascular disease by classifying PPG signals into three cardiovascular health categories.
\end{itemize}

\textbf{PhysioNetEMG} \citep{PhysioNet} is a publicly available dataset used for classifying neuromuscular conditions based on electromyography (EMG) signals. It comprises single-channel EMG recordings from the tibialis anterior muscle of three subjects: one healthy, one with neuropathy, and one with myopathy. The EMG signals were recorded at a sampling rate of 4,000 Hz. Each recording was segmented into time series samples using a fixed-length window of 6 second. Each segment is labeled according to the subject's condition—healthy, neuropathy, or myopathy—serving as the target output for classification tasks.

\textbf{Non-invasive Blood Pressure Estimation} \citep{ppg-bp} is a publicly available dataset used for cuff-less blood pressure (BP) estimation. It comprises data from 26 subjects, each with recorded electrocardiogram (ECG) and photoplethysmogram (PPG) signals, sampled at 1,000 Hz. Reference BP measurements were taken during signal acquisition. Each subject's data also includes demographic information such as age, weight, and height. The dataset is structured to facilitate regression tasks aimed at predicting systolic and diastolic BP values.

\textbf{PPG-HGB} \citep{ppg-hgb} is a publicly available dataset used for non-invasive hemoglobin (Hb) measurement based on photoplethysmography (PPG) signals. It comprises data from 68 subjects, aged 18 to 65 years, with a gender distribution of 56\% female and 44\% male. PPG signals were recorded using the MAX30102 sensor, which emits red and infrared light. The sensor's analog-to-digital converter (ADC) output data rate can be programmed from 50 samples per second (sps) to 3200 sps. Each subject contributed 12 sets of PPG signals, totaling 816 data records. We formulate regression tasks aimed at predicting Hb concertration levels. 

\textbf{Fetal-fPCG} \citep{india-fpcg} is a publicly available dataset designed for estimating fetal heart rate (FHR) using fetal phonocardiography (fPCG) signals. It includes recordings from 60 pregnant women, aged 18 to 37 years, with gestational ages between 31 and 40 weeks. The recordings were collected at St. John’s Hospital in Bangalore using an electronic stethoscope (SS30LA) connected to a Biopac MP36 data acquisition system. The stethoscope was placed on the lower abdomen of each subject to capture the fPCG signal, which was sampled at 2,000 Hz. The dataset supports regression tasks, where the goal is to predict continuous FHR values directly from the fPCG waveforms.

\section{Implementation Detail}
\label{appx:hyper}
\subsection{Data Preprocess.} For the data preparation, we set the uniform sampling rate and interval length to 65 HZ and 6 seconds respectively. In our case, 65 Hz covers most of the frequency bands of interest such as heart activity, physical motions, and neuron activity up to the beginning of Gamma power (above 30 Hz). And a great amount of samples are less than 6 seconds such as \citep{uci-har,ppg-china,ecg-abnormal}. \edit{We conduct basic pre-processing for each signal with identical setting: (1) de-trended \edit{by subtract the result of a linear least-squares fit to series data from the raw time series,} and (2) Gaussian smoothed with standard deviation of 1.3 (0.02 seconds), ensuring a highly consistent dataset for training.}

\edit{Since the Transformer's computational requirements scale quadratically with input length, to release the full potential of our self-supervised algorithm, we segment our multivariate time series into intervals with a uniform length and pad shorter samples with zeros. This approach not only enables parallel processing of samples in large minibatches but also addresses variation in the length of individual samples.}

For the downstream task, we split the data into train and test sets for linear probing evaluation with portion of $80\%$ and $20\%$ correspondingly. The split is stratified on the anonymized subject ID if this information is provided by the dataset.

\subsection{Data Augmentation.} Since there are very few publicly available datasets containing multiple devices or modalities, we aim to expand our curated training set to fully leverage the potential of self-supervised learning. Inspired by data augmentation techniques in computer vision and natural language processing \citep{zhang2017mixup,data_aug}, we adopt a heuristic approach to augment the dataset. Specifically, we augment each sub-dataset by a factor of 10. For each dataset, we sample two time series, randomly extract a segment from one, and substitute it with a transformed counterpart, as outlined in the pseudocode in Algorithm \ref{alg:augment_single_series}. As a result, our training set is expanded to 2,586,404 segments, corresponding to 4,294 hours of data.
\begin{algorithm}[htbp!]
\caption{Time Series Mixup Augmentation}
\label{alg:augment_single_series}
\textbf{Input:} Time series dataset $\mathcal{X}$, number of augmentations $n$ \\
\textbf{Output:} Augmented Dataset $\Tilde{\mathcal{X}}$
\begin{algorithmic}[1]
\FOR{$i = 1$ to $n$}
    \STATE Sample two time series $\mathbf{x}^{(1)}, \mathbf{x}^{(2)} \sim \mathcal{X}$
    \STATE Sample a chunk size $\lambda \sim \mathcal{U}(0, l)$
    \STATE Sample start indices $s_1, s_2 \sim \mathcal{U}(0, l - \lambda)$
    \STATE Swap chunk from $\mathbf{x}^{(2)}$ into $\mathbf{x}^{(1)}$: 
    \[
    \mathbf{x}^{(1)}_{s_1:s_1+\lambda} \gets \mathbf{x}^{(2)}_{s_2:s_2+\lambda}
    \]
    \STATE Append $\mathbf{x}^{(1)}$ into $\Tilde{\mathcal{X}}$
\ENDFOR
\STATE \textbf{return} $\Tilde{\mathcal{X}}$
\end{algorithmic}
\end{algorithm}

\vspace{4mm}
\subsection{Pretraining Framework.} 
\begin{figure*}[t!]
\vspace{-2mm}
\centering
\includegraphics[width=0.95\textwidth]{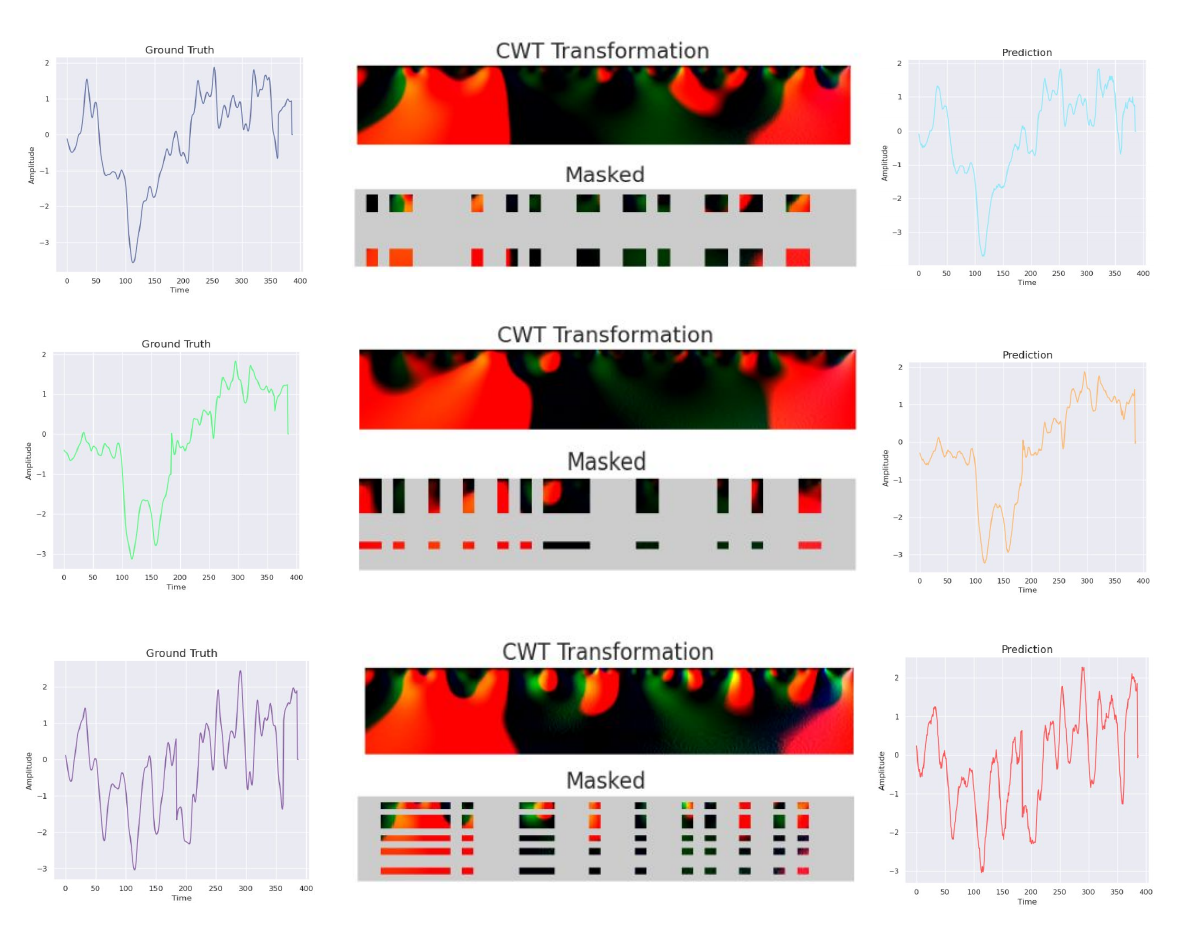}
\vspace{-4mm}
\caption{\textbf{Visualization of original time series (left), CWT transformation image with structured masking (middle), and reconstructed time series (right).}}
\label{fig:pretrain_demo}
\vspace{-4mm}
\end{figure*}

Normwear is derived from the Masked Autoencoder (MAE) \citep{mae}. The detailed hyper-parameter choice is descibe in \ref{tab:model-hyper}. We use a Conv2D layer with a kernel size of (9, 5) and a stride of (9, 5), ensuring no overlapping patches. This layer takes input with 3 channels and projects it to 768 channels, matching the hidden size of our encoders. In Normwear, we apply structured masking independently to each variate along both the frequency and time axes, with respective masking ratios of 0.6 and 0.5. This results in an expected overall masking ratio of 0.8 for each variate. Only the unmasked tokens are passed to the encoder, reducing computational complexity. To enhance representation learning, we introduce six additional transformer blocks as fusion layers, interleaved with the original 12 encoder blocks, creating a total of 18 blocks. Each transformer block has a hidden dimension of 768 and uses LayerNorm as in the original MAE. The latent embeddings obtained from the encoder are projected from 768 to 512 dimensions. Learnable masked tokens are reinserted at their original positions, and positional embeddings are added to guide the decoder in reconstructing the input series. The lightweight decoder consists of two transformer blocks with a hidden dimension of 512, followed by two Conv1D layers. The first Conv1D layer maps from the flattened multivariate signal embedding to an intermediate dimension, and the second Conv1D layer maps from this intermediate dimension back to the original multivariate signal space. A GELU activation function is used between these layers, with BatchNorm applied to the input. The decoder reconstructs the original input series, and the model is trained using Mean Squared Error (MSE) loss on all data points. Our models are pre-trained for 45,000 steps with a batch size of 256, using the AdamW optimizer with a learning rate of $10^{-4}$. We did not perform on-the-fly data augmentation, as suggested in the MAE framework, due to the high masking ratio. (An end-to-end example of the input and output of this pretraining pipeline is illustrated in Fig. \ref{fig:pretrain_demo})

\edit{All the models are pretrained on 4 NVIDIA RTX 3090 graphical computing unit (GPU), with 24GB of GPU memory on each card. }

\subsection{MSiTF.} For pretraining the representation alignment module, we have the training hyper-parameters in Table \ref{tab:zero-shot-hyper}.
\begin{table*}[ht]
    \centering
    \renewcommand{\arraystretch}{1.1} 
    \setlength{\tabcolsep}{4pt} 
    \begin{minipage}{.65\textwidth} 
        \centering
        \caption{\textbf{NormWear Pretraining Hyper-parameters.}}
        \begin{tabular}{|l|r|}
        \hlineB{2}
        \cellcolor{gray!20} \bf Hyper-parameter & \cellcolor{gray!20} \bf Value \\
        \hline
        \# cross-patches Transformer Encoder & 12 \\
        \# cross-channels Transformer Encoder & 6 \\
        \# Transformer Decoder & 2 \\
        \# Attention Heads & 12 \\
        Encoder Latent Size & 768 \\
        Decoder Latent Size & 512 \\
        Feedforward Latent Size & 3072 \\
        Normalization & LayerNorm \\
        Patch size (time axis) & 9 \\
        Patch size (scale axis) & 5 \\
        \hlineB{2}
        Optimizer & AdamW \\
        Loss Scalar & NativeScaler \\
        Base Learning Rate (blr) & 1e-3 \\
        Epochs & 140 \\
        Batch size & 192 \\
        \hlineB{2}
        \end{tabular}
        \label{tab:model-hyper}
    \end{minipage}%
    \hfill 
    \begin{minipage}{.35\textwidth} 
        \centering
        \caption{\textbf{MSiTF Hyper-parameter}}
        \begin{tabular}{|l|r|}
        \hlineB{2}
        \cellcolor{gray!20} \bf Hyper-parameter & \cellcolor{gray!20} \bf Value \\
        \hline
        Learning rate (lr) & 1e-3 \\
        Epochs & 40 \\
        Batch size & 32 \\
        L2 regularization & 5e-6 \\
        lr decay rate & 0.997 \\
        $\lambda$ & 0.5 \\
        $\tau$ & 0.5 \\
        \hlineB{2}
        \end{tabular}
        \label{tab:zero-shot-hyper}
    \end{minipage}
    
    \vspace{-4mm}
    \end{table*}
\newpage
\subsection{Aligner Module, Objective Function, and Pretraining.} \label{app:ctl_loss}
The Aligner Module matches two vectors: the fused representation $f(q, H) = \hat{Y} \in \mathbb{R}^E$ with the semantic embedding ($Y$) of ground truth sentence, which is obtained from prompting the ground truth label using a template, for example, \textit{``The subject is presently \{activity\_label\}"}. In the same manner as the query embedding, the ground truth sentence is encoded using the same pre-trained language model \citep{clinical-tinyllama}.
\edit{At this stage, $Y$ is leveraged to supervise the fused output $\hat{Y}$. The vanilla contrastive learning loss formula following \cite{unimts} is:
\begin{equation}
\label{eq:ctl-loss}
Loss_{ctl}(Y, \hat{Y}) = -\frac{1}{N}\sum_{i=1}^N \log \frac{\exp (\hat{Y_i}^T Y_i)^{\frac{1}{\gamma}}}{\sum_{k=1}^N\exp (\hat{Y_i}^T Y_k)^{\frac{1}{\gamma}}}
\end{equation}
where $N$ is the batch size and $\gamma$ is the learnable temperature parameter. We denote this loss function as contrastive loss with batch normalizer. We also leverage a refine process after contrastive learning using simlarity loss with per sample normalizer, which is essentially cosine similarity loss, with vector distance as supplemental penalty:
\vspace{-2mm}
\begin{equation}
\label{eq:semantic-loss}
Loss_{refine}(\hat{Y}, Y) = \frac{1}{N} \sum_{i=1}^N\left(\left(1-\frac{\hat{Y}_i^T Y_i}{\|\hat{Y}_i\| \|Y_i\|}\right) + \lambda |\hat{Y_i}-Y_i|\right)
\end{equation}
where $\lambda$ is hyper-parameters controlling the weight of the supplemental loss components. 
}

\subsection{Sentence template example for signal-sext alignment.} \label{appx:sent_example}
To enhance the expressiveness and diversity of supervision signals for our MSiTF alignment module, we convert categorical labels into natural language descriptions using varied prompt templates. We apply this strategy to several pretraining tasks. We present example sentence templates below for emotion recognition and activity recognition to demonstrate the general idea of how we derive text modality from the raw label:

For the \textbf{emotion recognition} task, we use:
\begin{tcolorbox}[promptbox]
    \begin{itemize}
        \item `` The emotion detected is \{\}.''
        \item `` This subject is feeling \{\}.''
        \item `` The emotional state is \{\}.''
        \item `` The identified emotion is \{\}.''
    \end{itemize}
\end{tcolorbox}

For the \textbf{activity recognition} task, we use:
\begin{tcolorbox}[promptbox]
    \begin{itemize}
        \item `` This subject is currently \{\}.''
        \item `` The subject is engaged in \{\}.''
        \item `` Current activity is \{\}.''
        \item `` Subject's activity is \{\}.''
    \end{itemize}
\end{tcolorbox}

By exposing the model to multiple phrasings for the same label, this design helps it learn modality-invariant representations that are more robust to linguistic variation and better aligned across modalities. 
\edit{Specifically, to increase the diversity of semantic representations of query and ground truth sentences 
in the pretraining signal corpus, we utilize large language models (GPT-3.5) \citep{chatgpt} to generate 20 alternative variations for each sentence, from which only one is randomly sampled during pre-training. 
During test-time inference on downstream datasets, each ground truth label is converted into a sentence (details in appendix  \ref{appx:sent_example}), which is transformed into a semantic embedding using a frozen text encoder. The sentence with the closest distance with the embedding from our foundation model is used as the final inferential result. }

\subsection{Statistical Feature List.} Our statistical baseline includes features extracted from both the time and frequency domains. In the time domain, we compute the mean, standard deviation, maximum, minimum, skewness, kurtosis, 25\% quantile, median, and 75\% quantile. In the frequency domain, we extract the spectral centroid, spectral spread, mean frequency, peak frequency, as well as the 25\%, 50\% (median), and 75\% quantile frequencies.

\subsection{Radar Plot or Performance Trend.} To enhance the visual contrast between model performances across tasks, we applied the Softmax function to the raw performance scores. This transformation rescales the scores to a range between 0 and 1, accentuating relative differences between models. While the Softmax transformation emphasizes the relative improvement of our model over others, we note that the absolute scores may differ from those originally reported.

\section{Ablation Study} \label{appx:ablation}
\edit{Due to computational constraints, we will conduct the ablation study on our smaller dataset (37k samples) to train and evaluate the model, establishing a proof of concept and demonstrating the effectiveness of our approach in a controlled setting.}

\textbf{Fusion Schemes.} Table \ref{tab:fusion-scheme-res-1} shows the performance of different fusion schemes, including (1) no fusion, (2) cross-attention fusion, (3) \textsc[CLS]-attention fusion, and (4) mean-pooling fusion. We excluded all-attention fusion in our ablation study because it is computationally prohibitible. Among all the compared strategies, the \textsc[CLS] token fusion generally achieves the best accuracy with a minor increase in parameters.
\begin{table*}[!htbp]
\vspace{-2mm}
\centering
\caption{
\textbf{Performance Comparison of Various Fusion Schemes}
}
\vspace{1mm}
\resizebox{\textwidth}{!}{
        \begin{tabular}{|l|c|c|c|c|}
        \hlineB{2}
        \hlineB{2}
        \bf \begin{tabular}{@{\hspace{-0.5\tabcolsep}}l@{\hspace{-0.5\tabcolsep}}}
            Downstream
            Tasks \\ 
        \end{tabular}  & 
        \bf \begin{tabular}{@{}c@{}}No fusion \end{tabular}
        & 
        \bf \begin{tabular}{@{}c@{}} Cross-Attention fusion\end{tabular}
        & 
        \bf \begin{tabular}{@{}c@{}}Mean pooling fusion\end{tabular} 
        &
        \bf \begin{tabular}{@{\hspace{-0.5\tabcolsep}}l@{\hspace{-0.5\tabcolsep}}}
            [CLS] Token fusion \\
        \end{tabular}  
        \\\hline
        \hlineB{2}
        \cellcolor{blue!20} WESAD & 
        72.209&	74.165&	71.99 & \bf 75.390
        \\\hline
        \cellcolor{blue!20} UCI-HAR & 
        97.793&	96.908&	97.566 & \bf 98.928
        \\\hline
        \cellcolor{blue!20} DriverFatigue & 
        73.252&	60.308&	72.552&	\bf 75.167
        \\\hline
        \hlineB{2}
        \bf Activity Recognition Avg. &
        81.085&	77.127&	80.703& \cellcolor{gray!40}\bf \textbf{\textbf{83.162}}
        \\\hline
        \hlineB{2}
        \hlineB{2}
        \cellcolor{teal!20} Epilepsy (eye open state) &
        90.966&	84.075&	89.817&	\bf 92.203
        \\\hline
        \cellcolor{teal!20} Epilepsy (eye relaxation) &
        94.399&	93.589&	93.912&	\bf 94.908
        \\\hline
        \cellcolor{teal!20} Epilepsy (health area) &
        87.866&	86.899&	87.248&	\bf 88.117
        \\\hline
        \cellcolor{teal!20} Epilepsy (tumor area) &
        86.599&	86.861&	\bf 87.152& 86.888
        \\\hline
        \cellcolor{teal!20} Epilepsy (seizure) &
        \bf 97.477&	96.351&	96.719&	96.638
        \\\hline
        \cellcolor{teal!20} GAMEEMO &
        \bf 57.695&	56.724&	58.079& 56.532
        \\\hline
        \hlineB{2}
        \bf EEG Main Tasks Avg. & 
        85.834&	84.083&	85.488	& \cellcolor{gray!60}\bf \textbf{85.881}
        \\\hline
        \hlineB{2}
        \hlineB{2}
        \cellcolor{red!20} ECG-Abnormal &
        99.429&	\bf 99.441&	99.268	& 99.041
        \\\hline
         \cellcolor{red!20} PPG-BP (HTN) &
         61.850&	60.983&	\bf 63.577	&	60.344
        \\\hline
        \cellcolor{red!20} PPG-BP (DM) &
         58.333&	\bf 62.800&	62.200	& 59.459
        \\\hline
        \cellcolor{red!20} PPG-BP (CVA) &
        61.319&	61.458&	59.236	& \bf 70.278
        \\\hline
        \cellcolor{red!20} PPG-BP (CVD) &
        48.417&	\bf 53.585&	46.961	& 52.596
        \\\hline
         \cellcolor{red!20} PhysioNet EMG &
        93.715&	95.49&	86.749	& \bf 98.184
        \\\hline
        \hlineB{2}
        \bf Risk Evaluation Avg. & 
        70.511&	72.293&	69.665& \cellcolor{gray!60}\bf73.317
        \\\hline
        \hlineB{2}
        \hlineB{2}
        \cellcolor{orange!10}Noninvasive-BP &
        88.356&	\bf 92.759&	88.719	& 92.470
        \\\hline
        \cellcolor{orange!10} PPG-Hgb &
        95.031&	93.413&	\bf 95.086	&	94.766
        \\\hline
        \cellcolor{orange!10} Fetal-fPCG &
        98.582&	\bf 99.145&	98.771	&	99.088
        \\\hline
        \hlineB{2}
        \bf Vital Signs Avg. & 
        93.990&	95.106&	94.192	& \cellcolor{gray!60}\bf \textbf{95.441}
        \\\hline
        \hlineB{2}
        \bf Micro Avg. & 
        81.294&	80.831&	80.867&	\cellcolor{gray!60}\bf82.833
        \\\hline
        \bf Macro Avg. & 
        82.855&	82.152&	82.512&	\cellcolor{gray!60}\bf\textbf{84.450}
        \\\hline
        \hlineB{2}
        \end{tabular}
}
\label{tab:fusion-scheme-res-1}
\end{table*}

\newpage
\textbf{Masking Strategies in Pre-training.} We ablated our masking strategy introduced in Section \ref{sec2.3}. Using a consistent mask ratio of 0.8 in all strategies, we found that applying masking along the scale and time axes produced the best performance (details in Table \ref{tab:mask-scheme-res-2}).
\begin{table*}[ht]
\vspace{-2mm}
\scriptsize
\centering
\caption{
Performance Comparison of Different Masking Strategies
}
\vspace{1mm}
\resizebox{\textwidth}{!}{
    \begin{tabular}{|l|c|c|c|c|}
        \hlineB{2}
        \bf Downstream Tasks &
        \bf \begin{tabular}{@{}c@{}}Unstructured Mask\\ ($P = 0.8$)\end{tabular} &
        \bf \begin{tabular}{@{}c@{}}Time Mask \\ ($P_t=0.8, P_f=0.0$)\end{tabular} &
        \bf \begin{tabular}{@{}c@{}}Scale Mask \\ ($P_t=0.0, P_f=0.8$)\end{tabular} &
        \bf \begin{tabular}{@{}c@{}}Structured Mask\\ ($P_t=0.6, P_f=0.5$)\end{tabular} \\
        \hlineB{2}
        \cellcolor{blue!20} WESAD &
        71.46 & 71.952 & 72.201 & \bf 75.390 \\
        \hline
        \cellcolor{blue!20} UCI-HAR &
        97.097 & 98.438 & 98.106 & \bf 98.928 \\
        \hline
        \cellcolor{blue!20} DriverFatigue &
        72.719 & 73.424 & \bf 78.354 & 75.167 \\
        \hlineB{2}
        \bf Activity Recognition Avg. &
        80.425 & 81.271 & 82.887 & \cellcolor{gray!40}\bf 83.162 \\
        \hline
        \hlineB{2}
        \cellcolor{teal!20} Epilepsy (eye open state) &
        89.521 & 91.895 & 89.407 & \bf 92.203 \\
        \hline
        \cellcolor{teal!20} Epilepsy (eye relaxation) &
        93.471 & 94.808 & 93.786 & \bf 94.908 \\
        \hline
        \cellcolor{teal!20} Epilepsy (health area) &
        86.812 & \bf 88.510 & 87.317 & 88.117 \\
        \hline
        \cellcolor{teal!20} Epilepsy (tumor area) &
        86.524 & \bf 88.254 & 85.502 & 86.888 \\
        \hline
        \cellcolor{teal!20} Epilepsy (seizure) &
        96.59 & \bf 97.791 & 95.29 & 96.638 \\
        \hline
        \cellcolor{teal!20} GAMEEMO &
        \bf 58.043 & 56.770 & 55.771 & 56.532 \\
        \hlineB{2}
        \bf EEG Main Tasks Avg. &
        85.160 & \cellcolor{gray!40}\bf 86.338 & 84.512 & 85.881 \\
        \hline
        \hlineB{2}
        \cellcolor{red!20} ECG-Abnormal &
        99.085 & \bf 99.316 & 98.296 & 99.041 \\
        \hline
        \cellcolor{red!20} PPG-BP (HTN) &
        58.880 & 55.333 & 59.230 & \bf 60.344 \\
        \hline
        \cellcolor{red!20} PPG-BP (DM) &
        61.074 & 48.386 & 58.896 & \bf 59.459 \\
        \hline
        \cellcolor{red!20} PPG-BP (CVA) &
        56.389 & 58.472 & 64.167 & \bf 70.278 \\
        \hline
        \cellcolor{red!20} PPG-BP (CVD) &
        52.572 & 46.557 & \bf 55.666 & 52.596 \\
        \hline
        \cellcolor{red!20} PhysioNet EMG &
        85.160 & 95.490 & 83.922 & \bf 98.184 \\
        \hlineB{2}
        \bf Risk Evaluation Avg. &
        68.860 & 67.259 & 70.030 & \cellcolor{gray!40}\bf 73.317 \\
        \hline
        \hlineB{2}
        \cellcolor{orange!10} Noninvasive-BP &
        90.124 & 90.650 & 91.152 & \bf 92.470 \\
        \hline
        \cellcolor{orange!10} PPG-Hgb &
        \bf 95.314 & 95.055 & 94.713 & 94.766 \\
        \hline
        \cellcolor{orange!10} Fetal-fPCG &
        98.630 & \bf 99.121 & 98.926 & 99.088 \\
        \hlineB{2}
        \bf Vital Signs Avg. &
        94.689 & 94.942 & 94.930 & \cellcolor{gray!40}\bf 95.441 \\
        \hline
        \hlineB{2}
        \bf Micro Avg. &
        80.526 & 80.568 & 81.150 & \cellcolor{gray!40}\bf 82.833 \\
        \hline
        \bf Macro Avg. &
        82.284 & 82.453 & 83.090 & \cellcolor{gray!40}\bf 84.450 \\
        \hlineB{2}
    \end{tabular}
}
\label{tab:mask-scheme-res-2}
\end{table*}
\newpage
\textbf{Input Representations.} Table\ref{tab:raw-series-input} compares the performance of two input representations: (1) CWT scalogram and (2) raw time series. The CWT scalogram converts the time series into a time-frequency representation, while the raw time series retains the original sensor data. Among the two representations, the model trained on CWT scalograms demonstrates better performance, suggesting that the time-frequency features enhance model accuracy.
\begin{table*}[ht]
\scriptsize
\centering
\caption{
Performance Comparison Between CWT Scalogram and Raw Time Series as Inputs.
}
\vspace{1mm}
\resizebox{\textwidth}{!}{%
        \begin{tabular}{|>{\centering\arraybackslash}m{0.5\textwidth}|c|c|}
        \hlineB{2}
        \hlineB{2}
        \bf \begin{tabular}{@{\hspace{-0.5\tabcolsep}}l@{\hspace{-0.5\tabcolsep}}}
            Downstream
            Tasks \\ 
        \end{tabular}  & 
        \bf \begin{tabular}{@{}c@{}}Raw Series Input\end{tabular} 
        &
        \bf \begin{tabular}{@{\hspace{-0.5\tabcolsep}}l@{\hspace{-0.5\tabcolsep}}}
            CWT Scalogram Input \\
        \end{tabular}  
        \\\hline
        \hlineB{2}
        \cellcolor{blue!20} WESAD & 
        70.862 & \bf 75.390
        \\\hline
        \cellcolor{blue!20} UCI-HAR & 
        97.969 & \bf 98.928
        \\\hline
        \cellcolor{blue!20} DriverFatigue & 
        73.854 &	\bf 75.167
        \\\hline
        \hlineB{2}
        \bf Activity Recognition Avg. &
        80.895 & \cellcolor{gray!40}\bf \textbf{\textbf{83.162}}
        \\\hline
        \hlineB{2}
        \hlineB{2}
        \cellcolor{teal!20} Epilepsy (eye open state) &
        91.978 &	\bf 92.203
        \\\hline
        \cellcolor{teal!20} Epilepsy (eye relaxation) &
        94.781 &	\bf 94.908
        \\\hline
        \cellcolor{teal!20} Epilepsy (health area) &
        88.045 &	\bf 88.117
        \\\hline
        \cellcolor{teal!20} Epilepsy (tumor area) &
        85.619 & \bf 86.888
        \\\hline
        \cellcolor{teal!20} Epilepsy (seizure) &
        \bf 97.722 &	96.638
        \\\hline
        \cellcolor{teal!20} GAMEEMO &
        54.651 & \bf 56.532
        \\\hline
        \hlineB{2}
        \bf EEG Main Tasks Avg. & 
        85.466	& \cellcolor{gray!60}\bf \textbf{85.881}
        \\\hline
        \hlineB{2}
        \hlineB{2}
        \cellcolor{red!20} ECG-Abnormal &
        97.701	& \bf 99.041
        \\\hline
         \cellcolor{red!20} PPG-BP (HTN) &
         52.614	&	\bf 60.344
        \\\hline
        \cellcolor{red!20} PPG-BP (DM) &
         \bf 62.012	& 59.459
        \\\hline
        \cellcolor{red!20} PPG-BP (CVA) &
        56.181	& \bf 70.278
        \\\hline
        \cellcolor{red!20} PPG-BP (CVD) &
        \bf 54.812	& 52.596
        \\\hline
         \cellcolor{red!20} PhysioNet EMG &
        93.756	& \bf 98.184
        \\\hline
        \hlineB{2}
        \bf Risk Evaluation Avg. & 
        69.513& \cellcolor{gray!60}\bf73.317
        \\\hline
        \hlineB{2}
        \hlineB{2}
        \cellcolor{orange!10}Noninvasive-BP &
        89.850	& \bf 92.470
        \\\hline
        \cellcolor{orange!10} PPG-Hgb &
        93.832	&	\bf 94.766
        \\\hline
        \cellcolor{orange!10} Fetal-fPCG &
        98.977	&	\bf 99.088
        \\\hline
        \hlineB{2}
        \bf Vital Signs Avg. & 
        94.220	& \cellcolor{gray!60}\bf \textbf{95.441}
        \\\hline
        \hlineB{2}
        \bf Micro Avg. & 
        80.845 &	\cellcolor{gray!60}\bf82.833
        \\\hline
        \bf Macro Avg. & 
        82.523 &	\cellcolor{gray!60}\bf\textbf{84.450}
        \\\hline
        \hlineB{2}
        \end{tabular}
}
\label{tab:raw-series-input}
\end{table*}

\textbf{Semi-Supervised Learning (Partial-shot).} To evaluate the generalizability and quality of learned representations, we conducted a semi-supervised learning evaluation following the protocol established by prior self-supervised methods \citep{swav}. Specifically, we assessed performance on the \textsc{Normwear} dataset using frozen features and a limited labeled subset (10\%). We deliberately excluded the commonly used 1\% label evaluation due to the inherently small sample size of our downstream medical dataset. A 1\% labeling scenario would provide fewer than ten labeled instances, rendering the results statistically unreliable and scientifically unjustified. Instead, we sampled 10\% of the training data while preserving the original label distribution, and then trained a linear classifier atop the frozen \textsc{Normwear} features for classification tasks and regression tasks. The results, summarized in Table \ref{tab:semi-supervised}, demonstrate the effectiveness of our method under realistic semi-supervised constraints.
\begin{table*}[ht]
    \vspace{-2mm}
    \scriptsize
    \centering
    \caption{
    \textbf{Semi-supervised learning on Downstream tasks}. We linear-prob the model with 10\% labels and report AUCROC scores.}
    \vspace{1mm}
    \resizebox{0.95\textwidth}{!}{
            \begin{tabular}{|l|c|c|c|c|c|c|}
            \hlineB{2}
            \hlineB{2}
            \bf \begin{tabular}{@{\hspace{-0.5\tabcolsep}}l@{\hspace{-0.5\tabcolsep}}}
                Downstream
                Tasks \\ 
            \end{tabular} &
            \bf Statistical
           &
            \bf Chronos
           &
            \bf CLAP 
           &
            \bf TF-C
           &
            \bf Modality-Specific 
           &
            \bf \begin{tabular}{@{\hspace{-0.5\tabcolsep}}l@{\hspace{-0.5\tabcolsep}}}
                \textsc{Normwear} (Ours) \\
            \end{tabular}  
            \\\hline
            \hlineB{2}
            \cellcolor{blue!20} WESAD&
            64.869&64.908&68.626&62.218&59.371&\bf 70.25
            \\\hline
            \cellcolor{blue!20} UCI-HAR&
            94.124&73.124&92.794&92.334&-&\bf 98.355
            \\\hline
            \cellcolor{blue!20} DriverFatigue&
            63.237&72.454&50.193&54.613&\bf69.004&55.094
            \\\hline
            \hlineB{2}
            \bf Activity Recognition Avg.&
            74.077&70.162&70.538&69.722&-&\cellcolor{gray!60}\bf74.566
            \\\hline
            \hlineB{2}
            \hlineB{2}
            \cellcolor{teal!20} Epilepsy (eye open state)&
            82.186&80.082&84.103&88.02&\bf89.152&85.456
            \\\hline
            \cellcolor{teal!20} Epilepsy (eye relaxation)&
            87.480&81.820&88.716&93.670&\bf95.191&92.369
            \\\hline
            \cellcolor{teal!20} Epilepsy (health area)&
            86.096&77.682&82.651&84.940&\bf87.377&85.471
            \\\hline
            \cellcolor{teal!20} Epilepsy (tumor area)&
            82.153&78.364&82.579&85.450&\bf86.962&83.033
            \\\hline
            \cellcolor{teal!20} Epilepsy (seizure)&
            88.179&96.786&\bf97.386&92.900&94.063&92.345
            \\\hline
            \cellcolor{teal!20} GAMEEMO&
            \bf 54.527&50.176&51.952&49.714&52.046&52.633
            \\\hline
            \hlineB{2}
            \bf EEG Main Tasks Avg.&
            80.104&77.485&81.231&82.449&\cellcolor{gray!60}\bf84.132&81.885
            \\\hline
            \hlineB{2}
            \hlineB{2}
            \cellcolor{red!20} ECG-Abnormal&
            96.420&\bf97.613&95.432&94.769&79.918&93.921
            \\\hline
             \cellcolor{red!20} PPG-BP (HTN)&
            52.491&49.407&48.397&53.800&\bf57.544&53.967
            \\\hline
            \cellcolor{red!20} PPG-BP (DM)&
            41.254&48.574&38.664&45.383&56.532&\bf57.545
            \\\hline
            \cellcolor{red!20} PPG-BP (CVA)&
            \bf83.056&51.944&48.125&51.667&64.792&66.597
            \\\hline
            \cellcolor{red!20} PPG-BP (CVD)&
            55.753&47.547&\bf59.505&55.651&47.586&54.614
            \\\hline
             \cellcolor{red!20} PhysioNet EMG&
            \bf92.993&70.248&92.415&79.412&-&87.503
            \\\hline
            \hlineB{2}
            \bf Risk Evaluation Avg.&
            \cellcolor{gray!60}\bf70.328&60.888&63.756&63.447&-&69.025 
            \\\hline
            \hlineB{2}
            \hlineB{2}
            \cellcolor{orange!10}Noninvasive-BP&
            90.589&\bf93.783&91.614&92.707&92.671&90.694 
            \\\hline
            \cellcolor{orange!10} PPG-Hgb&
            \bf95.068&94.999&94.712&94.981&94.916&94.633
            \\\hline
            \cellcolor{orange!10} Fetal-fPCG&
            \bf99.020&99.153&98.889&98.902&-&98.813
            \\\hline
            \hlineB{2}
            \bf Vital Signs Avg.&
            94.892&\cellcolor{gray!60}\bf95.978&95.072&95.530&-&94.713
            \\\hline
            \hlineB{2}
            \bf Micro Avg.&
            78.305&73.815&75.931&76.174&-&\cellcolor{gray!60}\bf78.516
            \\\hline
            \bf Macro Avg.&
            79.850&76.129&77.649&77.787&-&\cellcolor{gray!60}\bf80.047 
            \\\hline
            \hlineB{2}
            \end{tabular}
    }
    \label{tab:semi-supervised}
    \vspace{-4mm}
    \end{table*}
\newpage
\textbf{Permutation-Invariant Input Channel Analysis.} In many multimodal or multichannel sensing tasks, the input channel order is typically fixed and determined by hardware or preprocessing pipelines, limiting flexibility during deployment. This constraint raises the question of whether \textit{Normwear} relies on a specific channel ordering to perform well. To examine this, we conducted an experiment on datasets with multiple input channels by circularly shifting the channel order by one position and evaluating the resulting model performance. As shown in Table~\ref{tab:channel-permutation}, the model performance remains consistent across different permutations. These results suggest that our model does not rely on a fixed input channel configuration and is robust to variations in channel ordering, making it more applicable in practical scenarios where such inconsistencies may occur.

\textbf{$k$-fold Analysis.} To evaluate whether Normwear maintains consistent performance on datasets with limited subject diversity, we conducted 5-fold cross-validation stratified by subject ID. We applied this protocol to all downstream tasks containing 30 or fewer subjects to ensure a robust assessment. As shown in Table \ref{tab:5fold-res}, our model consistently outperformed the baselines across all tasks, demonstrating the robustness of our evaluation metric.
\begin{table*}[ht]
    \centering
    \begin{minipage}{0.38\textwidth}
      \centering
      \scriptsize
    \caption{\textbf{Performance of NormWear with original input channel order compared to random shuffling across tasks.}}
      \vspace{5.8mm}
      \resizebox{\linewidth}{!}{
      \begin{tabular}{|l|c|c|}
        \hlineB{2}
        \bf Task & \bf Original Order & \bf Random Shuffle \\
        \hlineB{2}
        \cellcolor{blue!20} WESAD (IMU, PPG, ECG, GSR) & 0.761 & 0.763 \\
        \hline
        \cellcolor{blue!20}UCI-HAR (IMU) & 0.989 & 0.975 \\
        \hline
        \cellcolor{blue!20}Drive Fatigue (EEG) & 0.743 & 0.721 \\
        \hline
        \cellcolor{teal!20}GAMEEMO (EEG) & 0.549 & 0.530 \\
        \hline
        \cellcolor{orange!10}Noninvasive-BP (PCG, PPG, ECG) & 0.924 & 0.914 \\
        \hline
        \cellcolor{orange!10}PPG-HGB (Red, IR) & 0.946 & 0.948 \\
        \hlineB{2}
      \end{tabular}}
      \label{tab:channel-permutation}
    \end{minipage}
    \hfill
    \begin{minipage}{0.60\textwidth}
      \centering
      \scriptsize
      \caption{\textbf{Performance on downstream health-related tasks under linear probing using 5-fold subject-stratified cross-validation.} Classification reports AUC ROC; regression reports relative accuracy. All metrics are higher-is-better.}
      \vspace{2mm}
      \resizebox{\linewidth}{!}{
      \begin{tabular}{|l|c|c|c|c|c|}
        \hlineB{2}
        \hlineB{2}
        \bf \begin{tabular}{@{\hspace{-0.5\tabcolsep}}l@{\hspace{-0.5\tabcolsep}}}Downstream Tasks\end{tabular}  & 
        \bf Statistical & 
        \bf Chronos & 
        \bf CLAP & 
        \bf TF-C &
        \bf \begin{tabular}{@{\hspace{-0.5\tabcolsep}}l@{\hspace{-0.5\tabcolsep}}}NormWear-L\\(Ours)\end{tabular}  \\
        \hlineB{2}
        \cellcolor{blue!20} WESAD & 79.992 $\pm$ 0.707 & 83.332 $\pm$ 0.841 & 87.824 $\pm$ 0.463 & 82.701 $\pm$ 0.536 & \bf 89.585 $\pm$ 0.683 \\
        \hline
        \cellcolor{blue!20} UCI-HAR & 95.602 $\pm$ 0.148 & 91.956 $\pm$ 0.256 & 96.864 $\pm$ 0.175 & 97.382 $\pm$ 0.138 & \bf 98.179 $\pm$ 0.06 \\
        \hline
        \cellcolor{blue!20} DriverFatigue & 69.614 $\pm$ 1.138 & \bf 72.48 $\pm$ 2.848 & 66.251 $\pm$ 0.471 & 65.026 $\pm$ 1.198 & 68.971 $\pm$ 1.32 \\
        \hline
        \cellcolor{teal!20} GAMEEMO & 64.281 $\pm$ 1.292 & 56.694 $\pm$ 0.878 & 64.119 $\pm$ 0.543 & 62.925 $\pm$ 0.999 & \bf 67.863 $\pm$ 0.72 \\
        \hline
        \cellcolor{orange!10}Noninvasive-BP & 92.83 $\pm$ 0.386 & 92.223 $\pm$ 0.356 & 92.612 $\pm$ 0.272 & 88.707 $\pm$ 0.622 & \bf 93.381 $\pm$ 0.516 \\
        \hline
        \hlineB{2}
        \bf Avg. & 80.464 $\pm$ 0.734 & 79.337 $\pm$ 1.036 & 81.534 $\pm$ 0.385 & 79.348 $\pm$ 0.699 & \bf 83.596 $\pm$ 0.660 \\
        \hlineB{2}
      \end{tabular}
      }
      \label{tab:5fold-res}
    \end{minipage}
  \end{table*}  
\newpage
\textbf{Demographic Anlysis.} 
\edit{Several previous works \citep{apple-ppg-ecg,google-scale-foundation} have used learned representations to infer demographic labels. These results suggest that wearable signals do contain demographic information. In Table \ref{tab:demo-res}, we wanted to investigate that NormWear does not extract only demographic information (e.g. age, sex, height, etc. depending on what is available within each dataset), hence indicating that the representation that our proposed model extracted and the demographic could be used as complementary features to each other during downstream modeling. }
From Table \ref{tab:demo-res}, we observe that demographic information and wearable signal representations each excel at different tasks. In most cases, concatenating them improves overall performance. However, the occasional performance drop after concatenation suggests a confounding relationship between the two, implying that demographic data and NormWear's wearable representations capture different aspects. 
\begin{table*}[t]
    \vspace{-4mm}
    \scriptsize
    \centering
    \caption{
        \textbf{Checking reliance on demographic information.} Simple baseline: for regression tasks (yellow), the mean prediction is used; for classification tasks (blue and red), the mode prediction is used. NormWear-Medium and NormWear-Large refer to NormWear's pretrained checkpoints trained on 2.58 million and 8.97 million signal segments, respectively.
    }
    \vspace{2mm}
    \resizebox{\textwidth}{!}{
            \begin{tabular}{|l|c|c|c|c|c|c|}
            \hlineB{2}
            \hlineB{2}
            \bf \begin{tabular}{@{\hspace{-0.5\tabcolsep}}l@{\hspace{-0.5\tabcolsep}}}
                Downstream
                Tasks \\ 
            \end{tabular}  & 
            \bf \begin{tabular}{@{\hspace{-0.5\tabcolsep}}l@{\hspace{-0.5\tabcolsep}}}
                \edit{Empirical Distribution}
            \end{tabular}
            & 
            \bf Demographic
            & 
            \bf NormWear-Medium 
            & 
            \bf \begin{tabular}{@{\hspace{-0.5\tabcolsep}}l@{\hspace{-0.5\tabcolsep}}}
                Demographic + \\
                NormWear-Medium 
            \end{tabular}
            & 
            \bf NormWear-Large 
            &
            \bf \begin{tabular}{@{\hspace{-0.5\tabcolsep}}l@{\hspace{-0.5\tabcolsep}}}
                Demographic + \\
                NormWear-Large 
            \end{tabular}  
            \\\hline
            \hlineB{2}
            \cellcolor{blue!20} WESAD & 
            50.000&	49.907&	\bf74.227&	69.06&	\bf76.06&	68.755
            \\\hline
            \cellcolor{orange!10} Noninvasive-BP & 
            92.988&	92.954&	91.427&	90.84&	92.42&	92.528
            \\\hline
            \cellcolor{orange!10} PPG-Hgb & 
            94.816&	95.634&	94.911&	95.835&	94.632&	96.384
            \\\hline
            \cellcolor{orange!10} Fetal-fPCG & 
            99.033&	99.039&	98.997&	99.001	&99.072	&99.097
            \\\hline
            Vital Signs Avg. & 
            95.612&	\bf95.876&	95.112&	95.225&	95.375&	\bf96.003
            \\\hline
            \hlineB{2}
            \cellcolor{red!20} PPG-BP (HTN) & 
            50.000&	59.899&	62.746&	64.482&	62.341	&61.291
            \\\hline
            \cellcolor{red!20} PPG-BP (DM) & 
            50.000&	47.297&	62.613&	47.86&	55.893&	60.135
            \\\hline
            \cellcolor{red!20} PPG-BP (CVA) & 
            50.000&	81.875&	67.639&	83.681&	70.625&	77.847
            \\\hline
            \cellcolor{red!20} PPG-BP (CVD) & 
            50.000&	71.011&	51.504&	70.37&	51.773&	67.466
            \\\hline
            Risk Evaluation Avg. & 
            50.000&	65.021&	61.126&	\bf66.598&	60.158&	\bf66.685
            \\\hline
            \hlineB{2}
            Micro Avg. & 
            67.105&	74.702&	75.508&	\bf77.641&	75.352&	\bf77.938
            \\\hline
            Macro Avg. & 
            65.204&	70.268&	76.821& \bf76.961&	\bf77.198&77.148
            \\\hline
            \hlineB{2}
            \end{tabular}
    }
    \label{tab:demo-res}
    \end{table*}
\section{Statistical significance on the model comparison}
We performed a statistical analysis to test the significance of the differences in model performance. First, we ran the downstream evaluations 100 times for each model on every task without fixing the random seed. The results remained consistent due to the stable optimization process.

\begin{figure}[hbt!]
    \centering
    \begin{minipage}[t]{0.4\textwidth}
    \resizebox{\textwidth}{!}{%
    \vspace{10mm}
    \begin{tabular}{|lcccc|}
    \hline
    \textbf{Ours/Baselines}                     & \textbf{Stats} & \textbf{Chronos} & \textbf{CLAP} & \textbf{TFC} \\ \hline
    \textbf{NormWear} - activity           & $P<.01$          & $P<.01$            & $P<.01$         & $P<.01 $       \\
    \textbf{NormWear} - eeg                & $P<.01$          & $P<.01$            & $P<.01$         & $P<.01$        \\
    \textbf{NormWear} - risk               & $P<.01 $         & $P<.01$            & $P<.01$         & $P<.01$        \\
    \textbf{NormWear} - vital              & $P<.01 $         & $P<.01 $           & $P<.01$         & $P<.01$        \\
    \textbf{NormWear} - micro avg.         & $P<.01$          & $P<.01$            & $P<.01$         & $P<.01$        \\
    \textbf{NormWear} - macro avg.         & $P<.01 $         & $P<.01  $          & $P<.01$         & $P<.01$        \\ \hline
    \hlineB{2}
    Conover post hoc              & $P<.001$          & $P<.001$            & $P<.001 $        & $P<.05$         \\ \hline
    \end{tabular}%
    }
    \caption{Permutation test on models' performance.}
    \label{tab:perm-test-res}
    \end{minipage}%
    \hfill
    \begin{minipage}[t]{0.58\textwidth}
    \centering
    \vspace{-11mm}
    \includegraphics[width=\textwidth]{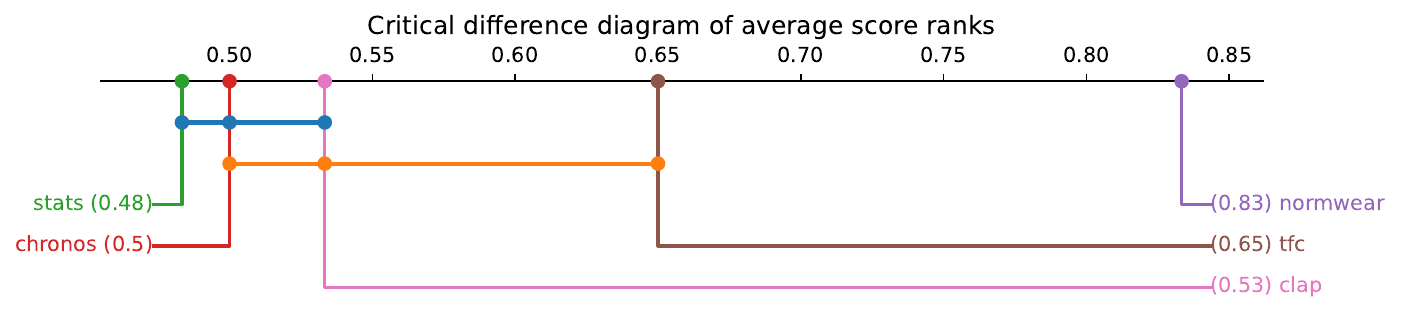}
    \caption{Critical Difference Diagram.}
    \label{fig:cd-diagram}
    \end{minipage}
\end{figure}

Next, we applied a permutation test on the results from these 100 runs to determine whether NormWear's AUC ROC score is greater than that of the baselines. The reported p-value represents the probability of observing a test statistic as extreme or more extreme than the observed difference under the null hypothesis, which assumes that NormWear's score is not higher than the baseline. In nearly all cases, the p-value is less than 0.01, confirming the statistical significance and indicating the robustness and superiority of our approach. Table \ref{tab:perm-test-res} presents the statistical test results across different task groups (as indicated by the color coding in the main tables) along with the overall average scores.

We also include a critical difference (CD) diagram to visually compare the performance of multiple models across datasets and highlight statistically significant differences. To generate the CD diagram, we first conducted a Friedman Chi-square test on the models' scores across all downstream tasks, which yielded a p-value of $P < 0.001$, confirming that the models' performances come from different distributions. We then applied the Conover post hoc test to examine pairwise differences between model performances; the p-values for NormWear compared with the baselines are shown in the last row of Table \ref{tab:perm-test-res}. Finally, based on these results, we generated the CD diagram displayed in Figure \ref{fig:cd-diagram}. In this diagram, our proposed model, NormWear, is well separated from the others, indicating its statistical superiority over the competitive baselines.
\newpage
\section{Supplementary Metrics}
\textit{Normwear}'s performance is summarized in Fig. \ref{fig:radar-plot} and detailed in Table \ref{tab:main-res-metrics}. Normwear consistently exceeds the baseline models by a wide margin, demonstrating a clear advantage.
\vspace{-2mm}
\begin{table*}[hbt!]
    \centering
    \caption{
    \textbf{Details of Incidental Performance Metrics.}
    }
    \vspace{2mm}
    \resizebox{\textwidth}{!}{
        \begin{tabular}{|l|l|c|c|c|c|c|c|}
        \hlineB{2}
        \hlineB{2}
        \bf Task Group & 
        \bf \begin{tabular}{@{\hspace{-0.5\tabcolsep}}l@{\hspace{-0.5\tabcolsep}}}
            Methods \\ 
        \end{tabular}  & 
        \bf AUC ROC
        & 
        \bf AUC PR
        & 
        \bf Accuracy
        & 
        \bf Precision
        &
        \bf Recall
        &
        \bf F1 Score
        \\\hline
        \hlineB{2}
        \cline{2-8} \cellcolor{blue!20}& Statistical & 
        75.082 & 63.996 & 65.298 & 61.450 & 61.56 & 61.034
        \\
        \cline{2-8} \cellcolor{blue!20}Activity &  Chronos & 
        79.935 & 65.622 & 66.175 & 62.044 & 61.512 & 60.522
        \\
        \cline{2-8} \cellcolor{blue!20}Recognition &  CLAP & 
        76.897 & 67.026 & 66.349 &	62.790 & 62.826 & 62.435
        \\
        \cline{2-8} \cellcolor{blue!20}&  TF-C & 
        77.880 & 68.228 & 67.175 & 64.967 & 64.798 & 64.783
        \\
        \cline{2-8} \cellcolor{blue!20}& NormWear (Ours) & 
        \cellcolor{gray!60}\bf 83.102 &	\cellcolor{gray!60}\bf 76.232 & \cellcolor{gray!60}\bf 75.254 & \cellcolor{gray!60}\bf 72.606 & \cellcolor{gray!60}\bf 72.177 & \cellcolor{gray!60}\bf 72.053
        \\\hline        
        \hlineB{2}
        \cline{2-8} \cellcolor{teal!20} &  Statistical & 
        79.720 & 50.172 & 73.921 & 63.567 & 57.529 & 57.948
        \\
        \cline{2-8} \cellcolor{teal!20}EEG Main &  Chronos & 
        80.677 & 55.507 & 75.285 & 72.442 & 52.520 & 47.671
        \\
        \cline{2-8} \cellcolor{teal!20}Tasks &  CLAP & 
        82.100 & 57.518 & 76.391 & 68.506 & 61.961 & 62.650
        \\
        \cline{2-8} \cellcolor{teal!20}&  TF-C & 
        84.302 & 61.864 & 76.825 & 71.702 & 65.517 & 67.889
        \\
        \cline{2-8} \cellcolor{teal!20}&  NormWear (Ours) & 
        \cellcolor{gray!60}\bf 85.883 &	\cellcolor{gray!60}\bf 66.841 & \cellcolor{gray!60}\bf 79.182 & \cellcolor{gray!60}\bf 72.485 & \cellcolor{gray!60}\bf 69.158 & \cellcolor{gray!60}\bf 69.698
        \\\hline        
        \hlineB{2}
        \cline{2-8} \cellcolor{red!20}&  Statistical & 
        71.032 & \cellcolor{gray!60}\bf 53.783 & 79.688 & 52.718 & 53.235 & 50.807
        \\
        \cline{2-8} \cellcolor{red!20}Disease Risk & Chronos & 
        62.060 & 40.673 & 71.910 & 45.512 & 43.739 & 40.569
        \\
        \cline{2-8} \cellcolor{red!20}Evaluation &  CLAP & 
        66.274 & 48.232 & 81.327 & 53.028 & 54.721 & 52.804
        \\
        \cline{2-8} \cellcolor{red!20}&  TF-C & 
        69.416 & 46.312 & 78.929 &	52.123 & 52.352 & 51.349
        \\
        \cline{2-8} \cellcolor{red!20} &  NormWear (Ours) & 
        \cellcolor{gray!60}\bf 73.165 &	51.666 & \cellcolor{gray!60}\bf 81.530 & \cellcolor{gray!60}\bf 54.133 & \cellcolor{gray!60}\bf 56.314 & \cellcolor{gray!60}\bf 54.428
        \\\hline        
        \hlineB{2}
        \cline{2-8} & \cellcolor{gray!0} Statistical & 
        75.317 & 51.596 & 74.503 & 58.804 & 56.618 & 55.709
        \\
        \cline{2-8} Micro & \cellcolor{gray!0} Chronos & 
        73.082 & 51.596 & 72.113 & 59.590 & 50.806 & 47.401
        \\
        \cline{2-8} Average & \cellcolor{gray!0} CLAP & 
        74.729 & 55.705 & 76.357 & 61.171 & 59.238 & 58.669
        \\
        \cline{2-8} & \cellcolor{gray!0} TF-C & 
        77.063 & 56.916 & 75.737 & 62.523 & 60.107 & 60.652
        \\
        \cline{2-8} & \cellcolor{gray!0} NormWear (Ours) & 
        \cellcolor{gray!60}\bf 80.240 &	\cellcolor{gray!60}\bf 62.649 & \cellcolor{gray!60}\bf 79.336 & \cellcolor{gray!60}\bf 65.168 & \cellcolor{gray!60}\bf 64.624 & \cellcolor{gray!60}\bf 64.061
        \\\hline        
        \hlineB{2}
        \cline{2-8} &  Statistical & 
        75.278 & 55.983 & 72.969 & 59.245 & 57.441 & 56.596
        \\
        \cline{2-8} Macro &  Chronos & 
        74.224 & 53.934 & 71.123 & 59.999 & 52.590 & 49.587
        \\
        \cline{2-8} Average &  CLAP & 
        75.091 & 57.592 & 74.689 & 61.441 & 59.836 & 59.296
        \\
        \cline{2-8} & TF-C & 
        77.199 & 58.801 & 74.310 & 62.931 & 60.889 & 61.340
        \\
        \cline{2-8} &  NormWear (Ours) & 
        \cellcolor{gray!60}\bf 80.717 &	\cellcolor{gray!60}\bf 64.913 & \cellcolor{gray!60}\bf 78.656 & \cellcolor{gray!60}\bf 66.408 & \cellcolor{gray!60}\bf 65.883 & \cellcolor{gray!60}\bf 65.393 
        \\\hline
        \hlineB{2}
        \end{tabular}
    }
    \label{tab:main-res-metrics}
\end{table*}
\clearpage
\section{\edit{Scaling up the Pretraining Data Size}}
\begin{wrapfigure}{r}{0.35\textwidth}
    \centering
    \vspace{-5mm}
    \includegraphics[width=\linewidth]{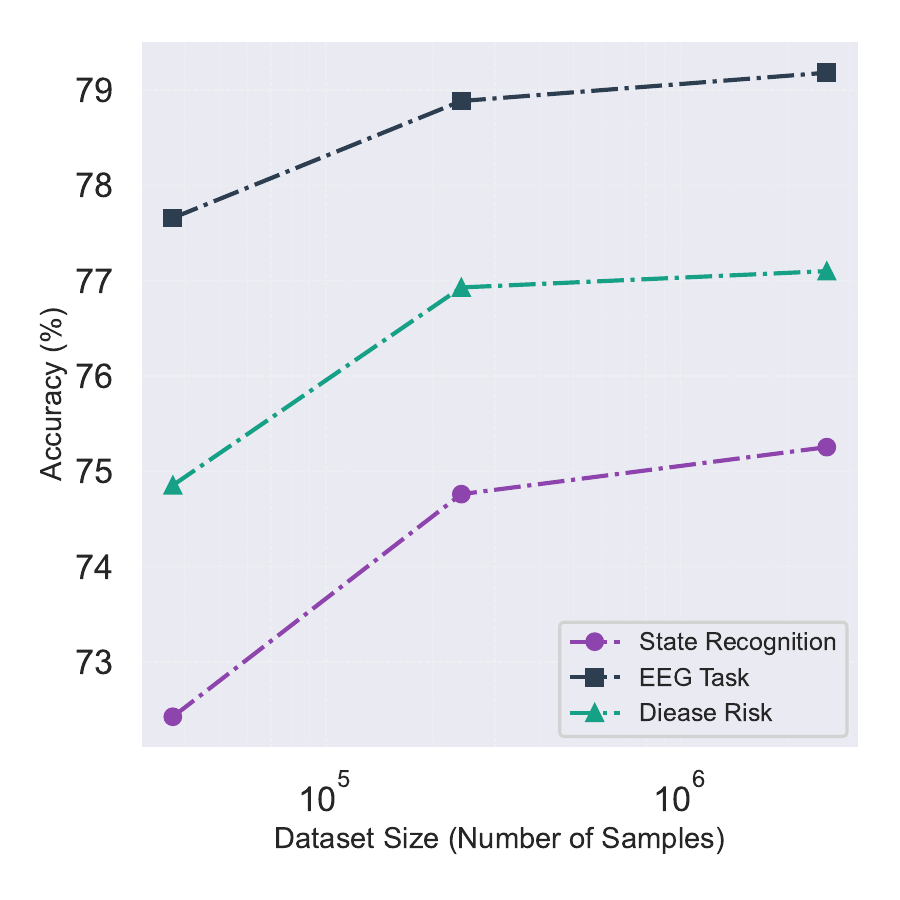}
    \vspace{-8mm}
    \caption{\textbf{Impact of scaling the pretraining dataset on downstream tasks.} The y-axis represents the average accuracy across tasks, while the x-axis denotes the size of the pretraining dataset in terms of the number of samples.}
    \label{fig:scale-law-plot}
\end{wrapfigure}
In addition to demonstrating that NormWear outperforms all strong baselines, we further investigate the effect of varying pretraining data size on the model's downstream performance to examine whether the scaling law applies to our proposed methodology. As shown in Figure \ref{fig:scale-law-plot}, the overall performance (measured by accuracy) significantly improves as the pretraining data size increases from approximately 37k (62 hours) to nearly 2.5M (4000 hours) samples of wearable signal data. This observation indicates that our model adheres to the scaling law, highlighting its potential scalability and suitability for future large-scale applications.

\section{Channel Fusion Complexity analysis}
\begin{figure}[hbt!]
    \vspace{-4mm}
    \centering
    \begin{tabular}{@{}c@{}}
        \includegraphics[width=0.49\textwidth]{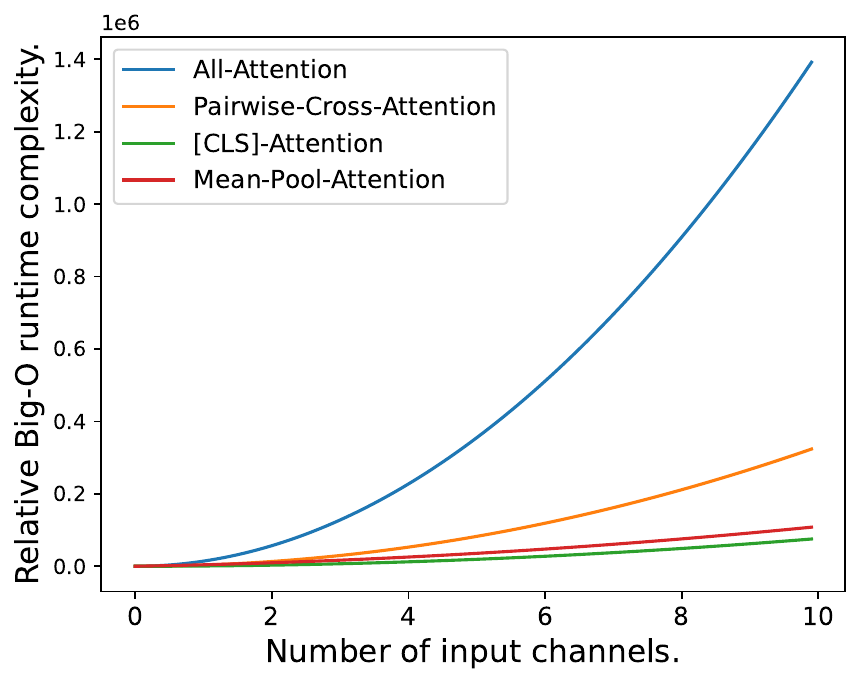} \\
        \small (a) Runtime Big-O when increasing \\
        \small the number of channels
    \end{tabular}
    \begin{tabular}{@{}c@{}}
        \includegraphics[width=0.49\textwidth]{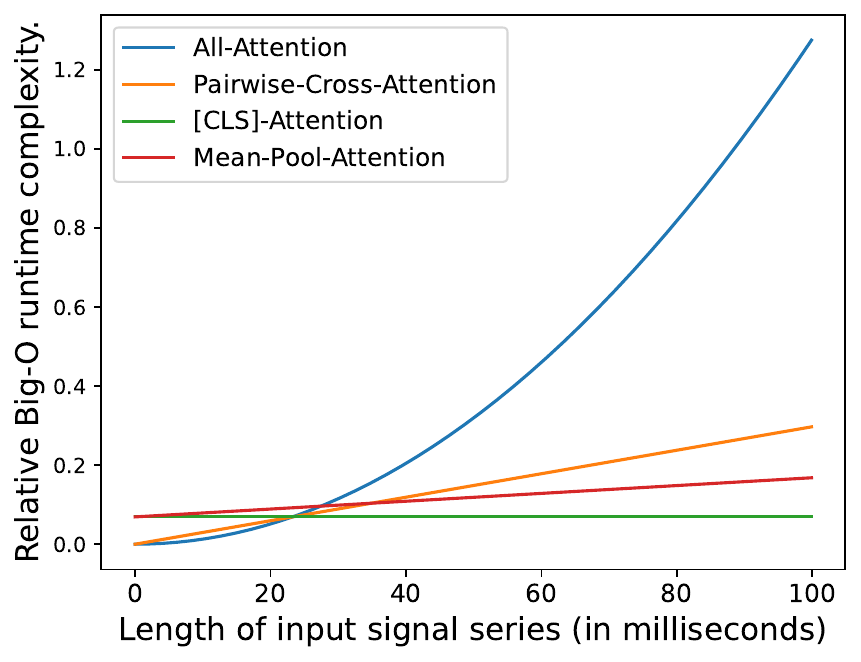}\\
        \small (b) Runtime Big-O when increasing \\
        \small the sequence length
    \end{tabular}
    \caption{\textbf{Visualization of runtime complexity when scaling up the number of channels or the sequence length.}}
    \label{fig:attns-comp-compare}
    \vspace{-4mm}
    \end{figure}

When conducting multi-channel modeling, for example, when the input comprises an arbitrary number of signals, a fusion operation needs to be conducted across all channels in order to let the model extract correlation information. Because we will deploy the model on an edge device like Jetson Nano, other than empirical evidence of the performance, we also have to consider the computation complexity of different approaches. A brief visualization of the runtime complexity of different approaches is presented in figure \ref{fig:attns-comp-compare}. The detailed derivation is presented in the following subsections. 
\subsection{All-Attention}
For the approach of conducting self-attention by concatenating all the patches, we arrive the Big-O complexity expression as follows:
\begin{itemize}
    \item We denote $C$ as the number of input channels, $d$ as the embedding size, $L$ as the number of patches convolved from the time series in each channel (proportional to sequence length), and $x \in \mathbb{R}^{C\times L\times d}$ as the input data before feeding into the fusion block. We have a total of $L \cdot C$ patches. 
    \item When calculating the attention scores, dot products are computed for each pair of the patches, which results in the following calculation process:

\begin{algorithm}
\caption{All-Attention Complexity}\label{alg:all-attn}
    \begin{algorithmic}
    \FOR{$i \in [1,2,\dots,C]$}
        \FOR{$j \in [1,2,\dots,L]$}
            \STATE $N \gets \exp(\text{attn}(x_{i,j})) \quad \implies O(L \cdot C)$
            \FOR{$k \in [1,2,\dots,C]$}
                \FOR{$l \in [1,2,\dots,L]$}
                    \STATE 1) Calculate dot product: $\text{attn}(x_{i,j}, x_{k,l}) = x_{i,j}^T x_{k,l} \quad \implies O(2d)$
                    \STATE 2) Softmax over attention scores: $\frac{\exp(\text{attn}(x_{i,j}, x_{k,l}))}{N} \quad \implies O(1)$
                    \STATE 3) Weighted average: $x_{i,j} + \text{attn}(x_{i,j}, x_{k,l}) \cdot x_{k,l} \quad \implies O(2d)$
                \ENDFOR
            \ENDFOR
        \ENDFOR
    \ENDFOR
    \end{algorithmic}
\end{algorithm}

    where "1), 2), 3)" represents the operations conducted at the first, second, and third rounds of entering the entire nested loops. The complexity for the first round of operation results in a complexity of:
    \begin{equation} \label{eq:att-comp}
        \sum_{i=1}^C\sum_{j=1}^L\sum_{k=1}^C\sum_{l=1}^L 2d = 
        \sum_{i=1}^C\sum_{j=1}^L\sum_{k=1}^C L \cdot 2d = 
        \sum_{i=1}^C\sum_{j=1}^L C \cdot L \cdot 2d = O(d \cdot (L\cdot C)^2)
    \end{equation}
    where in the case of multi-head attention, the dot product still has the complexity of $O(2d)$, and because the number of heads is a constant, the final complexity is equivalent to the result in equation \ref{eq:att-comp}. 
    \item Similarly, the softmax operation will result in a complexity of $O((L\cdot C)^2)$, and the final weighted average operation will also have a complexity of $O(d\cdot (L\cdot C)^2)$, which results in total complexity of
    \begin{equation}
        O(d\cdot (L\cdot C)^2) + O((L\cdot C)^2) + O(d\cdot (L\cdot C)^2) = O(d\cdot (L\cdot C)^2)
    \end{equation}
\end{itemize}

\subsection{Cross-Attention}
For the pairwise cross-attention approach following guidance of \citet{chen2021crossvit}, we have the operation defined as

\begin{algorithm}
\caption{Cross-Attention Complexity}\label{alg:cross-attn}
\begin{algorithmic}
\FOR{$i$ in $[1,2,...,C-1]$}
    \FOR {$j$ in $[1,2,...,C]$}
        \STATE 2) $N=\exp(\text{attn}(x_{i,1})), \implies O(L)$
        \FOR {$k$ in $[2,3,...,L]$}
            \STATE 1) Calculate $\text{attn}(x_{i,1}, x_{j,k}), \implies O(2d)$
            \STATE 2) Softmax over all-attention scores, $\frac{\exp(\text{attn}(x_{i,1}, x_{j,k}))}{N}$, $\implies O(1)$
            \STATE 3) Weighted average: $x_{i,1}+x_{j,k}$, $\implies O(2d)$
        \ENDFOR
    \ENDFOR
\ENDFOR
\end{algorithmic}
\end{algorithm}

with the same notion in the previous subsection. The total complexity is
\begin{equation}
    O(C^2 \cdot L \cdot 2d) + O(C^2 \cdot L) + O(C^2 \cdot L \cdot 2d) = O(d\cdot L \cdot C^2)
\end{equation}

\subsection{\textsc[CLS]-Attention}
This is the approach that we adopted for the final version of our proposed foundation model. Only the embedding corresponding to the $[\text{CLS}]$ token of each channel is involved during the self-attention operation. Therefore, the complexity is 
\begin{equation}
    O(d\cdot C^2)
\end{equation}

\subsection{Mean-pool Attention}
For fusion with mean-pool attention, we first calculate the mean representation for each channel, resulting in a complexity of $O(C\cdot L\cdot d)$. And self-attention with Tese mean representations has the same complexity as \textsc[CLS]-attention, which is $O(d\cdot C^2)$. Thus, the total complexity is 
\begin{equation}
    O(C\cdot L\cdot d) + O(d\cdot C^2) = O(d\cdot (L\cdot C + C^2))
\end{equation}

\section{MSiTF Complexity analysis}
\begin{algorithm}
\caption{MSiTF Runtime Complexity}\label{alg:msitf}
\begin{algorithmic}
\STATE key embedding $E_k$ = $k(S) \in \mathbb{R}^{p\times d}, \implies O(d^2)$
\STATE value embedding $E_v$ = $v(S) \in \mathbb{R}^{p\times d}, \implies O(d^2)$
\STATE Relevance score $Rel = E_k^T Q \in \mathbb{R}^{p}, \implies O(pd)$
\STATE likelihood parameter $E_l$ = $l(S) \in \mathbb{R}^{p\times 2}, \implies O(d^2)$
\STATE Importance score sampling $W_{imp} \in \mathbb{R}^{p}$ (equation \ref{eq:importance}) $\implies O(p)$
\STATE Fused embedding $E_{final} = E_v^T (\alpha W_{imp}+\beta W_{rel}+\kappa W_{rec}) \in \mathbb{R}^d, \implies O(pd)$
\STATE Inference final score $c = \argmax_{i \in |C|} \;\;C_i^T E_{final}, \implies O(cd)$
\end{algorithmic}
\end{algorithm}

Where $d$ being the latent size, $p$ being the number of total patches, $c$ being the number of available ground truth choice, $k$ and $v$ being the key and value linear mapping, $S \in \mathbb{R}^{p \times d}$ as the signal embeddings, $Q$ as the query sentence embedding, and $C$ as the list of available answer choice sentences. The total runtime complexity is $O(d^2 + pd + p + cd)$. Since $d$ is constant, we have runtime complexity of $O(p+c)$.

Regarding memory complexity of MSiTF, with $m$ being the size of text encoder, $w$ being the size of normwear, we have (i) Signal representations: $O(pd)$; (ii) Text representations: $O(cd)$; (iii) Total: $O(m+w+d(p+c))$. Since $m$, $w$, and $d$ are all constants, we have memory complexity of $O(p+c)$.

\section{Feature Visualization}
Feature visualization serves as a tool to interpret and analyze the internal representations learned by the model. By examining activation patterns or embedding structures at various layers, we aim to understand how the model encodes input signals and whether these representations align with relevant semantic or structural information. This analysis provides insight into the effectiveness of the learned features and can inform architectural or training modifications to improve performance and generalization.
\subsection{The model is agnostic to the input signals}\label{model-agnostic}
This section investigates whether, without requiring the signal modality type information as input, \textsc{NormWear} can effectively distinguish between different signal sources. We randomly sampled 500 samples for each sensor type and fed them into our pretrained model. We use t-SNE \citep{tsne}, with PCA \citep{pca} initialization to visualize the learned representations corresponding to the [CLS] special token at the last layer. The PCA preserves the global structure, while t-SNE emphasizes local relationships in the data. From Figure \ref{fig:cls-visual}(a), we observe that representations from sensors located at the same body position are clustered closely together, while representations from different body locations are clearly separated. 
This suggests that our model is signal-agnostic, as it can recognize the signal type differences, map their representations appropriately in the embedding space, and guide feature extraction within each Transformer block.

\subsection{Waveform visualization}\label{feature-visualization}
\begin{figure}[t!]
\centering
\begin{tabular}{@{}c@{}}
    \includegraphics[width=0.49\textwidth]{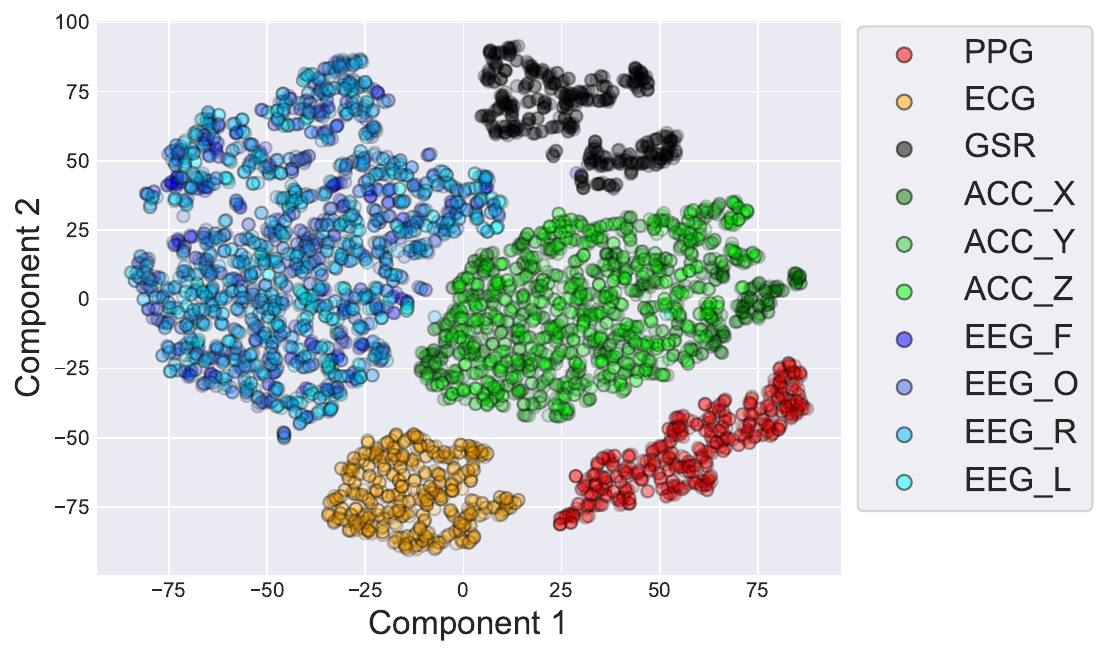}
    \\[\abovecaptionskip]
    \small \textbf{(a)} T-SNE plot of the embeddings of the \\
    \small $[\text{CLS}]$ special tokens, including \\
    \small signals of PPG, ECG, GSR, tri-axis \\
    \small accelerometer, and EEG at lobe of \\
    \small front, occiput, right, and left
\end{tabular}
\begin{tabular}{@{}c@{}}
    \includegraphics[width=0.45\textwidth]{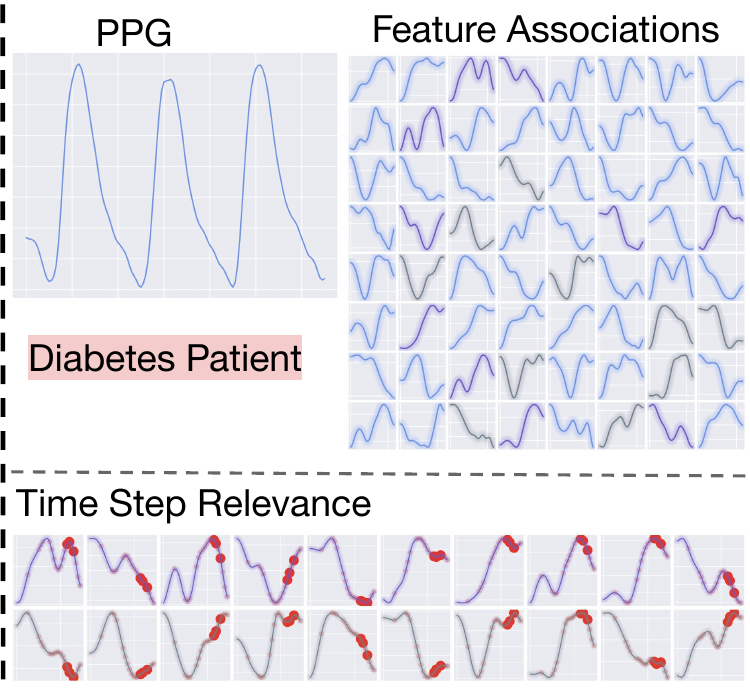}\\
    \small \textbf{(b)} Visualization of features \\
    \small extracted by the intermediate layer
\end{tabular}
\caption{Feature visualization.}
\label{fig:cls-visual}
\vspace{-2mm}
\end{figure}

Figure \ref{fig:cls-visual} (b) under ``Feature Associations" shows the features extracted by our model. Each patch corresponds to a representation with a vector size of $\mathbb{R}^{768}$. When ordered by time sequence, these representations form 768 waveforms per layer, representing the model's extracted features. The figure displays 64 randomly sampled waveforms from a selected layer. The features highlighted in purple and gray indicate the top 10 patterns positively and negatively associated with the target task (diabetes classification, in this example), with associations determined by linear regression parameters during linear probing.
Additionally, our relevance-based fusion mechanism identifies the contribution of each time step during inference, highlighted by red dots in the ``Time Step Relevance" section of Figure \ref{fig:cls-visual} (b). 

Such a visualization pipeline can assist researchers and clinicians by offering insights into how the model reaches its final predictions. Given the millions of parameters and hundreds of waveform features per layer, visualizing these features individually is inefficient for understanding the overall behavior of the proposed foundation model.
As a result, we use several techniques in nonlinear dynamic analysis \citep{nld-chaos} to quantify the overall patterns of these extracted features, which are discussed in detail in section \ref{quantify-visual}.

\subsection{Quantify the intrinsic behaviors: nonlinear dynamics analysis on the layer-wise waveforms}\label{quantify-visual}
\vspace{-2mm}
\edit{Understanding the representations extracted by intermediate layers is crucial to interpreting our model's behavior. To quantify the meaningfulness of these representations, we conducted a nonlinear dynamics analysis inspired by chaos theory. This method analyzes the features’ intrinsic behaviors through metrics like the Lyapunov exponent  \citep{lyapunov-exponent} (sensitivity to initial conditions), Hurst exponent \citep{hurst-exponent} (self-correlation/seasonality), and persistence entropy \citep{persist-entropy} (unpredictability in system states). We obtain the following key observations:}

\edit{
\textbf{1. Deeper Layers Capture Higher-Order Complexity.}
\begin{itemize}[itemsep=2pt, parsep=0pt]
    \item For signals such as GSR, EEG, and ACC, deeper layers show lower self-correlation (DFA \citep{dfa}) and higher unpredictability (persistence entropy), indicating a transition to representations that are less periodic and more chaotic.
    \item The decrease in the Lyapunov exponent across layers suggests reduced variation in extracted features, aligning with the idea that deeper layers capture more abstract, long-term patterns with broader receptive fields.
\end{itemize}
}

\edit{
\textbf{2. Modalities with Simpler Dynamics.} In contrast, PPG and ECG signals, dominated by regular heart activity, exhibit more stable patterns across layers. This aligns with their simpler waveform structures and less complex dynamics compared to signals related to neural and physical activities.
}

\edit{
These visualizations reveal that the model progressively transforms raw sensory data into representations aligned with the complexity of each signal. For GSR and EEG, deeper layers exhibit increased unpredictability and reduced periodicity, highlighting the extraction of nuanced, higher-order patterns critical for human sensing. In contrast, the stability of representations for PPG and ECG reflects their simpler dynamics, demonstrating the model's adaptability to varying signal characteristics. This analysis confirms that the intermediate representations are purposefully optimized to capture the temporal and structural nuances of each modality, supporting the conclusion that the model learns meaningful features tailored to human sensing tasks.
}

\begin{figure*}[ht]
\centering
\includegraphics[width=1.0\textwidth]{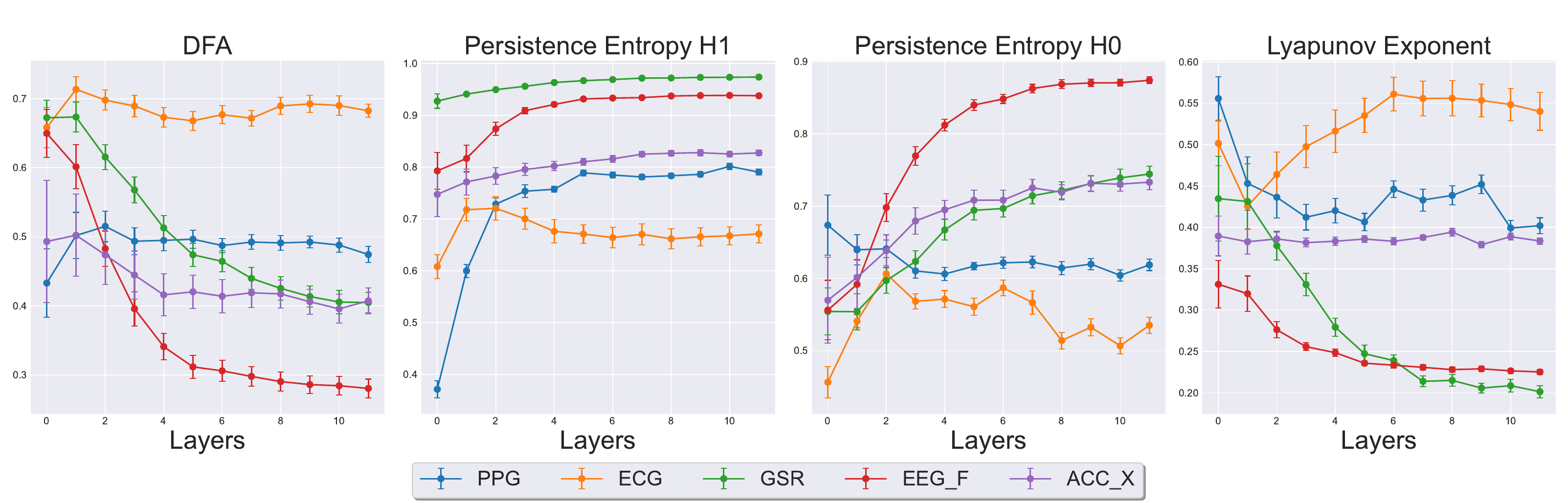}
\vspace{-5mm}
\caption{Nonlinear dynamic analysis on the waveforms extract at different layers of our model.}
\label{fig:nld-analysis}
\vspace{-4mm}
\end{figure*}

\subsection{\edit{T-SNE plot among classes}}
\begin{figure*}[ht!]
  \vspace{-2mm}
  \centering
  \begin{subfigure}[t]{0.49\textwidth}
    \centering
    \raisebox{0.05mm}{\includegraphics[width=\textwidth]{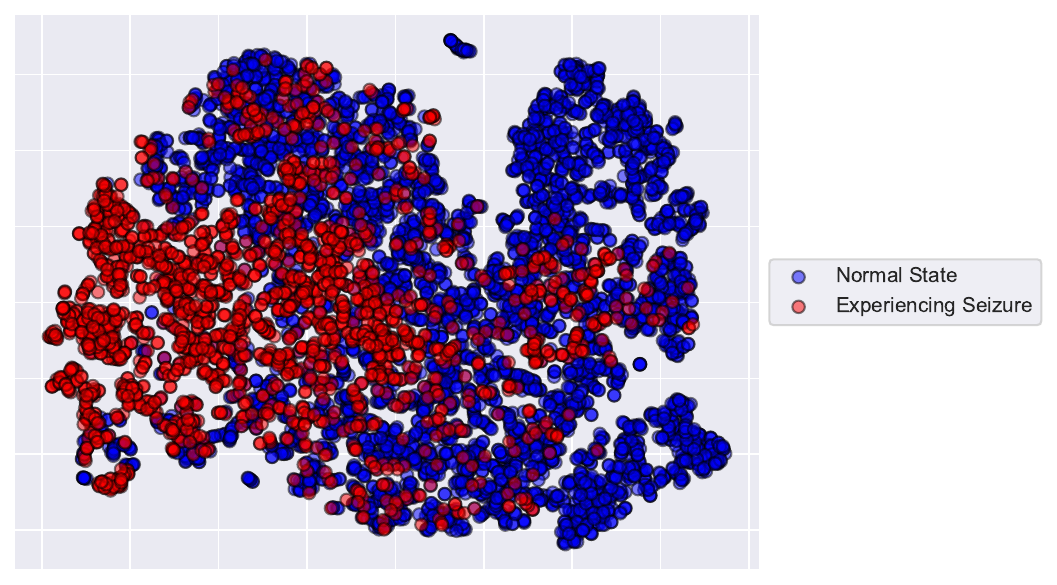}}
    \caption{\edit{Visualization of embedding on EEG signals.}}
    \label{fig:tsne-visual-eeg}
  \end{subfigure}
  \hfill
  \begin{subfigure}[t]{0.49\textwidth}
    \centering
    \includegraphics[width=\textwidth]{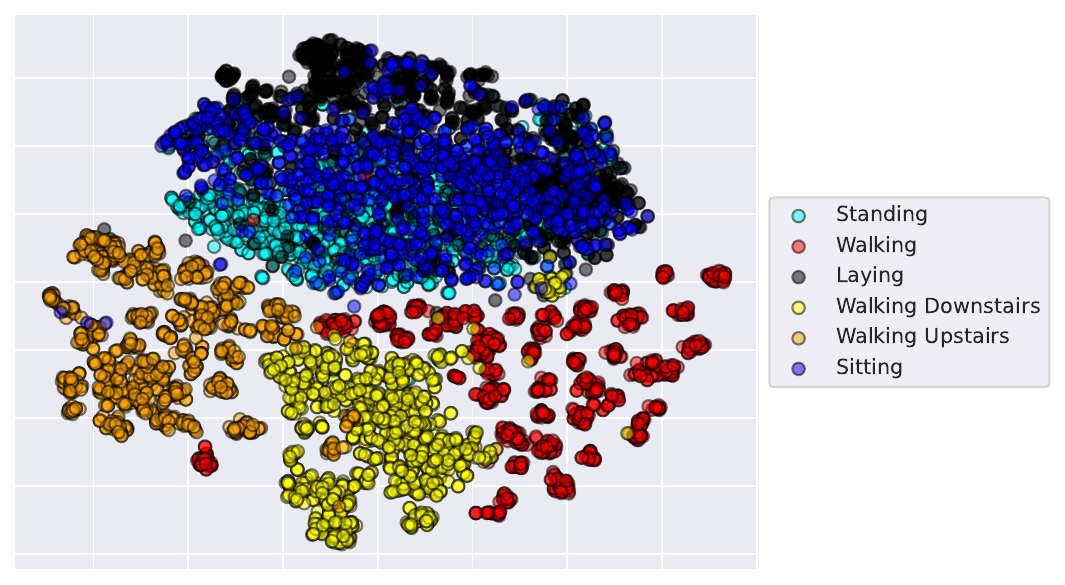}
    \caption{Visualization of embedding on signals from IMU sensors.}
    \label{fig:tsne-visual-imu}
  \end{subfigure}
  \vspace{-2mm}
  \caption{Visualization of example signal embeddings.}
  \vspace{-6mm}
\end{figure*}

\edit{In this section, we present T-SNE plots of NormWear's embeddings across different classes to provide insights into their structure and assess their suitability for sample similarity-based information retrieval. It is important to note that these plots are exploratory in nature and do not serve as a claim of the embeddings' superiority.
As shown in Figures \ref{fig:tsne-visual-eeg} and \ref{fig:tsne-visual-imu}, 
clear class separations can be observed in certain scenarios. For example, EEG samples from seizure subjects and normal subjects are distinctly separated, and physical activity types are well-clustered. For ECG data, abnormal heartbeats tend to form cohesive clusters.
However, it is essential to recognize that these T-SNE plots reduce the latent representations into a 2D space, which may not fully capture the inherent properties of the embeddings in their original high-dimensional form.}

\clearpage
\subsection{Supplementary Qualitative Analysis of MSiTF}
\begin{figure}[hbt]
    \centering
    \includegraphics[width=\linewidth, trim=0cm 0cm 3.2cm 0cm, clip]{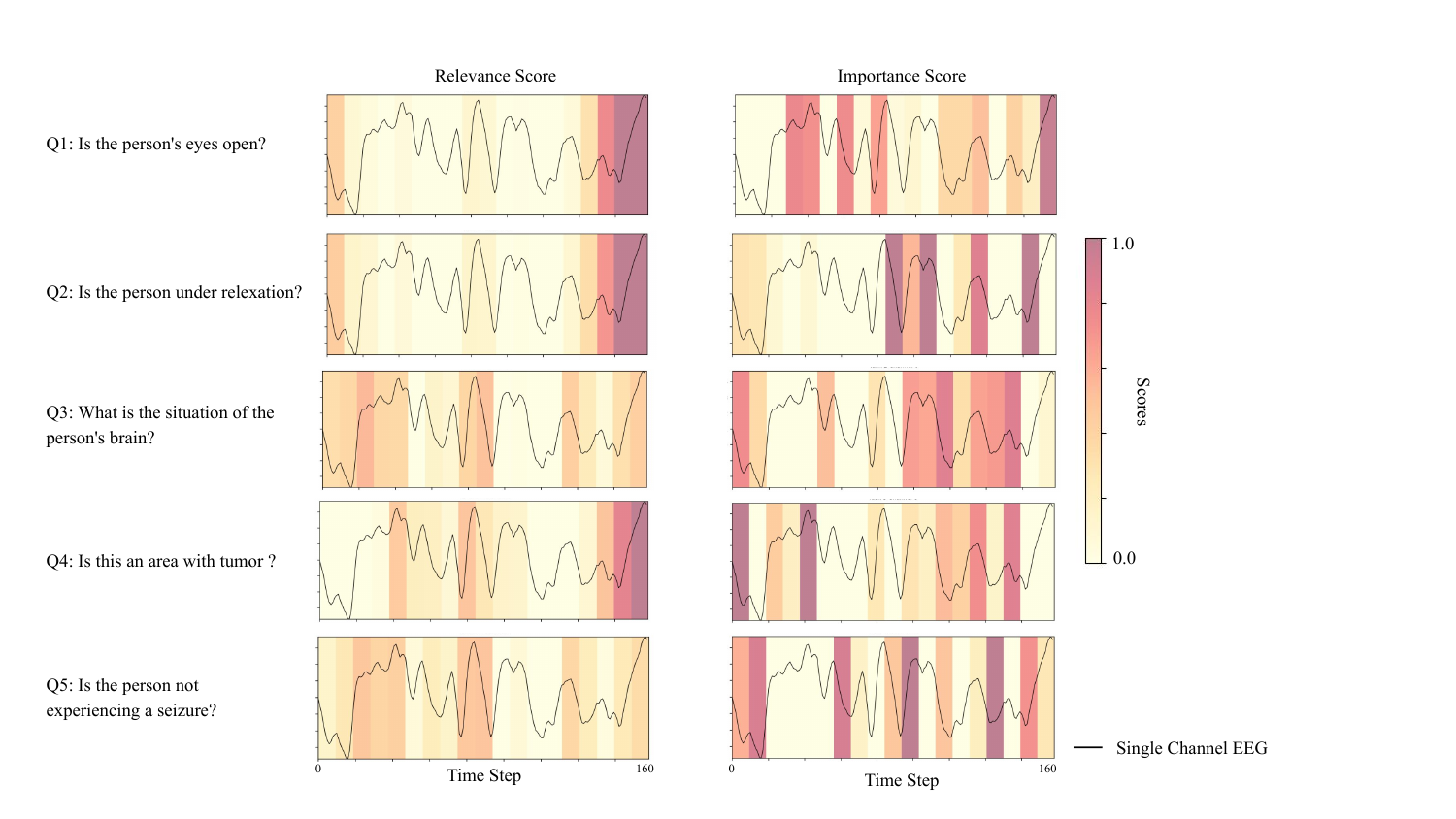}
    \caption{\textbf{Visualization of relevance scores(left) and importance scores (right) for a single channel EEG sample from the Epilepsy dataset under five task-specific questions.} The background color follows a yellow-to-red scale, where darker regions indicate higher scores.}
    \label{fig:heatmap}
\end{figure}
\begin{figure}[hbt]
    \centering
    \includegraphics[width=0.5\linewidth, trim=0cm 0cm 12cm 7cm, clip]{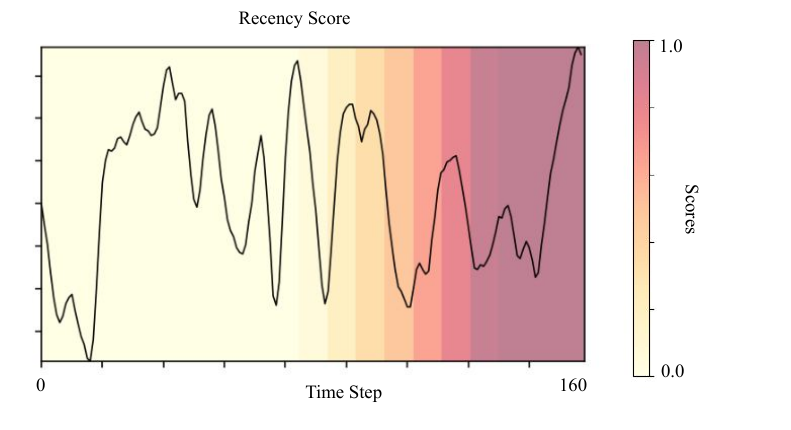}
    \caption{\textbf{Recency score generated by a decay function.} The sample is selected from the Epilepsy dataset.}
    \label{fig:recency_heatmap}
\end{figure}

To understand how each of our proposed gating modules in MSiTF—relevance, recency, and importance—select useful features for different tasks, we visualize the scores assigned to each time window. As shown qualitatively in Figure \ref{fig:heatmap}, the heatmaps reveal that both relevance and importance scores are sensitive to task differences. For example, in the eye closure detection task, the model focuses on the last few patches, whereas in the seizure detection task, it emphasizes patches with large fluctuations. A similar pattern is observed for the importance score, where patches are weighted differently across tasks. This suggests that our gating mechanism can adaptively select relevant features based on the task. We include a figure of the recency score (Figure \ref{fig:recency_heatmap}) for completeness. Since the recency score is derived from a fixed decay function and is not learned, it remains the same across tasks. 

To improve visualization, we aggregated token scores using a window size of 9, which matches our tokenization patch size. We then applied Z-score normalization to ensure comparability across tasks. The sample was selected from the Epilepsy dataset due to its multiple and diverse task types.

\clearpage

\section{Reconstruction Example}
\begin{figure}[htb!]
    \centering
    \begin{minipage}{0.5\textwidth}
        \centering
        \includegraphics[width=\textwidth]{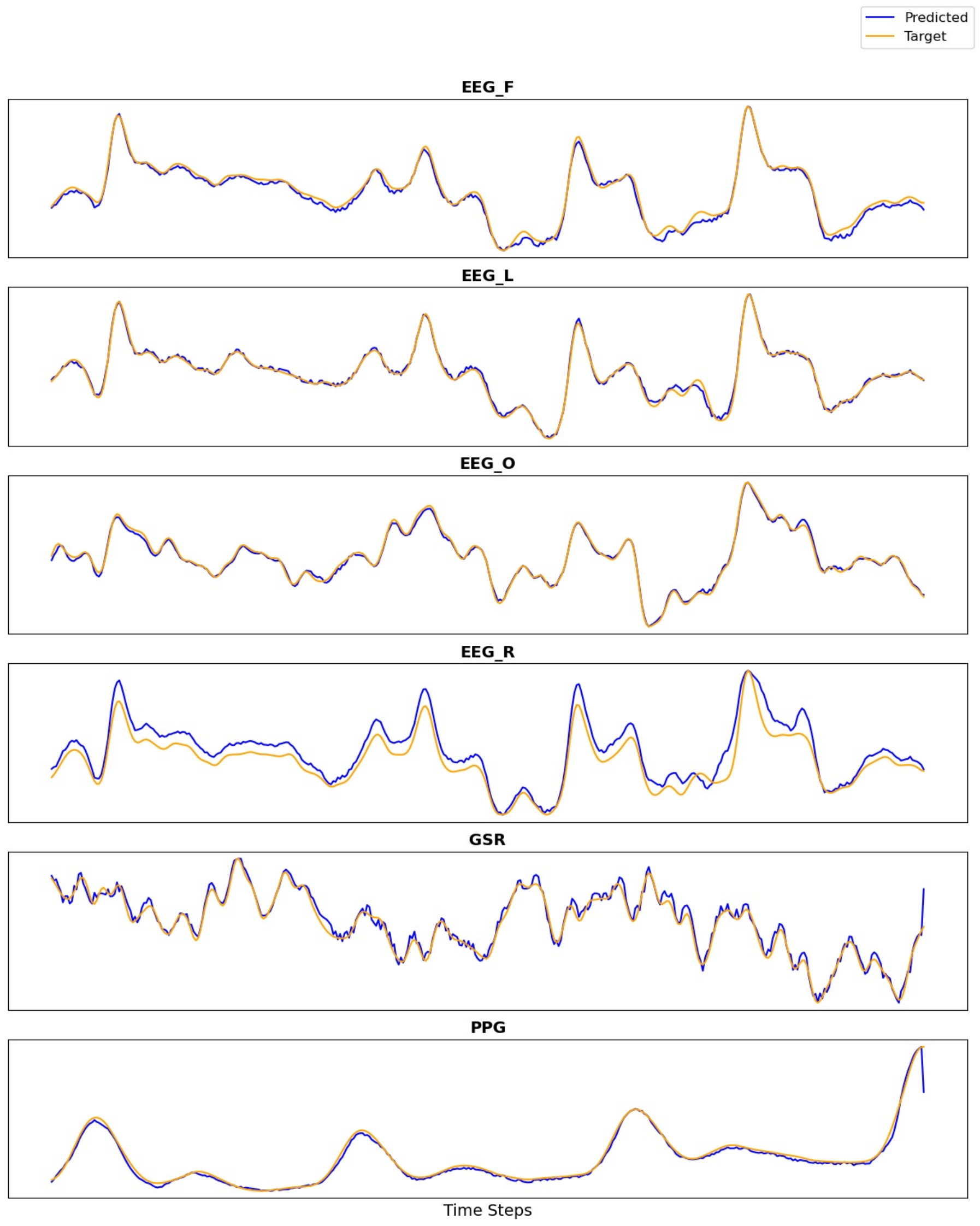}
        \label{fig:first}
    \end{minipage}\hfill
    \begin{minipage}{0.5\textwidth}
        \centering
        \includegraphics[width=\textwidth]{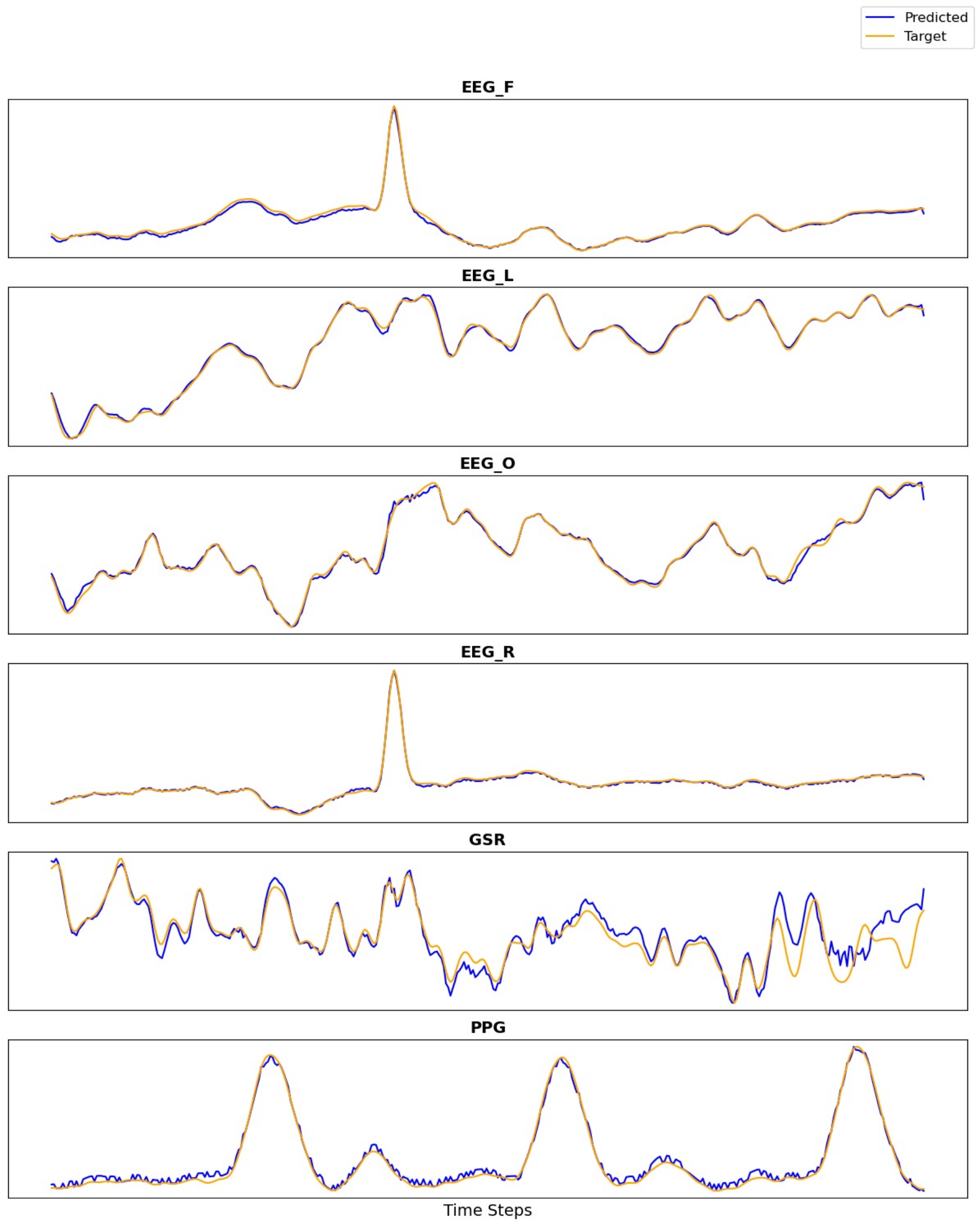}
        \label{fig:second}
    \end{minipage}

    \begin{minipage}{0.45\textwidth}
        \centering
        \includegraphics[width=\textwidth]{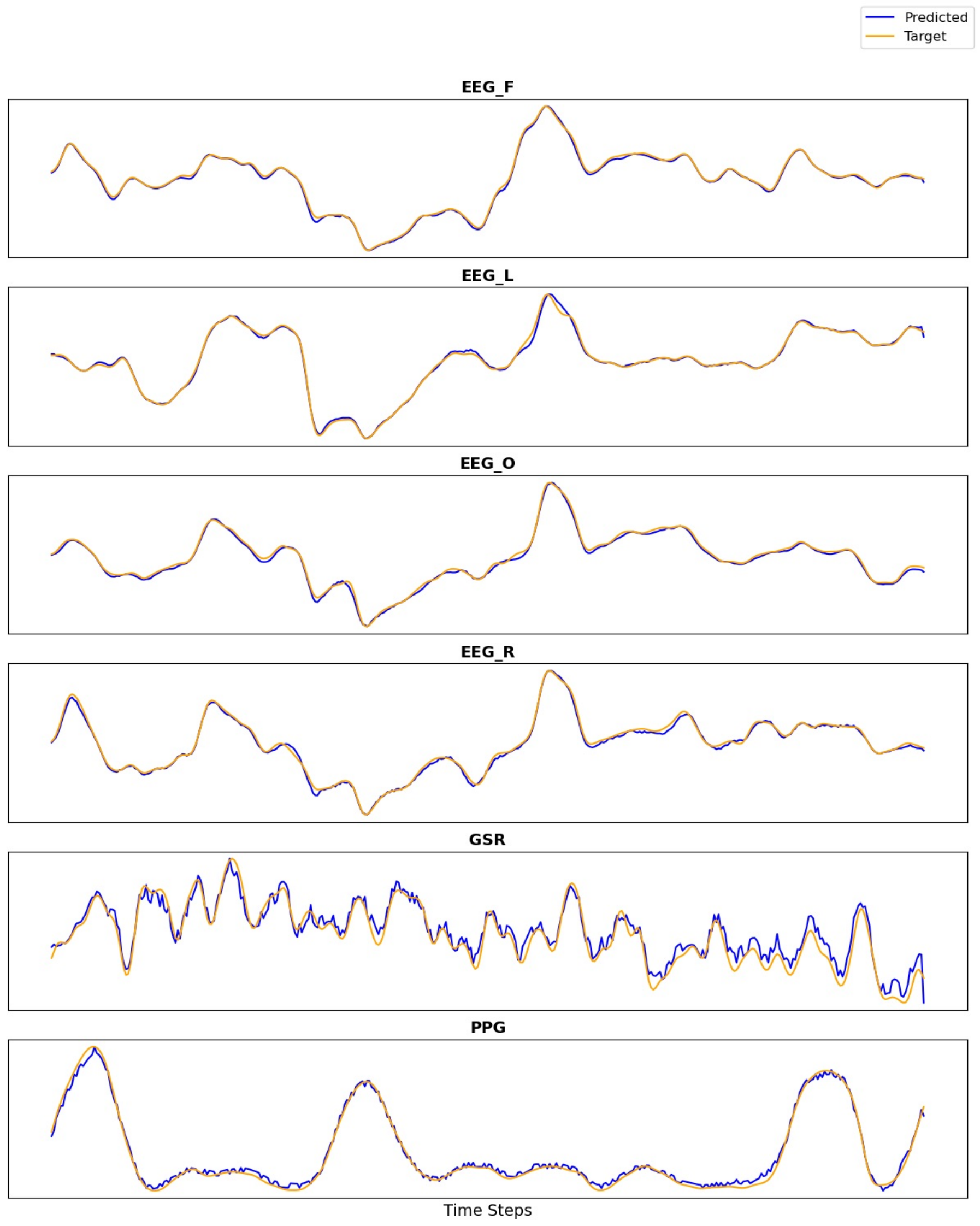}
        \label{fig:third}
    \end{minipage}\hfill
    \begin{minipage}{0.45\textwidth}
        \centering
        \includegraphics[width=\textwidth]{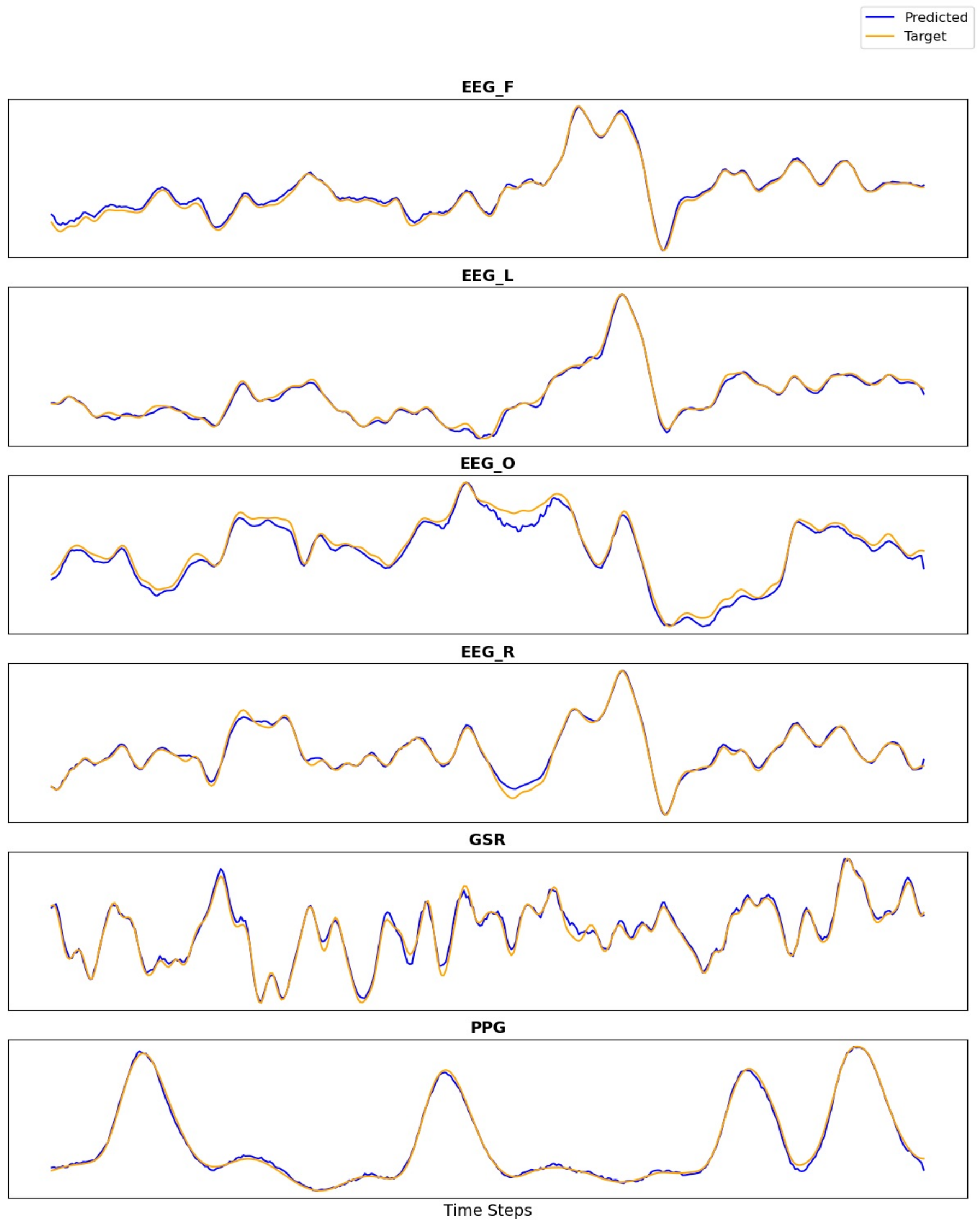}
        \label{fig:fourth}
    \end{minipage}

    \caption{\textbf{Uncurated random samples} on Phyatt scalogram, using a \textsc{NormWear} trained in our training set. The masking ratio is 80\%.}
    \label{fig:overall}
\end{figure}
\begin{figure}[!ht]
    \centering
    \begin{minipage}{0.47\textwidth}
        \centering
        \includegraphics[width=\textwidth]{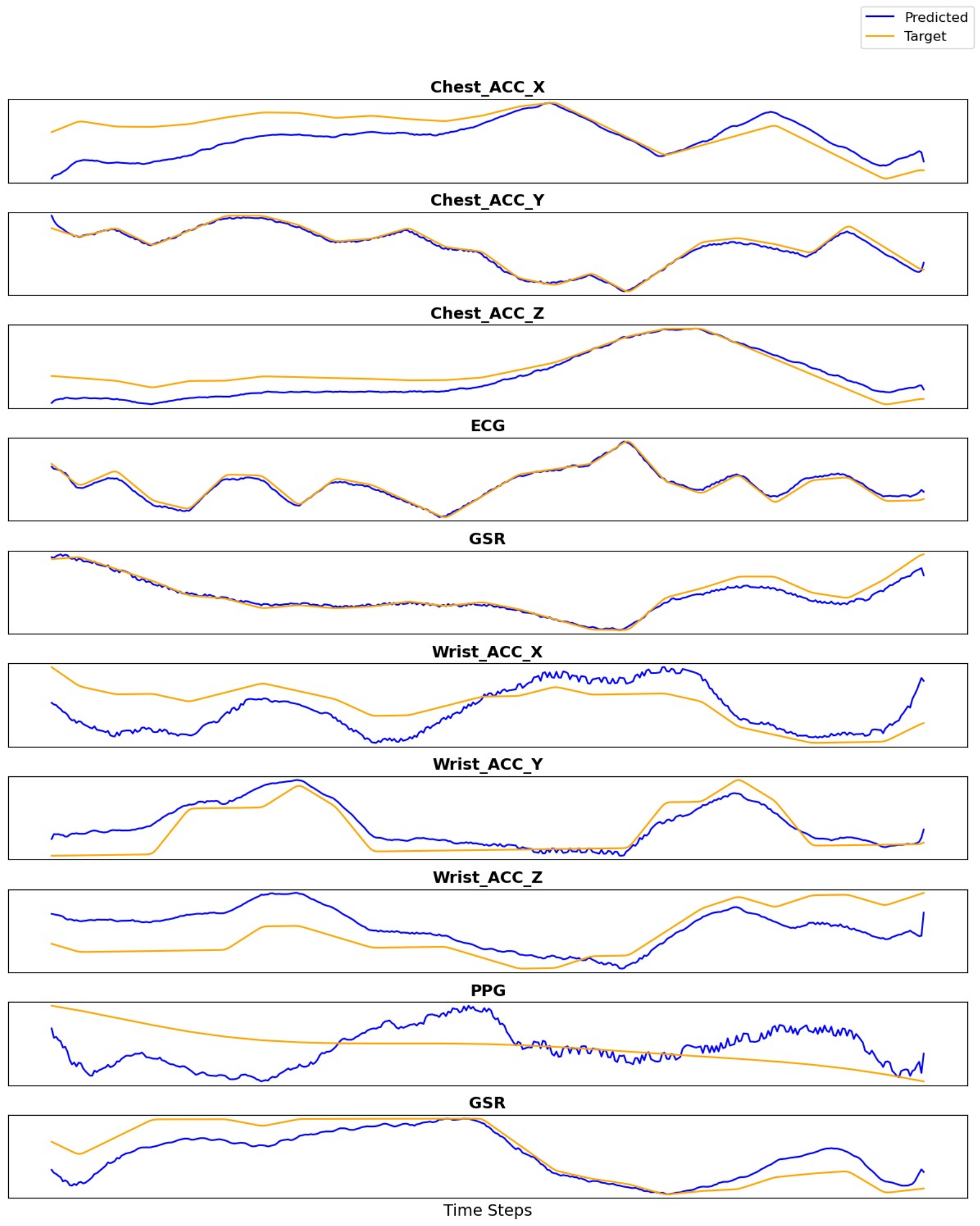}
        \label{fig:wesad-first}
    \end{minipage}\hfill
    \begin{minipage}{0.47\textwidth}
        \centering
        \includegraphics[width=\textwidth]{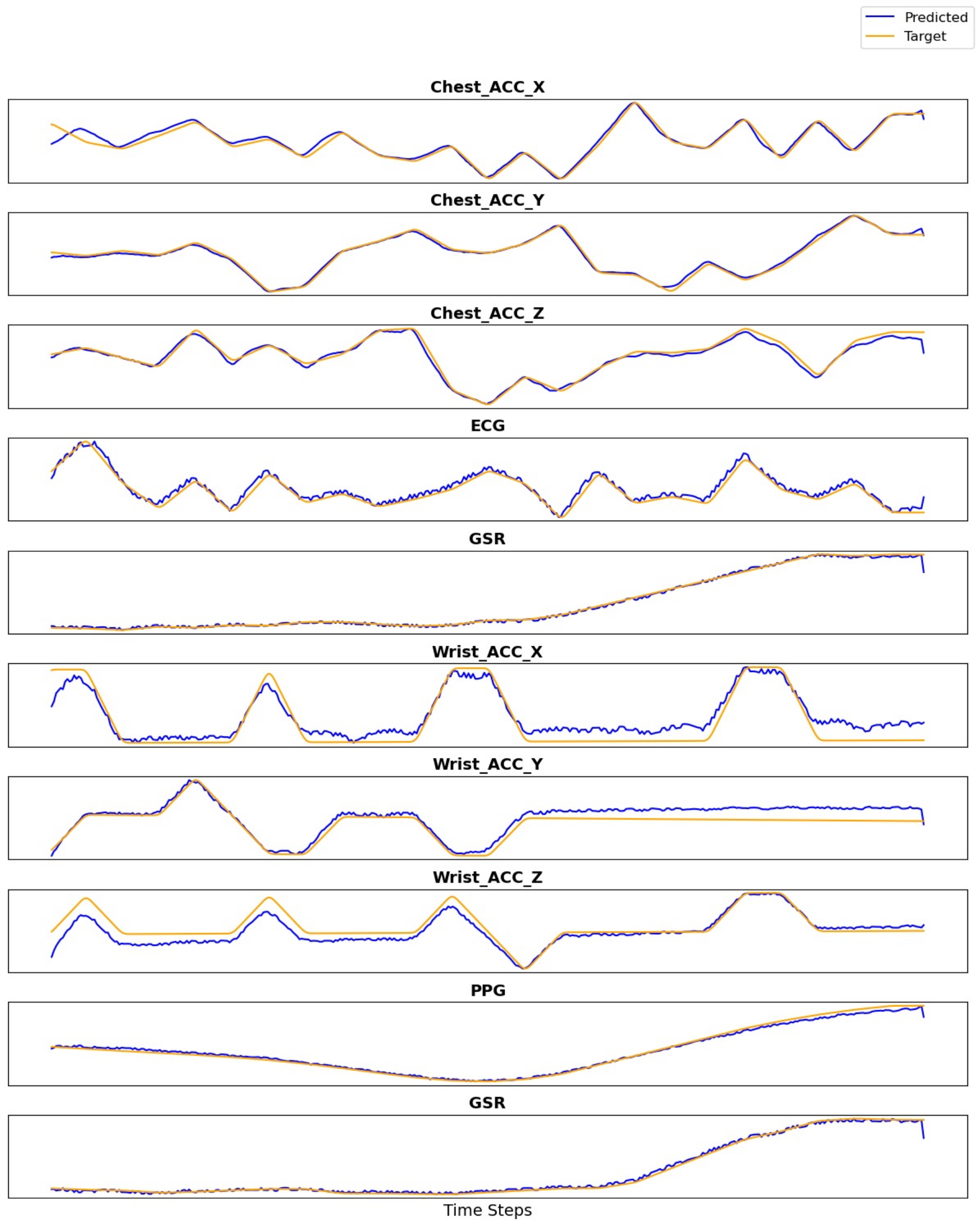}
        \label{fig:wesad-second}
    \end{minipage}

    \begin{minipage}{0.47\textwidth}
        \centering
        \includegraphics[width=\textwidth]{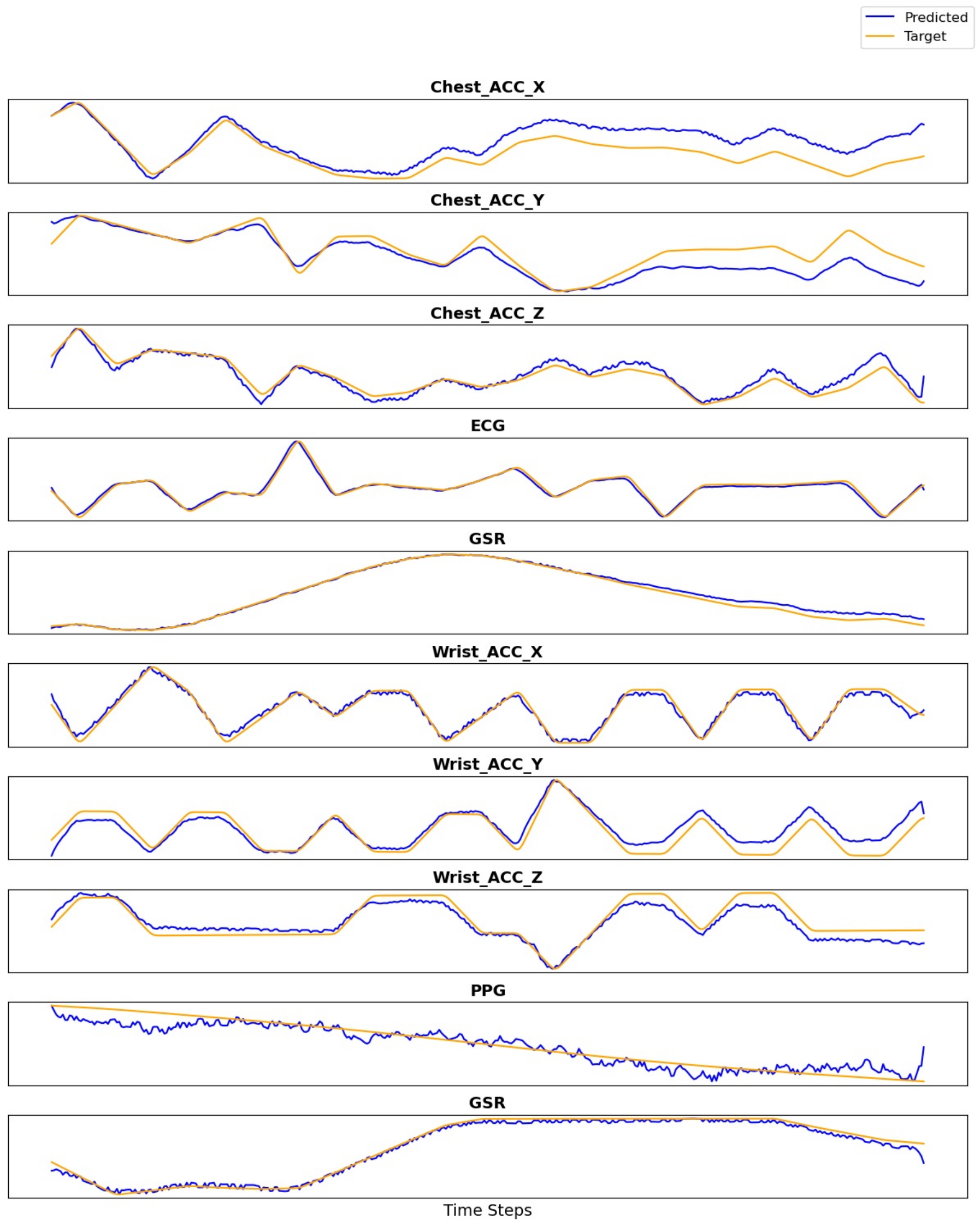}
        \label{fig:wesad-third}
    \end{minipage}\hfill
    \begin{minipage}{0.47\textwidth}
        \centering
        \includegraphics[width=\textwidth]{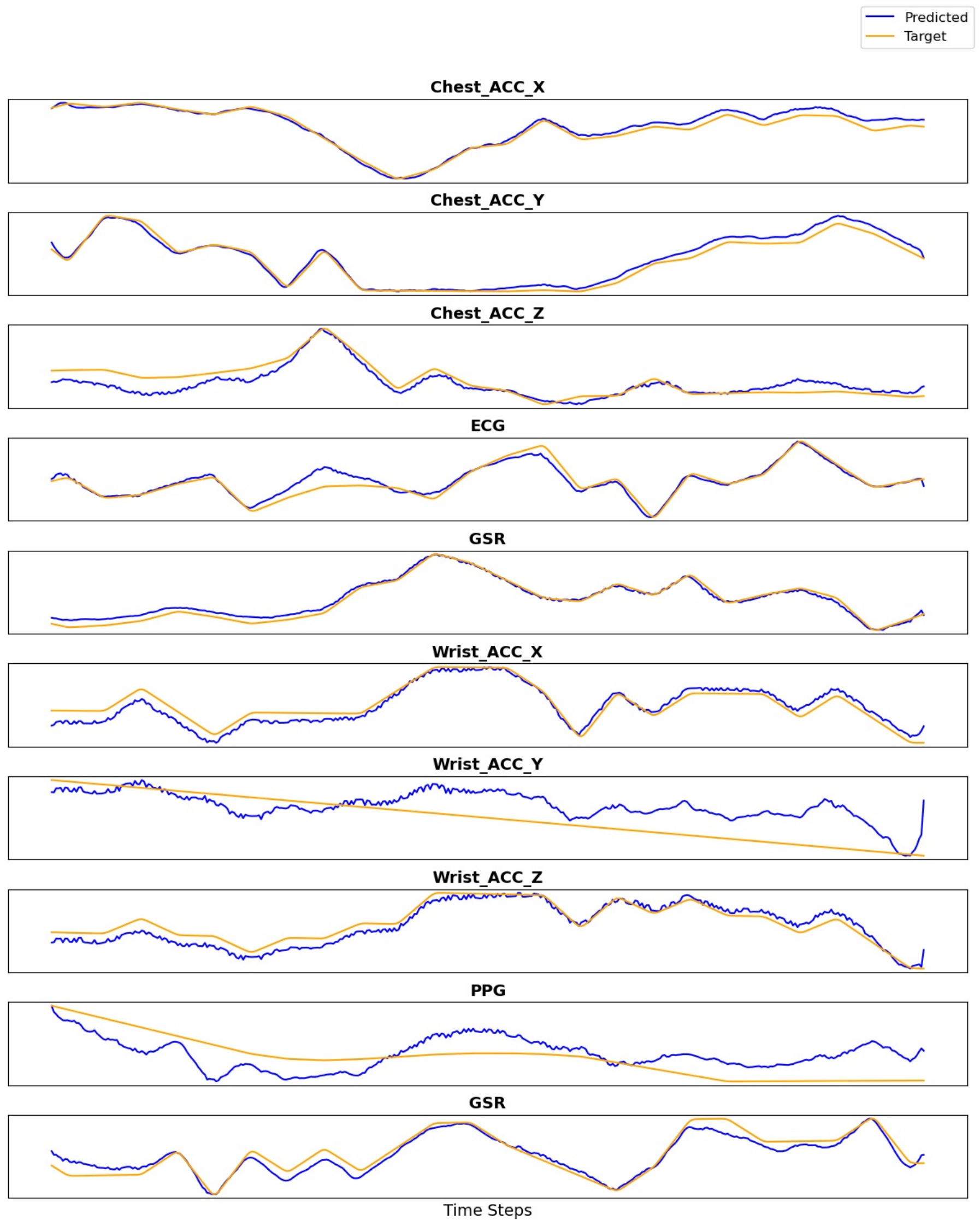}
        \label{fig:wesad-fourth}
    \end{minipage}

    \caption{\textbf{Uncurated random samples} on WESAD scalogram, using a \textsc{NormWear} trained in our training set. The masking ratio is 80\%. Note that the IMU data are not in the training set and, in general, \textsc{NormWear} is able to reconstruct this with high accuracy.}
    \label{fig:wesad-overall}
\end{figure}




\end{document}